\documentclass[pdflatex,sn-mathphys-num,iicol]{sn-jnl}

\usepackage{graphicx}
\usepackage{multirow}
\usepackage{amsmath,amssymb,amsfonts}
\usepackage{booktabs}
\usepackage{xcolor}
\usepackage{makecell}
\usepackage{placeins}
\usepackage[export]{adjustbox}
\usepackage{graphbox}
\usepackage{tikz}
\usetikzlibrary{shapes.geometric, arrows.meta, positioning, calc, fit, backgrounds}

\newcommand{\qualresult}[1]{\includegraphics[width=1.3cm, keepaspectratio, valign=m]{\detokenize{#1}}}
\newcommand{\qualresultners}[1]{\includegraphics[height=1.7cm, width=1.7cm, keepaspectratio, valign=m]{\detokenize{#1}}}
\newcommand{\qualresulth}[1]{\includegraphics[height=1.8cm, keepaspectratio, valign=m]{\detokenize{#1}}}

\DeclareRobustCommand{\flip}{FLIP}
\definecolor{cvprblue}{rgb}{0.21,0.49,0.74}

\renewcommand{\orcid}[1]{\,\href{#1}{\textcolor[HTML]{A6CE39}{\textsuperscript{\scriptsize iD}}}}

\usepackage{cleveref}

\begin{document}

\title[CADSplat]{CADSplat: Sparse-View 3D Gaussian Splatting Aided by CAD Models for Robust, Photorealistic Digital-Twin Reconstruction}

\author*[1]{\fnm{Kristof} \sur{Overdulve}\orcid{https://orcid.org/0000-0002-0535-9798}}\email{kristof.overdulve@uhasselt.be}

\author[1]{\fnm{Lode} \sur{Jorissen}\orcid{https://orcid.org/0000-0001-7228-9315}}\email{lode.jorissen@uhasselt.be}

\author[1]{\fnm{Nick} \sur{Michiels}\orcid{https://orcid.org/0000-0002-7047-5867}}\email{nick.michiels@uhasselt.be}

\affil[1]{\orgdiv{Digital Future Lab, Flanders Make}, \orgname{Hasselt University}, \orgaddress{\city{Hasselt}, \country{Belgium}}}

\abstract{We present CADSplat, a framework that reconstructs photorealistic, geometrically accurate digital twins from sparse ($<15$ views), wide-baseline posed images of an object by regularizing 3D Gaussian Splatting (3DGS) with an explicit CAD shape prior. Using such a prior requires finding a CAD model whose shape resembles the object depicted in the images and determining the pose of each camera relative to the object. We obtain both by matching segmented object silhouettes against silhouettes rendered from a CAD library and keeping the camera-to-object poses of the best-matching model. We then anchor 3D Gaussian primitives to the surface of the retrieved model and jointly optimize the 3DGS parameters, the camera-to-object registration, and a non-rigid deformation field to account for shape differences between the physical object and the CAD model. Across two real-world datasets, CADSplat outperforms unconstrained, few-shot, and mesh-texturing baselines and degrades gracefully to as few as 3 views. Our experiments show that most of the gain in rendering quality comes from how the splats are constrained---a fixed set of splats tied to a surface and moved by a single smooth deformation field---rather than from the CAD shape itself. The CAD model adds shape knowledge where views are scarcest, in the sparsest captures and on strongly self-occluded objects, and it places every camera in the object's own frame. This enables applications beyond novel-view synthesis, such as markerless augmented reality registration, per-image object pose estimation, physical simulations, and the transfer of part labels from the design to the reconstruction.}


\maketitle

\begin{figure*}[t]
    \centering
    \begin{minipage}[b]{0.15\textwidth}
        \centering
        \includegraphics[width=\textwidth]{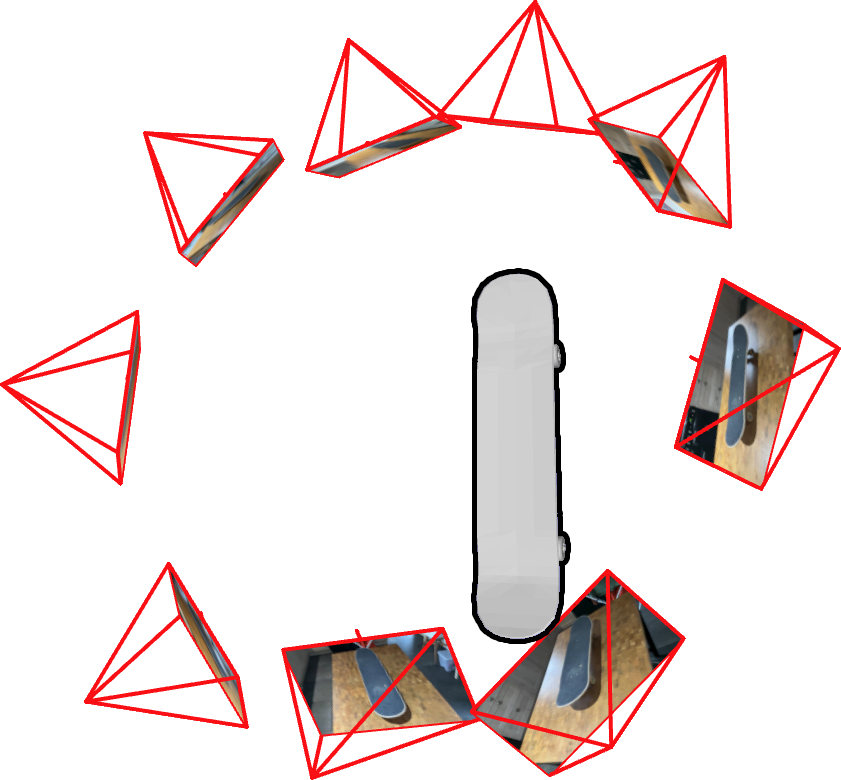}
        \vspace{4pt}

        {\small (a)}
    \end{minipage}
    \hfill
    \begin{minipage}[b]{0.15\textwidth}
        \centering
        \includegraphics[width=\textwidth]{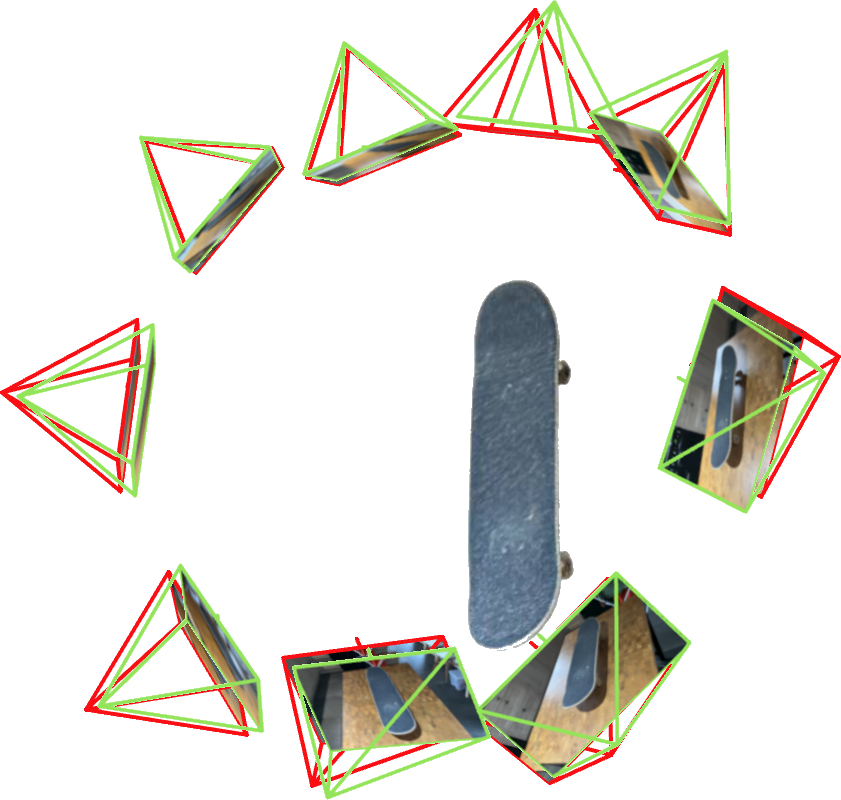}
        \vspace{4pt}

        {\small (b)}
    \end{minipage}
    \hfill
    \begin{minipage}[b]{0.65\textwidth}
        \centering
        \begin{tabular}{@{}c@{\hspace{1pt}}c@{\hspace{1pt}}c@{\hspace{1pt}}c@{\hspace{1pt}}c@{}}
            \includegraphics[width=0.17\textwidth]{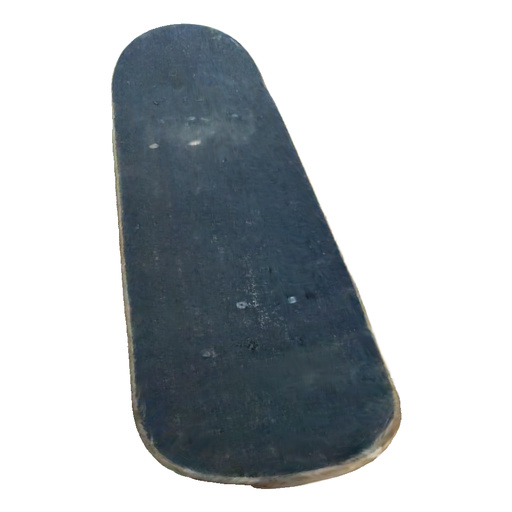} &
            \includegraphics[width=0.17\textwidth]{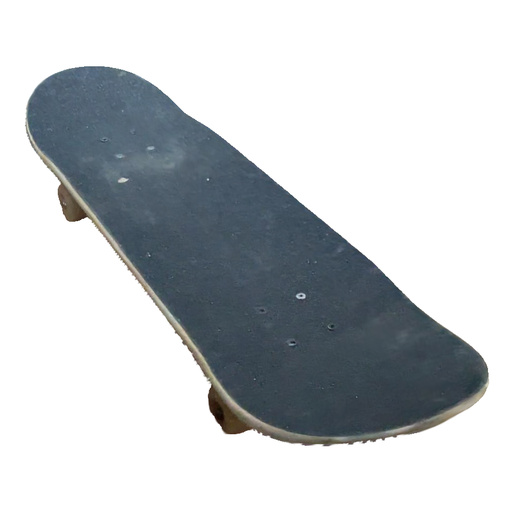} &
            \includegraphics[width=0.17\textwidth]{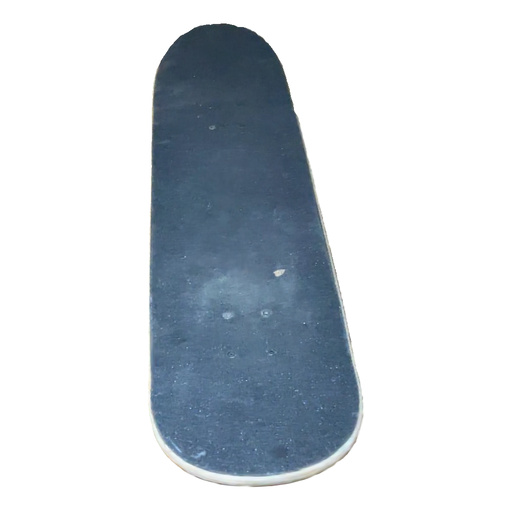} &
            \includegraphics[width=0.17\textwidth]{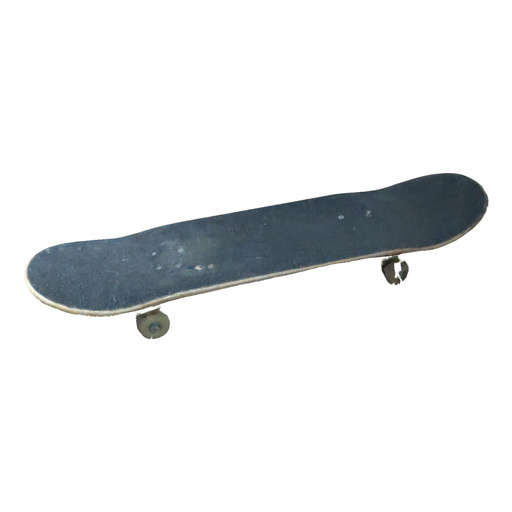} &
            \includegraphics[width=0.17\textwidth]{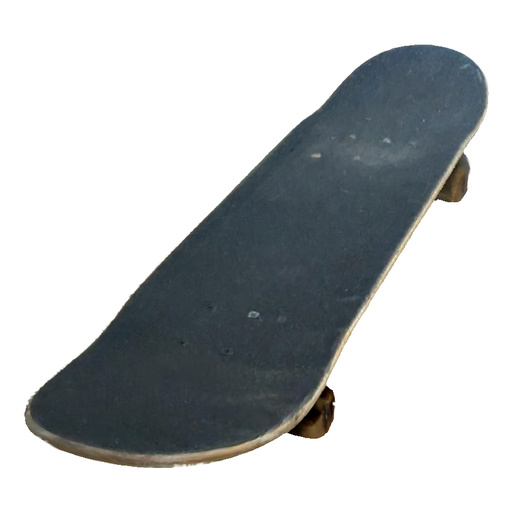}
        \end{tabular}
        \vspace{4pt}

        {\small (c)}
    \end{minipage}
    \caption{\textbf{CADSplat:} Our method regularizes sparse-view 3D Gaussian Splatting (3DGS) using CAD priors. \textbf{(a)} Given a handful of wide-baseline images with segmentation masks depicting the object silhouettes, the masks are matched against renderings of a CAD library, retrieving the most similar CAD model and registering the relative camera extrinsics to it to obtain the camera-to-object poses. \textbf{(b)} A joint optimization refines this registration, learns a non-rigid deformation field that corrects shape discrepancies between the imaged physical object and the matched CAD model, and 3DGS parameters to \textbf{(c)} recover a geometrically accurate, photorealistic 3D representation.}
    \label{fig:teaser}
\end{figure*}

\section{Introduction}
3D representations of physical objects fall into two paradigms: manually designed explicit geometric models and learned 3D representations from camera imagery. Explicit representations, specifically Computer-Aided Design (CAD) models, are commonly used in manufacturing, robotics, and e-commerce. They provide rigid, semantically interpretable, and topologically clean structures that encode the theoretical ground truth of a designed object. However, such models require significant manual labor to create and often lack the textures, material properties, and subtle geometric imperfections necessary for photorealism. Conversely, differentiable novel-view synthesis (NVS) methods such as Neural Radiance Fields (NeRF) \cite{mildenhall2021nerf} and 3D Gaussian Splatting (3DGS) \cite{kerbl20233d} generate photorealistic 3D representations directly from imagery. However, this is an underdetermined problem, especially when views are sparse, causing the geometry of these learned representations to become noisy and prone to hallucinating false geometry to satisfy the training input, yet failing to generalize well to novel views.

The dense-capture assumption underlying these novel-view synthesis algorithms limits their broad applicability. Two common data-capture regimes, in particular, violate it. First, multi-camera rigs, whether deployed on factory lines or as in-vehicle sensor suites, view objects from only a small number of fixed, wide-baseline viewpoints. Such rigs are typically pre-calibrated, so the \emph{relative} geometry of the cameras is approximately known. However, any physical calibration inevitably degrades over time due to mechanical vibration, accidental knocks during maintenance, and setup errors, compounding relative calibration errors. Second, in-the-wild multi-viewpoint captures, such as product listings on websites or photographs documenting insurance claims, are typically made with per-viewpoint aesthetics or damage documentation in mind, yielding sparse, wide-baseline, and incomplete coverage with no pre-calibration, so that relative poses must be recovered from the images themselves. Only recently have learned features and matchers \cite{sarlin2020superglue,sarlin2019coarse} made Structure-from-Motion (SfM) more viable in this regime.

Fortunately, a powerful structural prior is often available: a CAD model of a shape similar to the object depicted in the images. In industrial settings, the manufacturer almost always possesses accurate engineering models. In more general settings, large repositories such as ShapeNet \cite{chang2015shapenet} provide CAD models matching a broad range of real-world objects. \emph{We therefore pursue the ``best of both worlds'' in 3D representations: aligning sparse-view image captures with a topologically accurate CAD model to reconstruct a photorealistic, geometrically accurate digital twin.}

We propose \textbf{CADSplat}, a method that jointly recovers 6-Degrees-of-Freedom (DoF) camera-to-object poses and a photorealistic 3DGS representation \cite{kerbl20233d} of a physical object from a sparse, wide-baseline capture by using CAD priors. This problem is challenging for multiple reasons. First, we need to identify the CAD model whose shape best matches the imaged object and register the relative calibration with that CAD model's coordinate frame. We achieve this by matching segmented object silhouettes against masked CAD library renders (\cref{sec:retrieval}). This silhouette-based camera-to-object registration is only coarsely accurate. We therefore refine it during optimization (\cref{sec:pose_opt}). Second, the matched model rarely matches the physical object perfectly due to machining tolerances, wear, or the absence of the exact design. We bridge this gap with a learnable, non-rigid deformation field that warps the canonical CAD shape onto the physical observations (\cref{sec:deform}). By jointly optimizing the camera-to-object registration, the deformation field, and the 3DGS parameters, we recover improved camera-to-object poses and a photorealistic, view-dependent appearance (\cref{sec:opt}).

In addition to achieving state-of-the-art reconstruction quality, our pipeline enables multiple applications (\cref{sec:applications}) that are not covered by standard NVS algorithms: automated markerless registration for augmented reality-guided maintenance~\cite{kwatra2025splatoverflow}, per-image object-pose estimation for robotics, and physics simulations using the deformed CAD model as a rigid body. The CAD prior also extends 3DGS reconstructions with semantics: because every splat is sampled from a face of the CAD mesh, it inherits the part it sits on, so the model's part decomposition, labels, and metadata transfer onto the photorealistic twin without any manual annotation. Finally, the learned deformation provides an explicit, interpretable record of where the physical object deviates from its CAD design---a first step toward automated inspection.

In summary, our key contributions are:
\begin{enumerate}
    \item A constrained formulation of 3D Gaussian Splatting for sparse views: a fixed, dense splat set anchored to the surface of a deformable shape prior in a canonical frame. It achieves state-of-the-art rendering quality from $9$ to $13$ views and gracefully degrades to $3$ views. 
    \item A silhouette-based camera-to-object registration that retrieves a CAD model from a library, anchors a calibrated sparse capture to its canonical frame, and refines the registration jointly with the reconstruction.
    \item A non-rigid deformation field that warps the CAD geometry onto the physical object, together with two appearance regularizers to generalize better to novel views in the sparse-view regime.
\end{enumerate}

\section{Related Work}

\subsection{Sparse-View Synthesis}
Novel-view synthesis has been fundamentally revolutionized by Neural Radiance Fields (NeRF) \cite{mildenhall2021nerf}, which optimize continuous volumetric functions via ray marching, and by 3D Gaussian Splatting (3DGS) \cite{kerbl20233d}, which represents scenes as explicit 3D ellipsoids. However, these methods are notoriously data-hungry, requiring dense multi-view coverage to produce high-quality novel views. 

Dense capture is often impractical. As a result, specialized sparse-view methods such as FSGS \cite{zhu2024fsgs} and DNGaussian \cite{li2024dngaussian} have been developed that promise high-quality renderings from novel views using dense stereo or monocular depth priors \cite{li2024dngaussian, deng2022depth}, semantic consistency priors \cite{jain2021putting, chen2021mvsnerf}, or both \cite{zhu2024fsgs}. Others approach the problem leveraging generative priors to repair poorly reconstructed regions \cite{yang2024gaussianobject, wu2023reconfusion}. More recently, feed-forward networks regress Gaussian parameters directly from as few as one or two images: pixelSplat \cite{charatan2024pixelsplat}, and MVSplat \cite{chen2024mvsplat} from posed image pairs, Splatter Image \cite{szymanowicz2024splatter}, and SHARP \cite{mescheder2025sharp} from a single photograph. To overcome ambiguity, they pretrain a large-scale regression neural network on large multi-view corpora, leveraging learned priors for novel sparse-view scenes. As their prior is generative, regions that are unobserved or ambiguous are completed with content that is \emph{plausible} under the training distribution, but not necessarily faithful to the actual object depicted in the imagery. Moreover, priors learned from everyday scenes transfer poorly to the reflective, textureless objects common in industrial applications. Geometry foundation models offer a middle ground: InstantSplat \cite{fan2024instantsplat} does not regress Gaussians directly but uses MASt3R to predict camera poses and a dense point cloud from a few unposed images, then optimizes a 3DGS scene from that initialization while jointly refining the poses. Its prior is likewise learned from large-scale data rather than taken from the object at hand. Our method instead imposes a strict, explicit prior from the object's CAD model.

\subsection{Classical Mesh Texturing}
Prior to neural rendering, classical multi-view stereo decoupled geometry from texture mapping. Given a reconstructed or known mesh, MVS-Texturing \cite{Waechter2014Texturing} adds texture from input images onto fixed 3D geometry by solving a Markov Random Field (MRF) optimization to assign textures from an optimal view to each face, using local color adjustments to hide seams. While computationally efficient, these methods are exceptionally sensitive to sparse overlaps, pose inaccuracies, and lack support for view-dependent appearance. In practice, therefore, they are fragile and often produce fragmented, inconsistent textures.

\subsection{Mesh Anchoring}
A growing body of work couples Gaussian primitives with mesh-based structures because of their potential to integrate with existing 3D workflows. SuGaR \cite{guedon2024sugar} regularizes splats toward a surface to enable mesh extraction, GaMeS \cite{waczynska2024games} parameterizes splats on mesh faces for editing, and GaussianAvatars \cite{qian2024gaussianavatars} rigs splats to a parametric head template for animation. These methods anchor splats to a mesh that is either extracted from dense captures or given in known alignment; we instead anchor to a \emph{retrieved} CAD model whose pose relative to the cameras is initially unknown, and use the anchoring as a sparse-view regularizer and an explicit link back to the object's design. Relatedly, deformation fields have been used to model dynamic motion in video \cite{park2021nerfies, yang2024deformable}; we repurpose this architecture for \textit{static} alignment between the canonical CAD space and the physical observations \cite{deng2021deformed}.

\subsection{Pose Estimation and CAD-Based Tracking}
Traditional SfM (COLMAP \cite{schoenberger2016sfm}) expresses camera poses relative to one another, up to a global similarity transformation, without any notion of objects. Per-object model-based trackers \cite{drummond2002real} solve this camera-to-object problem by matching edge features to a projected CAD model, but require a separate pre-training step for every CAD object and often fail to overcome the appearance discrepancy between textureless CAD models and real imagery, returning no viable pose in such cases. Category-level estimators based on Normalized Object Coordinate Space (NOCS) \cite{wang2019normalized} instead regress a canonical-space correspondence and recover a pose against any retrieved CAD model. We adopt a related but lighter-weight strategy for pose initialization: rather than training a per-pixel coordinate regressor, we compare the silhouettes of each candidate CAD model from a dense grid of viewpoints to the object silhouettes in the sparse-view capture, and aggregate the per-view matches into a single multi-view-consistent registration of the calibrated capture. This shares the practical properties that make NOCS attractive, without requiring any learned per-pixel predictor. However, as with NOCS, the resulting poses are rather coarse, necessitating further joint-pose refinement (\cref{sec:pose_opt}).

\subsection{CAD Model Retrieval and Alignment}
A separate line of work picks a CAD model from a repository and aligns it to images of a real scene. Scan2CAD \cite{avetisyan2019scan2cad} learns 9-DoF alignment between CAD models and RGB-D scans; Mask2CAD \cite{kuo2020mask2cad} and ROCA \cite{gumeli2022roca} predict both the model and its pose from a single RGB image, with networks trained on those scan annotations; SPARC \cite{langer2022sparc} improves a single-image alignment by repeatedly rendering it and comparing it to the image; and DiffCAD \cite{gao2024diffcad} learns the same task from synthetic data only. These methods are built for cluttered indoor rooms; they require trained per-category predictors and hand-annotated alignments, and the single-image methods process one frame at a time. Our setting is different: a handful of wide-baseline views of a single object, already calibrated relative to each other. Rather than working with pretrained models that regress camera poses, we perform silhouette matching to propose a pose for each view and use the existing relative calibration to retain mutually consistent proposals. What comes out is a camera-to-object registration of the whole capture, which we then refine together with the reconstruction.

\subsection{Joint Novel-View Synthesis and Pose Estimation}
Analysis-by-synthesis methods such as iNeRF \cite{yen2021inerf} and BARF \cite{lin2021barf} optimize poses and radiance fields simultaneously but require dense image captures. Pose-free pipelines such as NoPe-NeRF \cite{bian2023nopenerf} and COLMAP-Free 3DGS \cite{fu2024colmapfree} remove the SfM preprocessing step altogether but are limited in applicability to video captures, as they assume temporal ordering and small inter-frame distances.

NeRS \cite{zhang2021ners} targets exactly the sparse, wide-baseline, in-the-wild regime we address. It represents the object as a neural deformation of a unit \emph{sphere} \cite{sdf, neus}, discretized into a mesh and Phong-shaded, and jointly optimizes shape, appearance, and camera parameters from roughly 8 views. Instead of deforming a generic sphere, we retrieve an actual CAD model, which both strengthens the geometric prior and provides a canonical frame for the cameras, enabling multiple new applications. Finally, our method inherits 3DGS's real-time rendering, whereas NeRS remains expensive to render.
Methodologically related to our work is CADNeRF \cite{wen2025cad}. We share the high-level idea of using a CAD repository for shape selection and pose initialization, but differ in three ways: (1) we adopt 3DGS rather than a NeRF density field, enabling surface-aligned initialization and real-time rendering; (2) our geometry stays a smooth deformation of the CAD surface---splats can only move through one shared, smooth deformation field, so no splat can drift off on its own into empty space to explain a single training view---whereas CADNeRF only softly regularizes an occupancy field and can still leave density in free space; and (3) our joint pose optimization absorbs coarse silhouette-matching hypotheses without CADNeRF's ordered-image multi-view pose retrieval or its assumption that all cameras observe the object from a similar distance. Closely related in spirit, CADSim \cite{cadsim} reconstructs vehicles for self-driving sensor simulation from sparse, noisy in-the-wild multi-camera and LiDAR data by optimizing a part-aware CAD mesh with a differentiable renderer; it shares our use of CAD priors and joint pose refinement under sparse observation. Its priors, however, are inherently vehicle-bound: a curated set of part-annotated vehicle CAD models and a wheel-articulation model, so applying it to any other object category would require redesigning the method. Our prior is simply a retrieved CAD model, which applies to any category covered by a shape repository.

\section{Methodology}
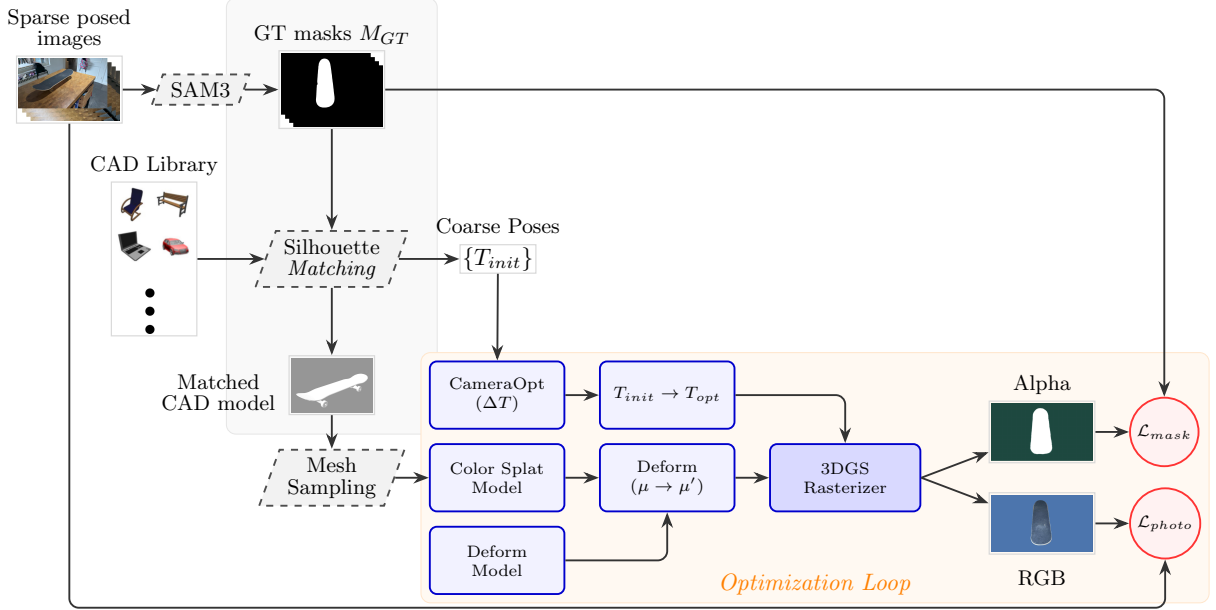
\begin{figure*}[!tp]
    \centering
    \adjustbox{max width=\textwidth}{%
    \begin{tikzpicture}[
        node distance=0.8cm and 1.0cm,
        imgnode/.style={draw=gray!30, line width=0.5pt, inner sep=1pt},
        imglabel/.style={font=\scriptsize, align=center, yshift=-3pt},
        module/.style={rectangle, draw=blue!80!black, fill=blue!5, thick, rounded corners=3pt, minimum height=1cm, minimum width=2cm, align=center, font=\footnotesize},
        fixed/.style={trapezium, trapezium left angle=70, trapezium right angle=110, draw=gray!60!black, fill=gray!10, dashed, thick, align=center, font=\scriptsize},
        loss/.style={circle, draw=red!80, fill=red!5, thick, inner sep=2pt, font=\footnotesize\bfseries},
        flow/.style={-Stealth, thick, draw=gray!40!black, rounded corners},
    ]

    \node[imgnode] (gt_imgs) {\includegraphics[width=1.5cm]{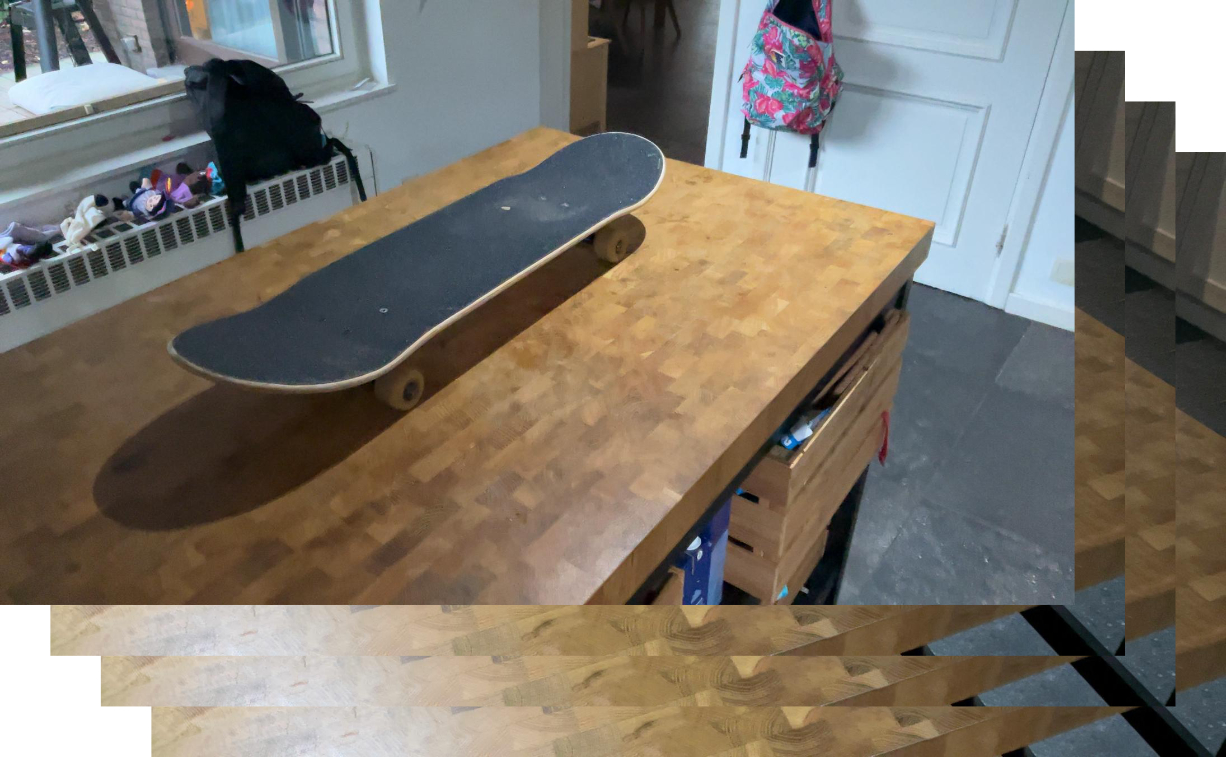}};
    \node[above=2pt of gt_imgs, imglabel] {Sparse posed \\ images};
    
    \node[fixed, right=0.5cm of gt_imgs] (sam) {SAM3};

    \node[imgnode, right=0.5cm of sam] (sam_masks) {\includegraphics[width=1.5cm]{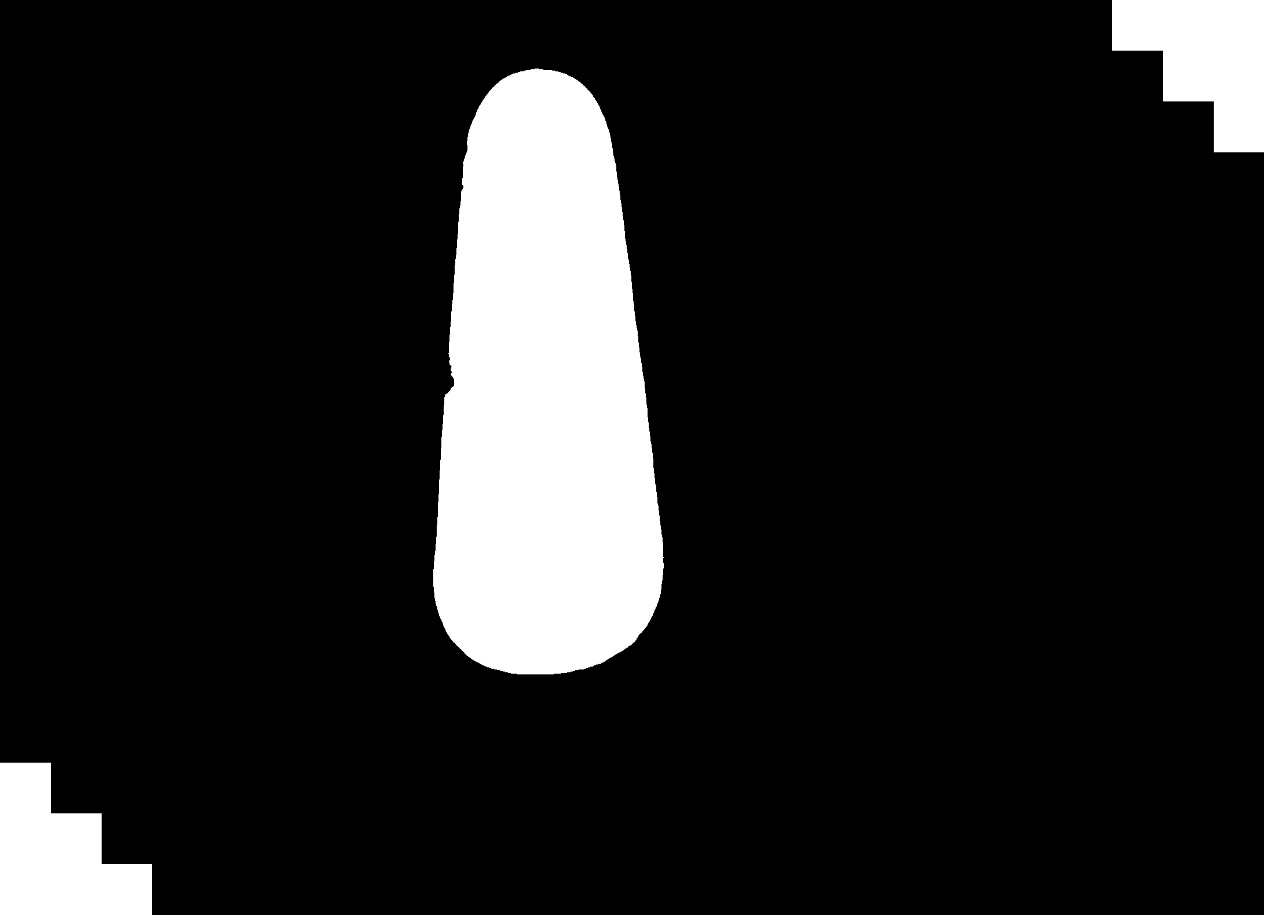}};
    \node[above=2pt of sam_masks, imglabel] (sam_masks_label) {GT masks $M_{GT}$};
     
    \node[fixed, below=1.5cm of sam_masks] (silhouette_match) {Silhouette\\ \textit{Matching}};
    
    \node[imgnode, left=1cm of silhouette_match] (cad_library) {\includegraphics[width=1.2cm]{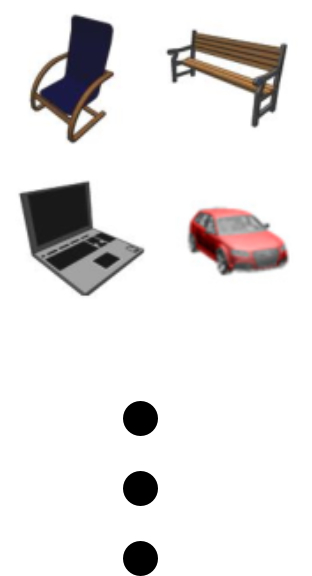}};
    \node[above=2pt of cad_library, imglabel] {CAD Library};
    
    \node[imgnode, below=1cm of silhouette_match] (gt_cad) {\includegraphics[width=1.2cm]{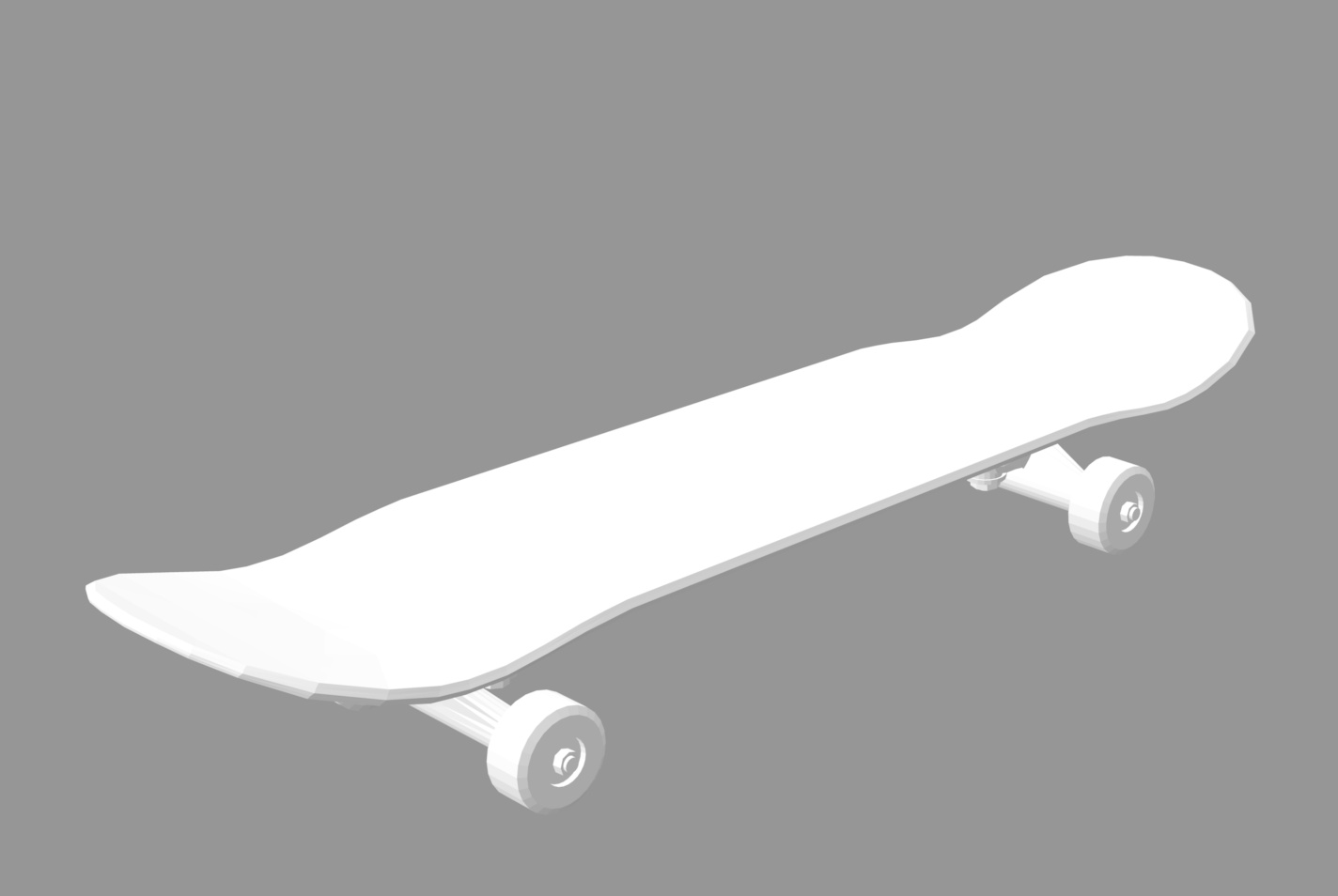}};
    \node[left=2pt of gt_cad, imglabel] {Matched \\ CAD model};
     
    \node[imgnode, right=0.9cm of silhouette_match] (coarse_vis) {$\{T_{init}\}$}; 
    \node[above=4pt of coarse_vis, imglabel] {Coarse Poses};

    \draw[flow] (gt_imgs) -- (sam);
    \draw[flow] (sam) -- (sam_masks);
    
    \draw[flow] (cad_library) -- (silhouette_match);
    \draw[flow] (sam_masks) -- (silhouette_match);
    \draw[flow] (silhouette_match) -- (gt_cad);

    \node[fixed, below=0.5cm of gt_cad] (sampler) {Mesh\\Sampling};
    
    \node[module, right=0.5cm of sampler] (color_splats) {Color Splat\\Model};
    \node[module, above=0.2cm of color_splats] (cam_opt) {CameraOpt\\($\Delta T$)};
    \node[module, below=0.2cm of color_splats] (deform_opt) {Deform\\Model};
    
    \node[module, right=0.5cm of color_splats] (deform_color) {Deform\\($\mu \rightarrow \mu'$)};
    
    \node[module, right=0.5cm of cam_opt] (cam_res) {$T_{init} \rightarrow T_{opt}$};
    
    \node[module, right=0.5cm of deform_color, fill=blue!15, text width=2cm] (raster_color) {3DGS\\Rasterizer};
    
    \node[imgnode, above right=-0.3cm and 1cm of raster_color] (mask_rend) {\includegraphics[width=1.5cm]{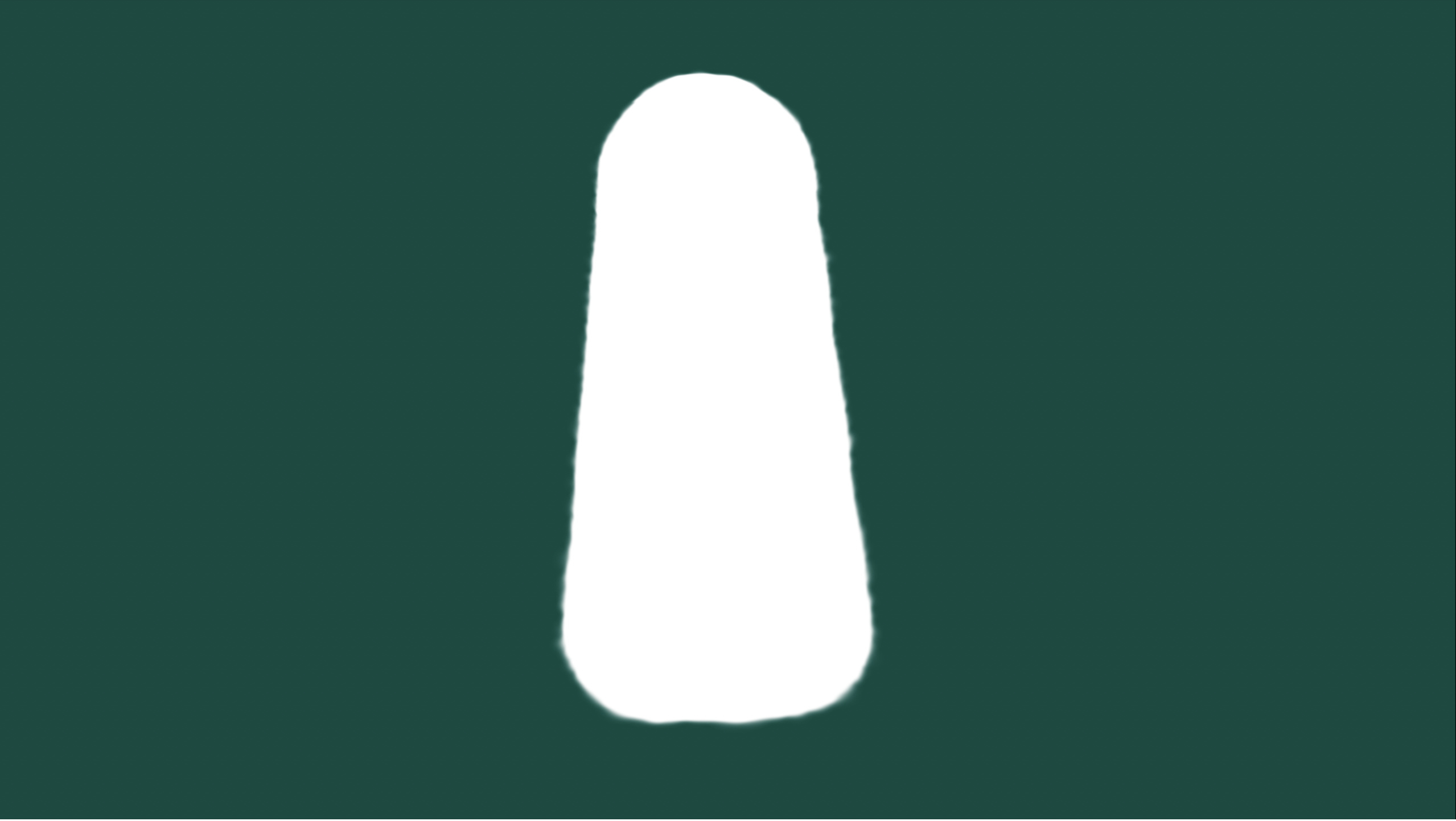}};
    \node[above=2pt of mask_rend, imglabel] {Alpha};
    
    \node[imgnode, below right=-0.3cm and 1cm of raster_color] (color_rend) {\includegraphics[width=1.5cm]{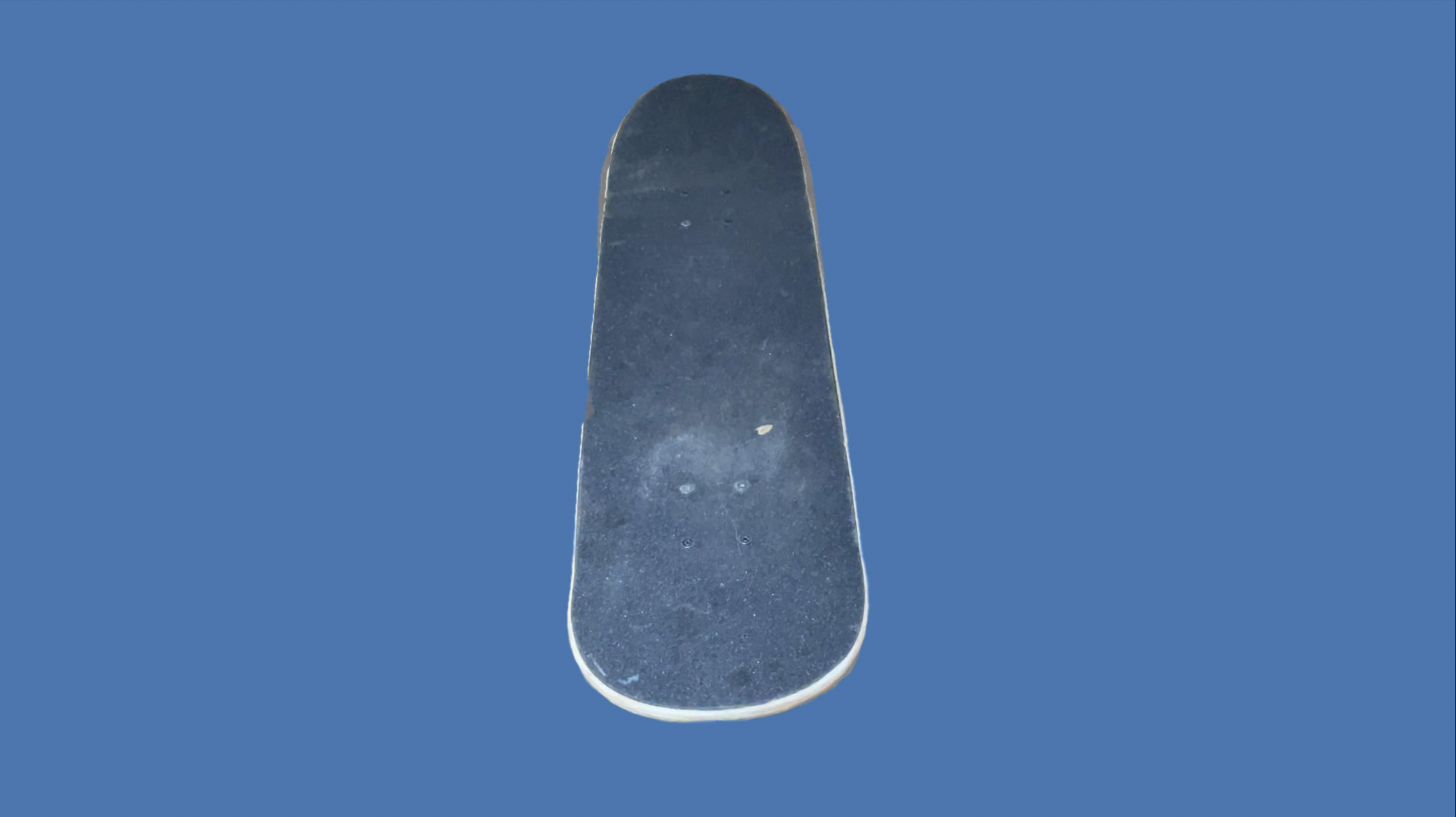}};
    \node[below=0pt of color_rend, imglabel] {RGB};

    \node[loss, right=0.5cm of mask_rend] (l_mask) {$\mathcal{L}_{mask}$};
    \node[loss, right=0.5cm of color_rend] (l_color) {$\mathcal{L}_{photo}$};

    \draw[flow] (silhouette_match) -- (coarse_vis);
    \draw[flow] (coarse_vis) -- (cam_opt); 
    \draw[flow] (cam_opt) -- (cam_res);
    \draw[flow] (cam_res) -| (raster_color);
    \draw[flow] (deform_opt) -| (deform_color);

    \draw[flow] (gt_cad) -- (sampler);
    \draw[flow] (sampler.east) -- (color_splats);
    
    \draw[flow] (color_splats) -- (deform_color);
    \draw[flow] (deform_color) -- (raster_color);

    \draw[flow] (raster_color.east) -- (mask_rend);
    \draw[flow] (raster_color.east) -- (color_rend);

    \draw[flow] (mask_rend) -- (l_mask);
    \draw[flow] (sam_masks) -| (l_mask); 
    
    \draw[flow] (color_rend) -- (l_color);
    \draw[flow] (gt_imgs) -- ++(0, -7.7cm) -| (l_color); 

    \begin{scope}[on background layer]
        \node[fit=(sam_masks_label)(gt_cad)(silhouette_match), draw=gray!20, fill=gray!5, rounded corners, inner sep=8pt, label={[anchor=north, gray]north:\textit{}}] {};
        
        \node[fit=(cam_opt)(color_splats)(l_color)(mask_rend)(deform_opt), draw=orange!20, fill=orange!5, rounded corners, label={[anchor=south, orange]south:\textit{Optimization Loop}}] {};
    \end{scope}

    \end{tikzpicture}%
    }%
    \caption{\textbf{CADSplat overview.} Object silhouettes from SAM3~\cite{carion2025sam3segmentconcepts} are matched against renderings of a CAD library to retrieve the most similar model and coarse camera-to-object poses. 3D Gaussians are sampled on the retrieved mesh, and the deformation network, camera correction, and 3DGS parameters are then optimized jointly. During the geometric warmup, only the rendered alpha channel is supervised against the mask $M_{GT}$ ($\mathcal{L}_{mask}$); afterward, the rendered RGB is compared with the masked photographs ($\mathcal{L}_{photo}$).}
    
    \label{fig:methodology}
\end{figure*}

Our goal is to reconstruct a photorealistic 3D asset and recover precise 6-DoF camera-to-object poses from a sparse set of images $\mathcal{I} = \{I_1, \dots, I_N\}$ with a known \emph{relative calibration}, i.e., camera poses relative to one another obtained from manual pre-calibration or Structure-from-Motion. As shown in \Cref{fig:methodology}, given a library of candidate CAD models, we retrieve the most similar CAD mesh and a coarse \emph{camera-to-object registration} of the calibrated camera trajectory to its canonical frame by matching the observed object silhouettes against the CAD library's masked renderings (\cref{sec:retrieval}), which is later refined differentiably (\cref{sec:pose_opt}). We anchor our Gaussian primitives strictly to the surface of the retrieved CAD mesh (\cref{sec:init}), accounting for potential geometric discrepancies between the CAD model and observations of the physical object via a learnable non-rigid deformation field (\cref{sec:deform}). Our optimization pipeline (\cref{sec:opt}) proceeds in two stages: a \textit{Geometric Warmup} that optimizes the pose estimates and the deformation field differentiably to align the ground-truth segmentation masks with the rendered alpha channel, followed by a \textit{Photometric Finetuning} stage that recovers view-dependent appearance by optimizing 3DGS attributes.

\subsection{Preliminaries: 3D Gaussian Splatting}
We adopt 3D Gaussian Splatting (3DGS) \cite{kerbl20233d} as our underlying representation. Each 3D Gaussian $g_i$ is defined by a center position $\mu_i \in \mathbb{R}^3$, three scaling factors $s_i \in \mathbb{R}^3$, a rotation quaternion $q_i \in \mathbb{R}^4$, and an opacity value $\alpha_i \in \mathbb{R}$. The view-dependent color is parameterized using spherical harmonics (SH) coefficients. To render an image, the 3DGS primitives are projected into screen space using splatting and $\alpha$-blending.

\subsection{Silhouette-Based CAD Matching}
\label{sec:retrieval}
Our pipeline begins with the relative calibration: a pre-calibrated rig or an SfM run on the sparse training views. This calibration is expressed in an arbitrary coordinate frame, locating the cameras relative to one another but not to the object. Silhouette matching identifies the most similar CAD model and anchors the camera poses in the canonical frame of the matched object by comparing observed object silhouettes with mask renderings from a CAD model library. 

For each posed input image of the sparse-view capture, we extract a binary foreground mask $M_{GT}$ with the Segment Anything Model 3 (SAM3) \cite{carion2025sam3segmentconcepts}. While this may seem like a tedious step, querying objects based on text prompts works well in practice, and in static multi-camera setups, obtaining object masks of moving objects is trivial. We then query a library of candidate CAD models. Each candidate is rendered with white textures, no environment lighting, and a black background, using a dense grid of viewpoints sampled over the viewing sphere. To compare silhouettes independently of where the object appears in the image or how large it is, we normalize every mask---both the rendered library silhouettes and the observed SAM3 masks---by cropping it to its tight bounding box and resizing it into a fixed-size square canvas while preserving aspect ratio. For each observed mask, we retain the top $50$ viewpoints whose normalized silhouettes achieve an Intersection-over-Union (IoU) of $0.7$ or higher as camera-to-object pose hypotheses. We render this viewpoint grid once per library, offline and in parallel, before any capture is taken. Matching a new capture then only compares its masks against these stored silhouettes.

\begin{figure}[!htbp]
    \centering
    \setlength{\tabcolsep}{2pt}
    \begin{tabular}{c}
        \includegraphics[width=0.9\linewidth]{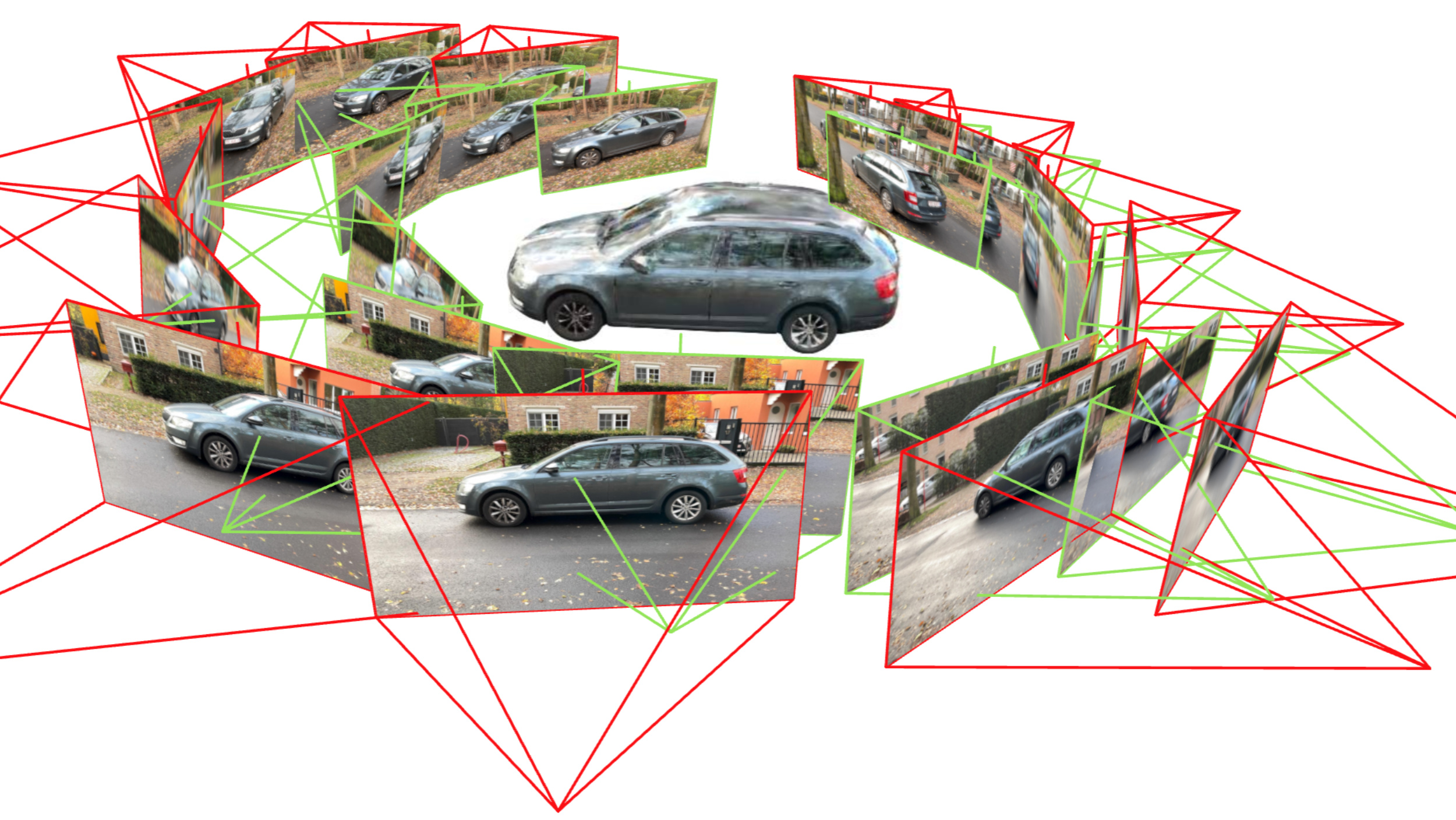} \\
        {\small (a) Multi-view consensus registration (ours)} \\
        \includegraphics[width=0.9\linewidth]{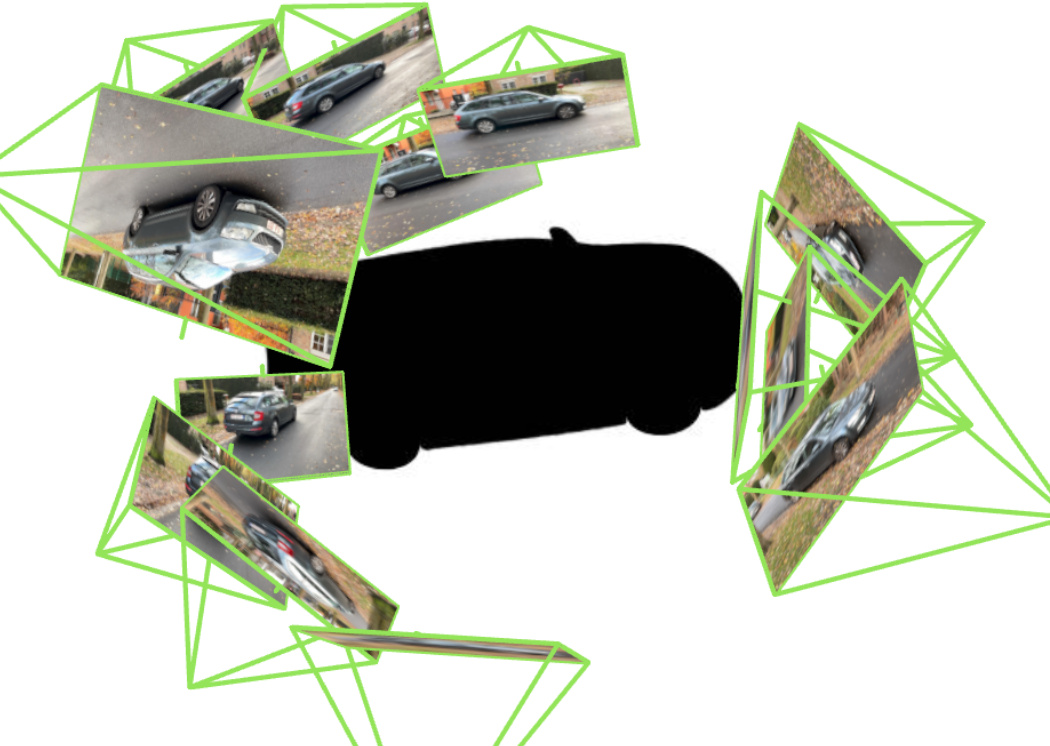} \\
        {\small (b) Independent per-camera silhouette matching}
    \end{tabular}
    \caption{Joint versus independent camera-to-object registration. \textbf{(a)} Our consensus registration (\cref{sec:retrieval}) recovers one similarity transform for the whole calibrated trajectory, so the initial poses $\mathbf{T}_{init}$ (red) form a consistent ring around the CAD model, which the joint optimization (\cref{sec:pose_opt}) refines to $\mathbf{T}_{opt}$ (green). \textbf{(b)} Matching each camera independently to its best library silhouette scatters the cameras: a silhouette alone does not fix a viewpoint, so each view picks a different, often mirrored, orientation. Using the relative calibration as a RANSAC consistency filter removes these mutually inconsistent matches.}
    \label{fig:alignment}
\end{figure}

These per-image hypotheses do not replace the need for the relative calibration, as they are individually unreliable. A silhouette does not uniquely determine a pose when objects are near symmetric from some viewpoints. The viewpoint library is also discrete, so the true viewing angle generally falls between the sampled angles, and an incorrect pose can then score a higher IoU than the correct one. We therefore never use these hypotheses in isolation. Instead, we combine them with the relative calibration information, treating it as a consistency filter. Concretely, we use RANSAC over $\mathrm{Sim}(3)$. Each iteration samples one pose hypothesis from each of two randomly sampled images and computes the transformation to register the relative calibration into the object's canonical frame (rotation and translation from the first hypothesis, scale from the ratio of inter-camera distances). It then scores the candidate by rendering the CAD silhouette in \emph{all} views and accumulating per-view IoU scores. The CAD model $\mathcal{M}$ and transformation $\mathbf{T}_{init}$ with the highest aggregate support are then selected. Applying $\mathbf{T}_{init}$ to the relatively calibrated cameras yields the initial per-camera camera-to-object poses $\mathbf{T}_{init}^{(i)}$. In this way, a single discriminative viewpoint suffices to anchor the trajectory correctly, even when other views are individually ambiguous. \Cref{fig:alignment} makes the difference visible: our consensus registration places the cameras in a consistent ring around the object (\cref{fig:alignment}a), whereas matching each camera to its closest library silhouette independently scatters them incoherently (\cref{fig:alignment}b).

Note that this registration works well even for texture-less or reflective objects: the relative calibration can rely predominantly on background features in such cases, while silhouette matching consumes only silhouettes and is therefore unaffected by material properties.

\subsection{Camera Pose Optimization}
\label{sec:pose_opt}
The silhouette-based registration from \cref{sec:retrieval} is only coarsely accurate and must be refined jointly with the reconstruction. We model this as a learnable pose correction optimized together with the reconstruction. When the relative calibration is accurate, as on our own SCO-CAD dataset (\cref{sec:scocad}), a single global correction $\Delta \mathbf{T}_{g}$ shared by all cameras suffices:
\begin{equation}
\mathbf{T}_{opt}^{(i)} \;=\; \Delta \mathbf{T}_g \cdot \mathbf{S}_{s}\!\bigl(\mathbf{T}_{init}^{(i)}\bigr),
\qquad
\Delta \mathbf{T}_g \;=\; \begin{bmatrix}\mathbf{R}_{\delta}(\boldsymbol{r}_{\delta}) & \boldsymbol{t}_{\delta} \\ \mathbf{0} & 1\end{bmatrix},
\label{eq:pose}
\end{equation}
where $\mathbf{S}_{s}$ scales the camera center by a learnable scalar $s = \exp(\Delta s)$ that absorbs the unknown scale between the calibrated trajectory and the CAD canonical frame. For a camera-to-world pose $(\mathbf{R}_i, \mathbf{c}_i)$, this means $\mathbf{c}_i \mapsto \mathbf{R}_{\delta}(s\,\mathbf{c}_i) + \boldsymbol{t}_{\delta}$ and $\mathbf{R}_i \mapsto \mathbf{R}_{\delta}\mathbf{R}_i$. Following Zhou et al. \cite{zhou2019continuity}, the rotation is parameterized as a continuous 6D vector $\boldsymbol{r}_{\delta} \in \mathbb{R}^6$ mapped to $\mathbf{R}_{\delta} \in SO(3)$ via Gram--Schmidt to avoid the discontinuities of Euler angles or quaternions. All ten parameters (translation, 6D rotation, log-scale) are initialized to the identity correction. When the provided poses are themselves per-view unreliable rather than only globally misaligned, as in the NeRS~\cite{zhang2021ners} datasets, the same module instead learns an independent correction $\Delta \mathbf{T}^{(i)}$ per camera.

\subsection{Geometric Initialization}
\label{sec:init}
Rather than initializing the 3DGS representation from the SfM sparse point cloud, we sample splats directly from the retrieved CAD mesh $\mathcal{M}$. We define a target number of primitives $N_{splats}$ and draw an area-weighted uniform random sample of points from the mesh surface to obtain the initial canonical positions $\mu_i^{can}$. The associated surface normals $n_i$ are recovered by casting a ray from each sample back toward the mesh interior and reading the triangle normal of the intersected face. We initialize the rotation quaternion $q_i$ of each Gaussian to the shortest-arc rotation that aligns its local $z$-axis with $n_i$, so each primitive begins as an oriented disk tangent to the mesh surface. This provides a strong initial inductive bias. Importantly, we do \textit{not} employ densification or pruning strategies (e.g., cloning or splitting) used in standard 3DGS. 

\subsection{Deformable Mesh-to-Reality Alignment}
\label{sec:deform}
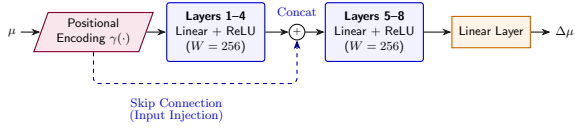
\begin{figure}[!htbp]
    \centering
    \resizebox{\columnwidth}{!}{%
    \begin{tikzpicture}[
        node distance=0.6cm and 1.2cm,
        font=\sffamily\small,
        tensor/.style={draw=gray!80, fill=white, thick, minimum width=0.8cm, minimum height=0.6cm, align=center},
        fixed/.style={trapezium, trapezium left angle=70, trapezium right angle=110, draw=purple!70!black, fill=purple!10, thick, minimum height=1.0cm, align=center},
        mlp/.style={rectangle, draw=blue!80!black, fill=blue!5, thick, rounded corners=2pt, minimum height=1.5cm, minimum width=2.5cm, align=center},
        head/.style={rectangle, draw=orange!80!black, fill=orange!10, thick, minimum height=0.8cm, minimum width=2.0cm, align=center},
        act/.style={trapezium, trapezium left angle=120, trapezium right angle=60, draw=red!70!black, fill=red!10, thick, font=\scriptsize},
        arrow/.style={-Stealth, thick, draw=gray!20!black},
        skip/.style={-Stealth, dashed, thick, draw=blue!60!black, rounded corners}
    ]
    \node (input) {$\mathbf{\mu}$};
    \node[fixed, right=0.5cm of input] (posenc) {Positional\\Encoding $\gamma(\cdot)$};
    \node[mlp, right=0.5cm of posenc] (block1) {\textbf{Layers 1--4} \\ Linear + ReLU \\ ($W=256$)};
    \node[circle, draw, inner sep=0pt, minimum size=0.4cm, right=0.6cm of block1] (concat) {\footnotesize $+$};
    \node[above=2pt of concat, font=\scriptsize, blue] {Concat};
    \node[mlp, right=0.5cm of concat] (block2) {\textbf{Layers 5--8} \\ Linear + ReLU \\ ($W=256$)};
    \node[head, right=0.7cm of block2] (head_pos) {Linear Layer};
    \node[right=0.5cm of head_pos] (out_pos) {$\Delta \mu$};
    \draw[arrow] (input) -- (posenc);
    \draw[arrow] (posenc) -- (block1);
    \draw[arrow] (block1) -- (concat);
    \draw[arrow] (concat) -- (block2);
    \draw[arrow] (block2.east) -- (head_pos);
    \draw[arrow] (head_pos) -- (out_pos);
    \draw[skip] (posenc.south) -- ++(0, -0.8) -| (concat.south);
    \node[below=0.8cm of block1, font=\scriptsize, blue, align=center, xshift=-1cm] {Skip Connection\\(Input Injection)};
    \end{tikzpicture}%
    }
    \caption{\textbf{Architecture of the deformation network $\mathcal{D}_\theta$.} The network maps a canonical coordinate $\mathbf{\mu}$ to high-dimensional features via an 8-layer MLP. A skip connection injects the positional encoding at Layer 4. A single linear head then predicts the position offset $\Delta \mu$.}
    \label{fig:deform}
\end{figure}

To correct discrepancies between the shape of $\mathcal{M}$ and the physical objects depicted in the images --- whether because no exact CAD model is present in the CAD library or due to manufacturing tolerances or errors --- we introduce a coordinate-based deformation network $\mathcal{D}_\theta$ similar to Yang et al.~\cite{yang2024deformable}. $\mathcal{D}_\theta$ is parametrized by a Multi-Layer Perceptron (MLP) with $D=8$ layers, width $W=256$, and a skip connection at the middle layer as depicted in \figurename~\ref{fig:deform}.

The network takes the canonical splat position $\mu$ as input and predicts a position offset $\Delta \mu$:

\begin{equation}
    \Delta \mu = \mathcal{D}_\theta(\gamma(\mu)),
\end{equation}
where $\gamma(\cdot)$ is a sinusoidal positional encoding. The final deformed position of the splat in world space is given by:
\begin{equation}
    \mu' = \mu + \Delta \mu.
\end{equation}
The output head is initialized such that $\Delta \mu \approx 0$ at the start of training. Note that we define the deformation in the canonical space. Therefore, the learned deformation represents a correction to the mesh shape shared across all viewpoints. Because $\mathcal{D}_\theta$ is a smooth coordinate MLP predicting offsets from this shared canonical shape, it injects a spatial smoothness prior that standard 3DGS lacks, which helps in avoiding overfitting deformations to sparse training views.

\subsection{Optimization Pipeline}
\label{sec:opt}
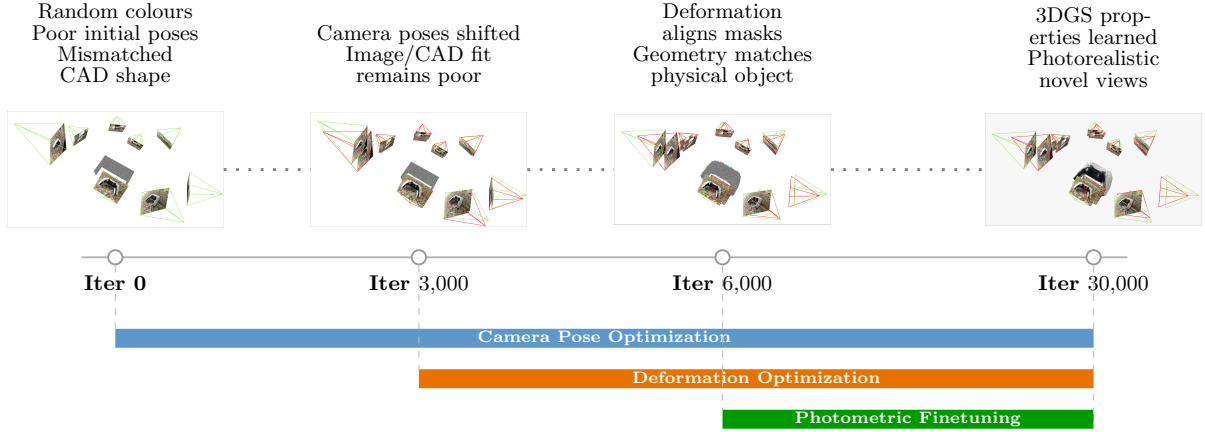
\begin{figure*}[!tp]
    \centering
    \adjustbox{max width=\textwidth}{%
    \begin{tikzpicture}[
        image node/.style={inner sep=0pt, minimum width=3.2cm, minimum height=2.2cm},
        timeline mark/.style={circle, draw=gray!80, fill=white, thick, minimum size=6pt, inner sep=0pt},
        phase bar/.style={line width=8pt, rounded corners=3pt},
        phase label/.style={font=\footnotesize\bfseries, white, anchor=west, xshift=2pt},
        desc/.style={font=\scriptsize, align=center, text width=3.2cm, anchor=south}
    ]

    \def\xO{0}
    \def\xA{4.5}
    \def\xB{9.0}
    \def\xC{14.5} 

    \node[image node] (img0) at (\xO, 1.3) {\includegraphics[width=3.2cm]{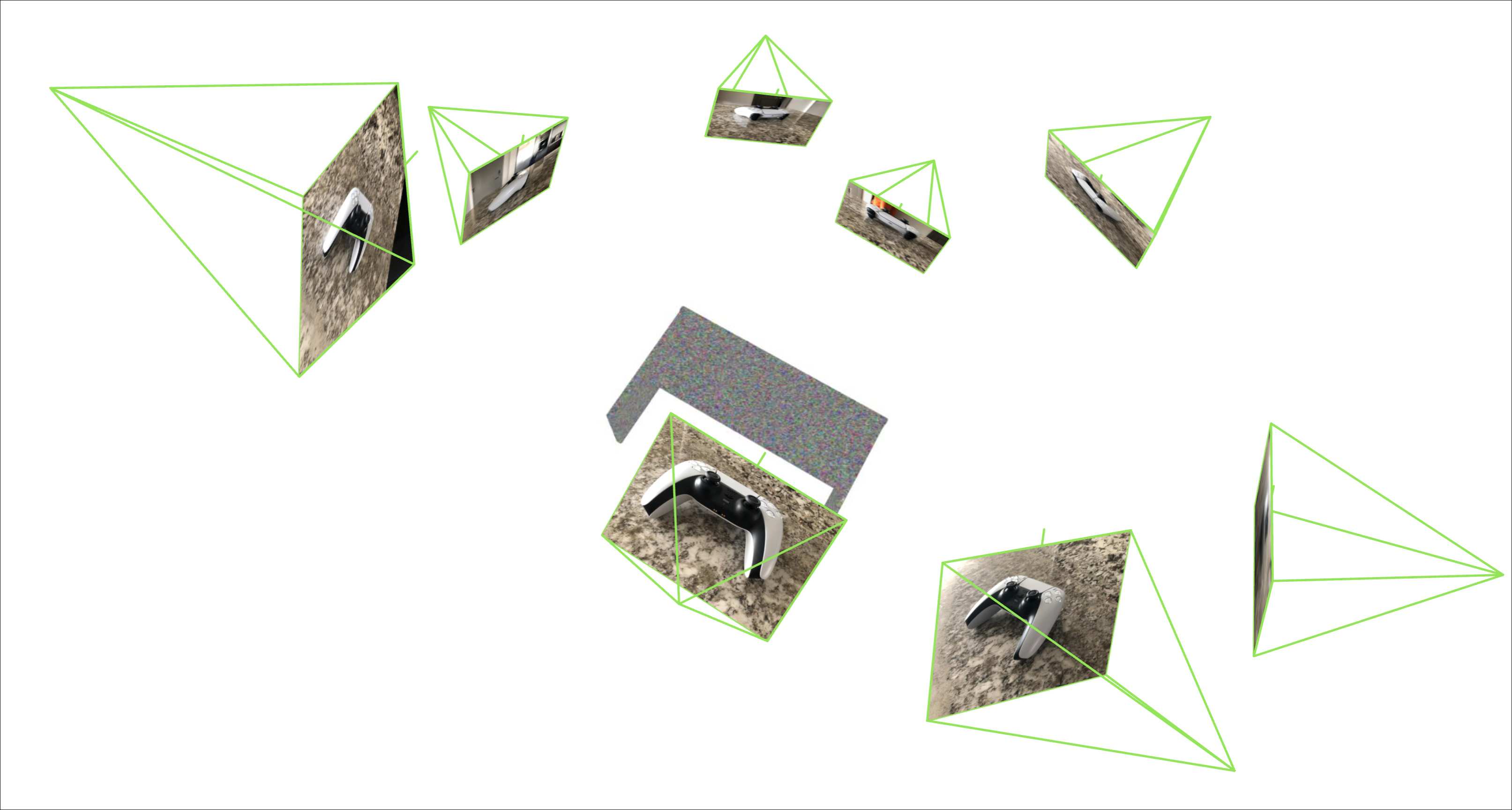}};
    \node[desc] at (img0.north) {Random colours \\ Poor initial poses \\ Mismatched CAD shape};

    \node[image node] (img3) at (\xA, 1.3) {\includegraphics[width=3.2cm]{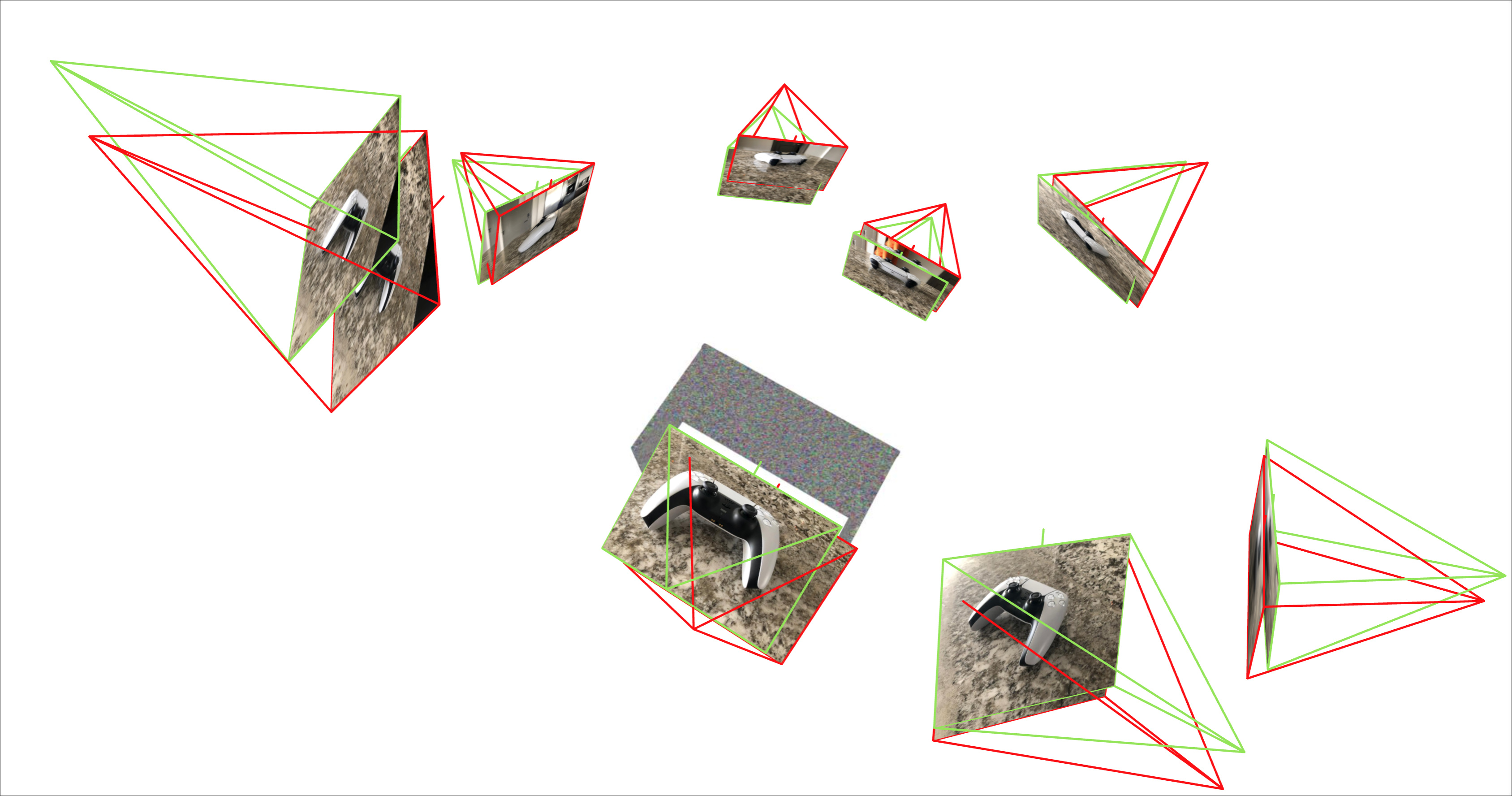}};
    \node[desc] at (img3.north) {Camera poses shifted \\ Image/CAD fit remains poor};

    \node[image node] (img6) at (\xB, 1.3) {\includegraphics[width=3.2cm]{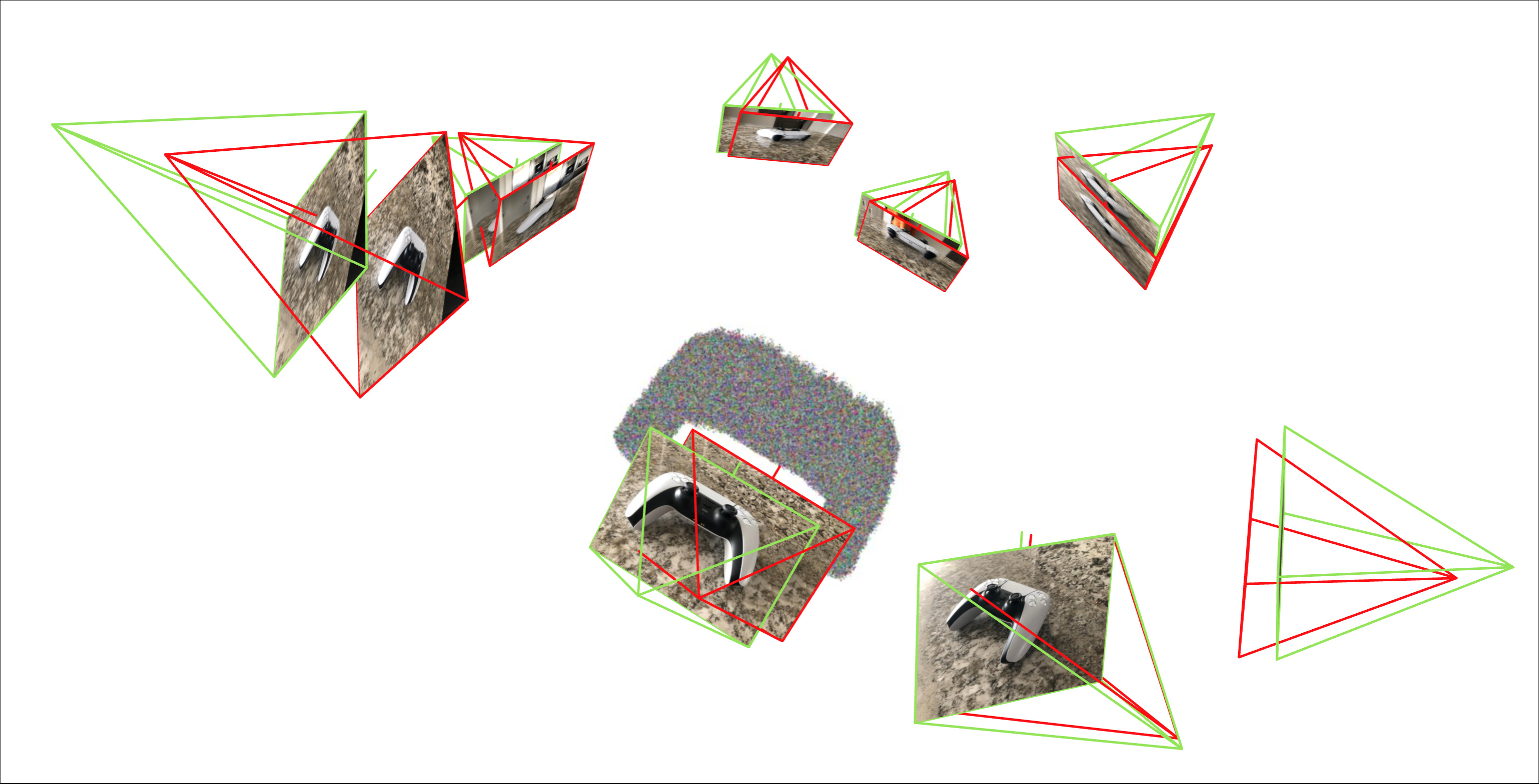}};
    \node[desc] at (img6.north) {Deformation aligns masks \\ Geometry matches physical object};

    \node[image node] (img30) at (\xC, 1.3) {\includegraphics[width=3.2cm]{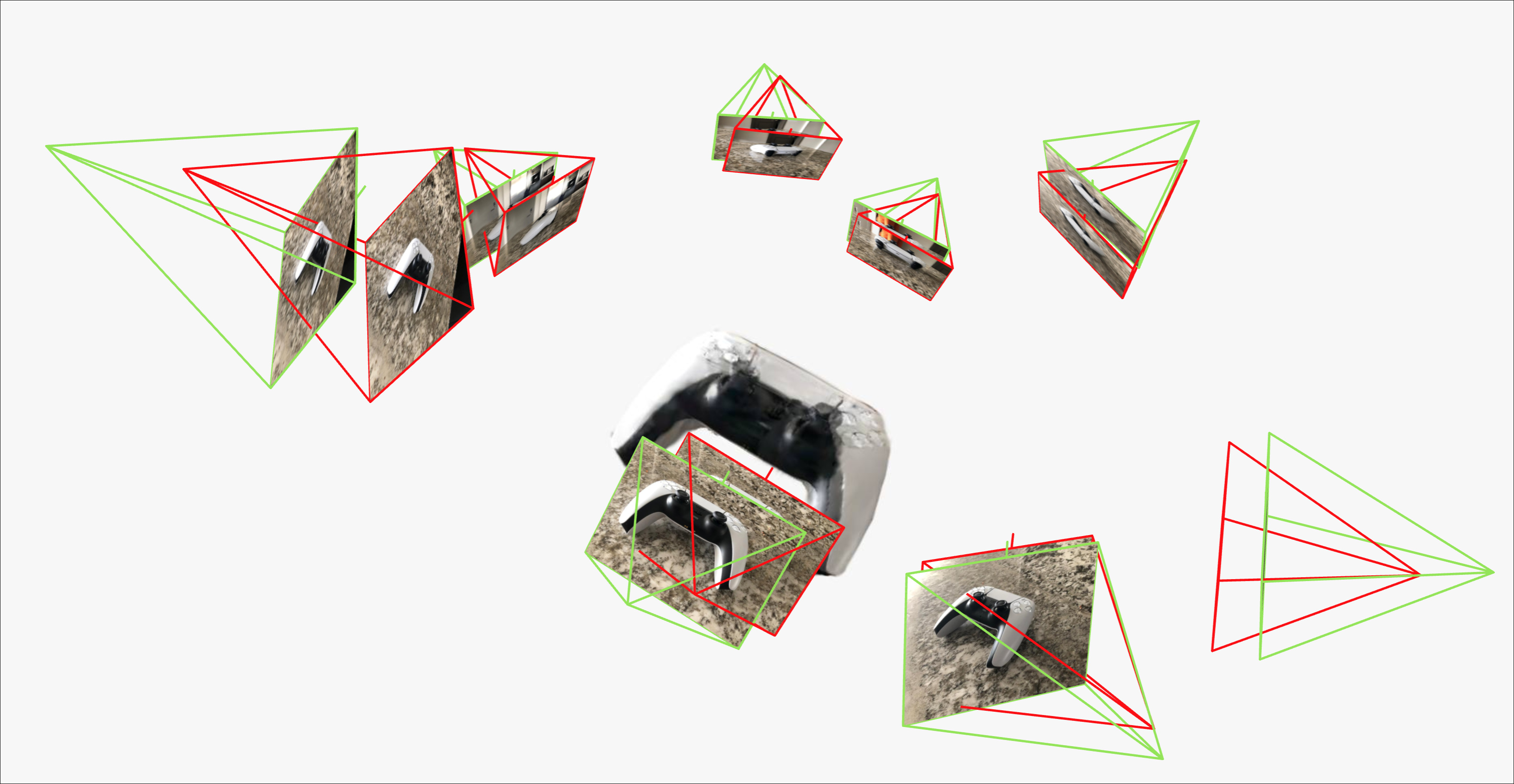}};
    \node[desc] at (img30.north) {3DGS properties learned \\ Photorealistic novel views};

    \draw[loosely dotted, very thick, gray] (img0.east) -- (img3.west);
    \draw[loosely dotted, very thick, gray] (img3.east) -- (img6.west);
    \draw[loosely dotted, very thick, gray] (img6.east) -- (img30.west);

    \draw[thick, gray!60] (\xO-0.5, 0) -- (\xC+0.5, 0);

    \node[timeline mark] (t0) at (\xO, 0) {};
    \node[below=4pt, font=\small\bfseries] at (t0) {Iter 0};

    \node[timeline mark] (t3) at (\xA, 0) {};
    \node[below=4pt, font=\small\bfseries] at (t3) {Iter $3{,}000$};

    \node[timeline mark] (t6) at (\xB, 0) {};
    \node[below=4pt, font=\small\bfseries] at (t6) {Iter $6{,}000$};

    \node[timeline mark] (t30) at (\xC, 0) {};
    \node[below=4pt, font=\small\bfseries] at (t30) {Iter $30{,}000$};

    \def\yPose{-1.2}
    \def\yDef{-1.8}
    \def\yPhoto{-2.4}

    \draw[phase bar, draw=cvprblue!80] (\xO, \yPose) -- (\xC, \yPose) node[midway, font=\footnotesize\bfseries, white] {Camera Pose Optimization};
    
    \draw[phase bar, draw=orange!90!black] (\xA, \yDef) -- (\xC, \yDef) node[midway, font=\footnotesize\bfseries, white] {Deformation Optimization};
    
    \draw[phase bar, draw=green!60!black] (\xB, \yPhoto) -- (\xC, \yPhoto) node[midway, font=\footnotesize\bfseries, white] {Photometric Finetuning};

    \draw[dashed, gray!50] (t0) -- (\xO, \yPose+0.1);
    \draw[dashed, gray!50] (t3) -- (\xA, \yDef+0.1);
    \draw[dashed, gray!50] (t6) -- (\xB, \yPhoto+0.1);
    \draw[dashed, gray!50] (t30) -- (\xC, \yPhoto-0.1);

    \end{tikzpicture}%
    }%
    \vspace{-5pt}
    \caption{\textbf{Evolution of the CADSplat Optimization process.} (Top) The visual state of the 3DGS representation at key iteration milestones. (Bottom) The active optimization parameters over the course of training.}
    \label{fig:opt_stages}
\end{figure*}

Attempting to jointly learn pose, deformation, and color from a noisy initialization can lead to local minima in early stages. We therefore decouple the geometric and photometric objectives into a \textit{Geometric Warmup} phase, which itself is split into a pose-only sub-phase and a pose-plus-deformation sub-phase, followed by a \textit{Photometric Finetuning} phase (\cref{fig:opt_stages}). Throughout training, the per-splat means, and the normal-aligned rotations $q_i$ are frozen, so that the deformation field absorbs all geometric corrections and the primitives remain surface-tangent-oriented disks. 

\subsubsection{Stage I: Geometric Warmup via Mask Alignment}
In Stage I, the objective is purely geometric: aligning camera poses and shape to the silhouettes. We render the splats with the 3DGS rasterizer and compare the alpha channel $A$ of the rendered RGBA image with the object masks $M_{GT}$ from \cref{sec:retrieval} using an $L_1$ loss:
\begin{equation}
    \mathcal{L}_{mask} = \| A - M_{GT} \|_1.
\label{eq:mask}
\end{equation}
For the first $3{,}000$ iterations, only the camera-pose delta is optimized; the deformation network is held at its zero-init identity, so it does not start absorbing the pose error. Between $3{,}000$ and $6{,}000$ iterations, we additionally enable the deformation network, which then begins ``wrapping'' the CAD silhouette onto the observed mask while the pose continues to be refined. 

\subsubsection{Stage II: Photometric Finetuning}
After $6{,}000$ total Stage-I iterations, Stage II enables the photometric optimization of the 3DGS primitives. We optimize the view-dependent color (Spherical Harmonics), per-splat scales, and opacities to capture the object photorealistically, while jointly updating pose and deformation using $\mathcal{L}_{mask}$. To isolate the object and prevent memorization of environmental artifacts, the RGB image is rendered with a random background and compared with the ground-truth image, with the background region replaced by the same random color. The resulting masked photometric loss combines $L_1$ and D-SSIM on the RGB channels:
\begin{equation}
    \mathcal{L}_{rgb} = (1 - \lambda) \mathcal{L}_1 + \lambda \mathcal{L}_{D-SSIM}.
\end{equation}

\subsubsection{Sparse-View Appearance Regularizers}
\label{sec:reg}
Even with the geometry pinned to the CAD manifold, the photometric loss alone leaves the per-splat appearance severely underdetermined in the sparse-view regime. We therefore add two complementary regularizers that act on the appearance parameters during Stage II, yielding the total Stage-II objective
\begin{equation}
    \mathcal{L}_{photo} \;=\; \mathcal{L}_{rgb} \;+\; \lambda_{mask}\, \mathcal{L}_{mask} \;+\; \lambda_{SH}\, \mathcal{L}_{SH} .
\label{eq:stage2_loss}
\end{equation}

\textbf{SH neighbor smoothness:} At initialization, we precompute the indices of the $K{=}20$ nearest neighbors of every splat $i$ in the canonical CAD frame; we denote this set $\mathcal{N}(i)$. Throughout Stage II, we add an SH smoothness term that penalizes the squared difference between every splat's SH coefficient vector $c_i = (c_i^{(0)}, \dots, c_i^{(N)})$ and that of each of its precomputed neighbors:
\begin{equation}
    \mathcal{L}_{SH} \;=\; \frac{1}{N K} \sum_{i=1}^{N} \sum_{j \in \mathcal{N}(i)} \| c_i - c_j \|_2^2 .
\label{eq:sh_smooth}
\end{equation}
Because the neighbor set is fixed in canonical space, the loss is invariant to the (small) deformations $\Delta\mu$ and acts as a graph Laplacian on appearance over the CAD surface.

\textbf{Stochastic splat dropout:} To further prevent any single splat from over-committing to a particular training view, at each Stage II iteration, we randomly subsample a fraction $p_{\text{drop}}$ of the splats uniformly without replacement and rasterize only the surviving subset. The mask-loss render in the same iteration uses the undropped set, so dropout regularizes appearance only and never destabilizes silhouette supervision. Because the surface is densely tessellated by $\sim$$\,$$600\mathrm{k}$ splats, the rendered silhouette and texture remain stable under random subsampling. In contrast, every splat is forced to share the explanation of each observation with its CAD-surface neighbors. We drop half the splats at the start of Stage II and reduce that fraction to zero over the first $20\mathrm{k}$ iterations using a cosine schedule, leaving the last $4\mathrm{k}$ iterations without any dropout. The regularization is thus strongest early, when the per-splat colors are least determined, and the final fit uses every splat.

\section{Experiments and Evaluation}
\label{sec:results}

We evaluate CADSplat on three object-centric datasets to demonstrate its broad applicability: \textbf{SCO-CAD}, our own dataset of sparse captures of common physical objects paired with approximate CAD models, and the two publicly available datasets of NeRS \cite{zhang2021ners}: \textbf{NeRS MVMC} and \textbf{NeRS Misc}. All datasets are annotated with ground-truth segmentation masks. 

\subsection{Implementation Details}
\label{sec:implementation_details}
We implement CADSplat on top of the \texttt{gsplat} 3DGS framework. We initialize $N_{splats}{=}600{,}000$ Gaussian primitives on the canonicalized CAD mesh surface, with their initial $\log$-scales set so each splat covers the mean distance to its three nearest neighbors, their quaternions set to the shortest-arc rotation aligned with the surface normal, and their opacities initialized to the post-sigmoid activation $0.5$.

As we lock Gaussian mean optimization and the per-splat quaternions entirely, geometric corrections are absorbed by the coordinate-based deformation field $\mathcal{D}_\theta$, optimized with Adam at a learning rate of $1.6 \times 10^{-5}$; the global pose correction, initialized from the silhouette-based consensus registration, is optimized with Adam at a learning rate of $5 \times 10^{-4}$. The optimization pipeline trains for $30{,}000$ iterations: an initial $3{,}000$-iteration pose-only sub-phase, a subsequent $3{,}000$-iteration pose-plus-deformation sub-phase, and a final $24{,}000$-iteration photometric finetuning phase, during which pose, deformation, and the 3DGS parameters are optimized jointly. \Cref{tab:hyperparams} lists the complete set of hyperparameters; all optimizers are Adam.

\begin{table}[!htbp]
\centering
\caption{Hyperparameters used in all experiments. The splat count is fixed (no densification or pruning); the dropout fraction is cosine-annealed from its initial value to $0$ over the first $20$k Stage-II iterations.}
\label{tab:hyperparams}
\begin{tabular*}{\linewidth}{@{\extracolsep{\fill}}llr@{}}
\toprule
\textbf{Group} & \textbf{Parameter} & \textbf{Value} \\ \midrule
Representation & splat count $N_{splats}$ & $600{,}000$ \\
 & SH degree & $3$ \\
 & means / quaternions & frozen \\ \midrule
Retrieval & hypotheses per view & top $50$ \\
 & silhouette IoU threshold & $0.7$ \\ \midrule
Deformation & MLP depth $D$ / width $W$ & $8$ / $256$ \\
 & PE frequency bands & $10$ \\ \midrule
Losses & D-SSIM weight $\lambda$ & $0.2$ \\
 & $\lambda_{mask}$ / $\lambda_{SH}$ & $1.0$ / $1.0$ \\
 & SH neighbors $K$ & $20$ \\
 & initial dropout $p_{\text{drop}}$ & $0.5$ \\ \midrule
Learning rates & deformation $\mathcal{D}_\theta$ & $1.6{\times}10^{-5}$ \\
 & pose correction & $5{\times}10^{-4}$ \\
 & scales & $5{\times}10^{-3}$ \\
 & opacities & $5{\times}10^{-2}$ \\
 & SH degree $0$ & $2.5{\times}10^{-3}$ \\
 & SH degrees $1$--$3$ & $1.25{\times}10^{-4}$ \\ \midrule
Schedule & Stage I: pose only & $3$k iters \\
 & Stage I: pose + deform & $3$k iters \\
 & Stage II: photometric & $24$k iters \\
 & test-time refinement & $400$ steps \\ \bottomrule
\end{tabular*}
\end{table}

\subsection{The SCO-CAD Dataset}
\label{sec:scocad}
\textbf{SCO-CAD}---Sparse Common Objects with associated CAD models---consists of 6 real-world objects. Authentic industrial CAD models are typically protected by strict corporate intellectual property (IP) rights. To avoid leaking sensitive, proprietary data, we deliberately constructed our dataset using common physical objects (e.g., a skateboard or scissors) for which approximate CAD models are publicly available under Creative Commons licenses. A direct consequence of this choice is that the associated CAD models often exhibit noticeable geometric deviations from the physical objects they capture. However, this intentional "reality gap" serves as an excellent stress test, necessitating and demonstrating the robustness of our non-rigid deformation network ($\mathcal{D}_\theta$). The six objects, along with their CAD models, will be released together with the full pipeline code. The CAD library for this dataset is the small set of these six approximate models, and every capture may pick any of them. We test large-scale retrieval on NeRS MVMC, where the library is the full ShapeNetCore \cite{chang2015shapenet} car category ($3{,}533$ models), and NeRS Misc using the other ShapeNetCore categories.

\textbf{Poses and calibration:}
We recover the relative calibration \emph{strictly from the sparse training images} using the hierarchical localization pipeline HLoc \cite{sarlin2019coarse} with learned features and matching \cite{sarlin2020superglue}, which is markedly more robust than traditional SIFT features \cite{lowe2004distinctive} in the wide-baseline sparse regime, and register this trajectory to the CAD canonical frame with the silhouette-based consensus registration of \cref{sec:retrieval}, refining a single global $\mathrm{Sim}(3)$ during training (\cref{sec:pose_opt}). At no point does the method observe poses or points derived from dense capture. This departs deliberately from the evaluation protocol common in sparse-view NVS \cite{zhu2024fsgs,li2024dngaussian}, in which the sparse training views are subsampled from a densely captured scene together with the camera poses (and often the SfM point cloud) computed from that dense capture---inputs that could not exist in a genuine sparse capture.

\textbf{Ground-truth views for evaluation:}
Each object is captured as a single dense video, from which we extract the $9$--$13$ training frames. The complete video is calibrated with COLMAP \cite{schoenberger2016sfm} and serves as ground truth for evaluation only; the held-out views used for evaluation are all frames of this trajectory except those used for training. The sparse HLoc calibration (training views) and the dense COLMAP calibration (training and held-out views) reside in different coordinate frames. We bridge them with a 7-DoF Umeyama alignment~\cite{umeyama1991least} fitted to the training-camera centers, which appear in both.

\subsection{Evaluation Protocol and Metrics}
\label{sec:eval_protocol}

Evaluating object-centric reconstructions from a genuinely sparse capture raises two difficulties that standard novel-view-synthesis benchmarking does not have: relative calibration discrepancies between training and testing images and a fair comparison between object-level algorithms such as ours and NeRS~\cite{zhang2021ners}, and scene-level algorithms such as FSGS~\cite{zhu2024fsgs} and InstantSplat~\cite{fan2024instantsplat}.

\begin{figure}[!htbp]
    \centering
    \setlength{\tabcolsep}{2pt}
    \begin{tabular}{ccc}
        \includegraphics[width=0.31\linewidth]{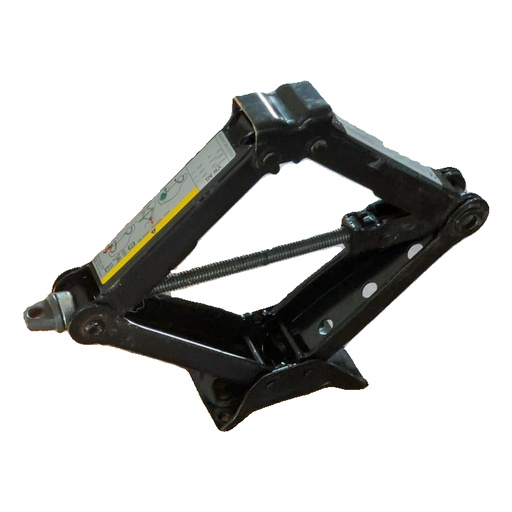} &
        \includegraphics[width=0.31\linewidth]{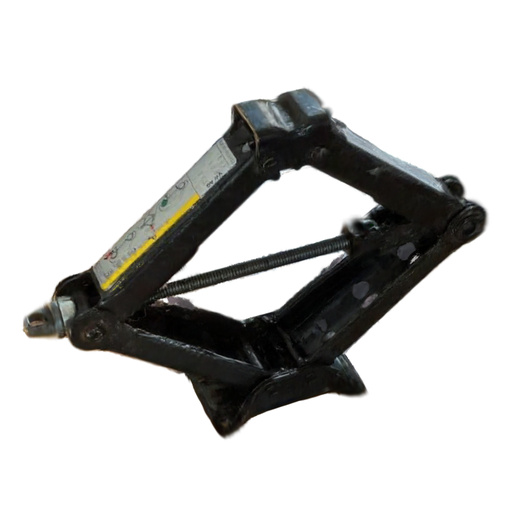} &
        \includegraphics[width=0.31\linewidth]{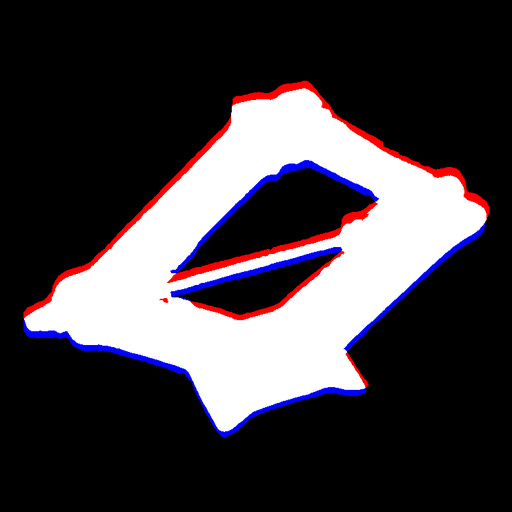} \\
        {\small Ground truth} & {\small Our render} & {\small Silhouette overlay} \\
    \end{tabular}
    \caption{Calibration residual on a held-out \emph{carjack} view. Training and held-out cameras come from two independent calibrations, so the rendered image (middle) is shifted by a few pixels relative to the photograph (left). The overlay (right) shows the shift: white where both silhouettes agree, red where only the photograph shows the object, and blue where only the render shows the object. Pixel metrics penalize these bands heavily, yet say nothing about reconstruction quality, which motivates test-time pose refinement and the eroded-mask metrics in \cref{sec:eval_protocol}.}
    \label{fig:eval_misalignment}
\end{figure}

If the training and testing relative calibration do not align perfectly, renders will be shifted by a few pixels relative to their ground-truth photograph. \Cref{fig:eval_misalignment} shows what this shift does at the object boundary: the ground-truth and rendered silhouettes disagree in a thin band, which is penalized heavily by pixel metrics, even when the reconstruction itself is accurate. This changes the question the evaluation asks. Standard novel-view synthesis asks whether a method reproduces the test photograph exactly at its given camera pose. With two independent calibrations, that camera is itself uncertain by a few pixels. We therefore reframe the evaluation to ask whether the object renders accurately from a camera \emph{near} its nominal pose---in other words, whether the reconstructed object is right. Following NeRS \cite{zhang2021ners}, we refine each held-out camera individually for $400$ optimization steps against its own target view with the reconstruction frozen: only the per-camera pose delta is learnable, so the reconstruction cannot be fitted to the held-out views. All metrics are reported \emph{with} refinement (\cref{tab:eval_avg}) and \emph{without} it (\cref{tab:eval_noref}); the difference reflects the portion of each score attributable to pose alignment rather than reconstruction quality.

The second difficulty is that two of our baselines, FSGS~\cite{zhu2024fsgs} and InstantSplat~\cite{fan2024instantsplat}, reconstruct the full scene rather than the object. Neither can be turned into an object-centric method: FSGS's semantic and depth priors need background context, and InstantSplat's dense stereo model reconstructs whatever is in view. There are different ways object-level renders can be compared against scene-level renders, each differing in which pixels from the renders and the ground truth are compared.

\textbf{Object-crop image quality (PSNR, SSIM, LPIPS, \flip{}):}
This is the evaluation protocol of NeRS \cite{zhang2021ners}, and the metrics in which we later report the NeRS datasets (\cref{tab:ners_mvmc_results}). The ground-truth image is composited onto white using the ground-truth mask, the prediction is rendered as-is over a white background, and both are cropped to the same $10\%$-padded square around the ground-truth mask and Lanczos-resized to $256{\times}256$. We report PSNR, SSIM \cite{wang2004image}, and two perceptual metrics: LPIPS \cite{zhang2018unreasonable} (AlexNet backbone), and \flip{} \cite{andersson2020flip}. The advantage of these metrics is that, because the prediction is \emph{not} masked, any geometry outside the true silhouette is penalized. However, it cannot report performance on full-scene methods as the background would be penalized. We mark such cells as not applicable.

\textbf{Eroded-mask image quality (mPSNR, mSSIM, mLPIPS, m\flip{}, fgPSNR):} Here \emph{both} the ground truth and the prediction are composited onto white using the ground-truth mask before cropping, as used by PixelNeRF \cite{yu2021pixelnerf} and RegNeRF \cite{niemeyer2022regnerf} for masked evaluations. We erode the mask by $5$ pixels to avoid unreliable segmentation at the silhouette edges. These metrics measure the correctness of the reconstruction inside the ground-truth mask. mPSNR, mSSIM, mLPIPS, and m\flip{} are computed on the full image. These metrics still favor scene-level reconstructions as their misaligned band contains background pixels rather than pure white. As a result, we also report fgPSNR as used in CO3D~\cite{reizenstein2021common}, which averages the squared error over the intersection of the rendered and ground-truth masks at full-image resolution, scoring only the foreground.

\textbf{Masking scene-level renders:} For FSGS~\cite{zhu2024fsgs} we additionally segment the foreground object in its renders with SAM3~\cite{carion2025sam3segmentconcepts} (manually verified) and use that mask in place of its opacity, so that it is scored on its own silhouette like every object-level method. This is reported as \textbf{FSGS+SAM3}. Because these masks are segmented from the pose-refined renders, this variant is only reported with test-time refinement.

\textbf{Silhouette accuracy:} Intersection-over-union (IoU) between the predicted mask (pixel opacity ${>}0.5$) and the \emph{un-eroded} ground-truth mask. This quantifies the misalignment between renders and ground truth.

\subsection{Baselines}
\label{sec:baselines}
We compare against two object-level and two scene-level methods. \textbf{Vanilla 3DGS} \cite{kerbl20233d} is trained on the same mask-composited images as our method, and \textbf{MVS-Texturing} \cite{Waechter2014Texturing} projects the images as textures onto the CAD mesh. \textbf{FSGS} \cite{zhu2024fsgs} and \textbf{InstantSplat} \cite{fan2024instantsplat} reconstruct the whole scene. FSGS, vanilla 3DGS, and MVS-Texturing use the same sparse HLoc calibration as our method; InstantSplat estimates its own poses from the training views. Among sparse-view methods with depth and semantic priors, we chose FSGS over DNGaussian \cite{li2024dngaussian} because it is newer and uses a superset of DNGaussian's priors.

Two related methods are absent from this comparison. CADNeRF~\cite{wen2025cad}, the closest CAD-based method, released no code. NeRS~\cite{zhang2021ners} assumes objects with smooth materials and a known category prior to operate well. We therefore compare against both on the NeRS MVMC and Misc datasets (\cref{sec:ners_mvmc}), using their published numbers and renders.

\subsection{Quantitative Results}
\label{sec:quantitative}

\begin{table*}[!tp]
\centering
\caption{Novel-view synthesis on held-out views of \textbf{SCO-CAD}: all metrics (defined in \cref{sec:eval_protocol}), averaged over the six scenes, \emph{with} test-time pose refinement. Some cells show ``--'' when the method renders full scenes and has no object silhouette to score for the metric. MVS-Texturing is reported without test-time refinement, which degrades it; its row therefore equals that of \cref{tab:eval_noref}.}
\label{tab:eval_avg}
\resizebox{\textwidth}{!}{%
\begin{tabular}{l|cccc|ccccc|c}
\toprule
& \multicolumn{4}{c|}{\textbf{Object crop}} & \multicolumn{5}{c|}{\textbf{Eroded mask}} & \textbf{Silhouette}\\
\textbf{Method} & \textbf{PSNR}\,$\uparrow$ & \textbf{SSIM}\,$\uparrow$ & \textbf{LPIPS}\,$\downarrow$ & \textbf{\flip{}}\,$\downarrow$ & \textbf{mPSNR}\,$\uparrow$ & \textbf{mSSIM}\,$\uparrow$ & \textbf{mLPIPS}\,$\downarrow$ & \textbf{m\flip{}}\,$\downarrow$ & \textbf{fgPSNR}\,$\uparrow$ & \textbf{IoU}\,$\uparrow$ \\ \midrule
FSGS \cite{zhu2024fsgs} & -- & -- & -- & -- & 27.183 & 0.930 & 0.046 & 0.057 & 18.814 & -- \\
FSGS+SAM3 & 20.250 & 0.861 & 0.125 & 0.101 & 24.836 & 0.916 & 0.082 & 0.064 & 16.398 & 0.821 \\
Vanilla 3DGS (masked) \cite{kerbl20233d} & 21.244 & 0.852 & 0.072 & 0.092 & 26.625 & 0.907 & 0.044 & 0.059 & 18.222 & 0.913 \\
MVS-Texturing \cite{Waechter2014Texturing} & 11.765 & 0.706 & 0.355 & 0.253 & 14.644 & 0.813 & 0.274 & 0.164 & 6.827 & 0.400 \\
InstantSplat \cite{fan2024instantsplat} & -- & -- & -- & -- & 27.898 & 0.930 & 0.045 & 0.062 & 19.552 & -- \\
\textbf{Ours (Full)} & \textbf{23.239} & \textbf{0.888} & \textbf{0.058} & \textbf{0.075} & \textbf{28.953} & \textbf{0.932} & \textbf{0.031} & \textbf{0.048} & \textbf{20.360} & \textbf{0.942} \\ \bottomrule
\end{tabular}%
}
\end{table*}

\begin{table*}[!tp]
\centering
\caption{Same as \cref{tab:eval_avg} but \emph{without} test-time pose refinement (Sim(3) alignment only). FSGS+SAM3 is omitted: its evaluation masks are segmented from the pose-refined renders (\cref{sec:eval_protocol}), and the unrefined renders would yield a different mask set, so no comparable row exists. The method ranking matches \cref{tab:eval_avg} on eight of ten metrics (mSSIM and m\flip{} shift by ${\leq}0.001$), so no conclusion depends on the refinement step. ``--'' as in \cref{tab:eval_avg}.}
\label{tab:eval_noref}
\resizebox{\textwidth}{!}{%
\begin{tabular}{l|cccc|ccccc|c}
\toprule
& \multicolumn{4}{c|}{\textbf{Object crop}} & \multicolumn{5}{c|}{\textbf{Eroded mask}} & \textbf{Silhouette} \\
\textbf{Method} & \textbf{PSNR}\,$\uparrow$ & \textbf{SSIM}\,$\uparrow$ & \textbf{LPIPS}\,$\downarrow$ & \textbf{\flip{}}\,$\downarrow$ & \textbf{mPSNR}\,$\uparrow$ & \textbf{mSSIM}\,$\uparrow$ & \textbf{mLPIPS}\,$\downarrow$ & \textbf{m\flip{}}\,$\downarrow$ & \textbf{fgPSNR}\,$\uparrow$ & \textbf{IoU}\,$\uparrow$ \\ \midrule
FSGS \cite{zhu2024fsgs} & -- & -- & -- & -- & 25.375 & \textbf{0.911} & 0.054 & 0.066 & 17.200 & -- \\
Vanilla 3DGS (masked) \cite{kerbl20233d} & 20.929 & 0.850 & 0.070 & 0.091 & 26.594 & 0.910 & 0.040 & \textbf{0.056} & 18.216 & 0.920 \\
MVS-Texturing \cite{Waechter2014Texturing} & 11.765 & 0.706 & 0.355 & 0.253 & 14.644 & 0.813 & 0.274 & 0.164 & 6.827 & 0.400 \\
InstantSplat \cite{fan2024instantsplat} & -- & -- & -- & -- & 23.818 & 0.892 & 0.061 & 0.075 & 15.877 & -- \\
\textbf{Ours (Full)} & \textbf{21.077} & \textbf{0.851} & \textbf{0.068} & \textbf{0.091} & \textbf{26.725} & 0.910 & \textbf{0.038} & 0.057 & \textbf{18.404} & \textbf{0.922} \\ \bottomrule
\end{tabular}%
}
\end{table*}

\Cref{tab:eval_avg} reports all results averaged over the six scenes \emph{with} test-time pose refinement, and \cref{tab:eval_noref} \emph{without} it. CADSplat achieves the best six-scene average across all metrics. Scored inside the ground-truth mask, FSGS is competitive on pixel metrics but consistently trails on perceptual metrics (mLPIPS $0.046$ vs.\ our $0.031$, with ours better on every individual scene; m\flip{} $0.057$ vs.\ our $0.048$). Once its renders are masked with its own silhouette for a proper object-to-object comparison (FSGS+SAM3), it drops across the board (mPSNR $27.18{\rightarrow}24.84$, fgPSNR $18.81{\rightarrow}16.40$, IoU $0.821$ vs.\ our $0.942$). Inspecting the qualitative renders, FSGS's over-smoothed textures are kind to PSNR but visibly worse, as captured by LPIPS and \flip{}. Attempting to train FSGS on mask-composited images, with its depth and pseudo-view losses restricted to the mask, confirms that it cannot reconstruct the object without scene context, with PSNR falling from $20.25$ to $13.43$. We therefore omit this variant from the tables.

\textbf{Vanilla 3DGS.} The vanilla 3DGS baseline---unconstrained masked 3DGS trained on the same background-masked images---trails CADSplat everywhere (mPSNR $28.95$ vs.\ $26.63$, fgPSNR $20.36$ vs.\ $18.22$, IoU $0.942$ vs.\ $0.913$, and lower LPIPS/mLPIPS/m\flip{}). The gap is largest on voluminous objects with significant occlusion (car1, carjack, senseo: $+2.7$ to $+3.1$\,dB mPSNR) and smallest on the flat scissors object ($+0.9$\,dB). The reason the gap between our method and Vanilla 3DGS is not larger is that we train on masked objects and randomize the background colors, which prevents floaters outside the object hull. Training 3DGS on the unmasked sparse training views produces noticeably more artifacts, but that would be an unfair comparison.

\textbf{MVS-Texturing.} Classical texturing of the CAD mesh is the weakest baseline in the comparison: a six-scene average PSNR of $11.77$, mPSNR $14.64$, fgPSNR $6.83$. The textures appear as a patchwork with visible seams; when the pose or CAD shape is inaccurate, faces fail to be textured properly, leading to holes, missing parts, and artifacts. Test-time refinement does not help, as it diverges away from the actual object. Therefore, we only report the metrics without test-time refinement.

\textbf{InstantSplat.} Geometry foundation models extend novel-view synthesis to the sparse-view regime by replacing classical SfM with learned dense stereo. InstantSplat \cite{fan2024instantsplat} uses MASt3R to predict camera poses and a dense point cloud from the sparse views, then optimizes a 3DGS scene from that initialization while jointly refining the poses. Despite being scored inside the ground-truth mask, it trails our method on every metric (mPSNR $27.9$ vs.\ $29.0$, mLPIPS $0.045$ vs.\ $0.031$, fgPSNR $19.6$ vs.\ $20.4$).

\begin{figure*}[!tp]
    \begin{center}
        \adjustbox{max width=\textwidth}{%
        \begin{tabular}{lc@{}c@{\hspace{10pt}}c@{\hspace{10pt}}c@{\hspace{10pt}}c@{\hspace{10pt}}c@{\hspace{10pt}}c @{}}
            GT & \qualresult{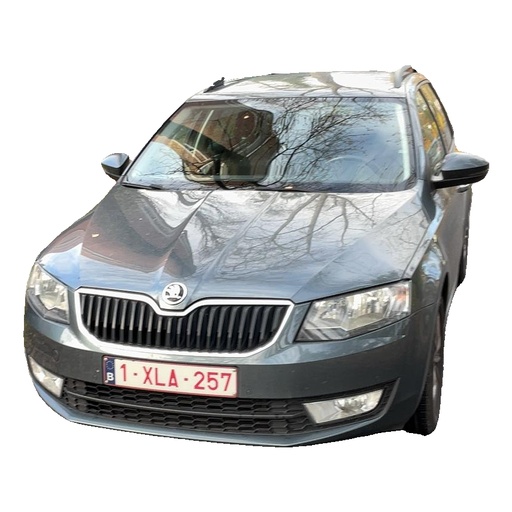} & \qualresult{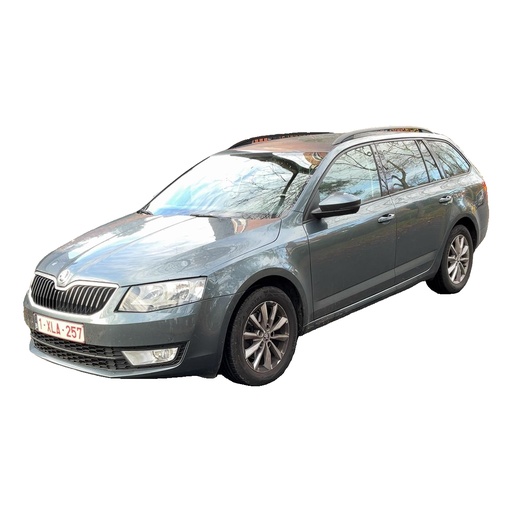} & \qualresult{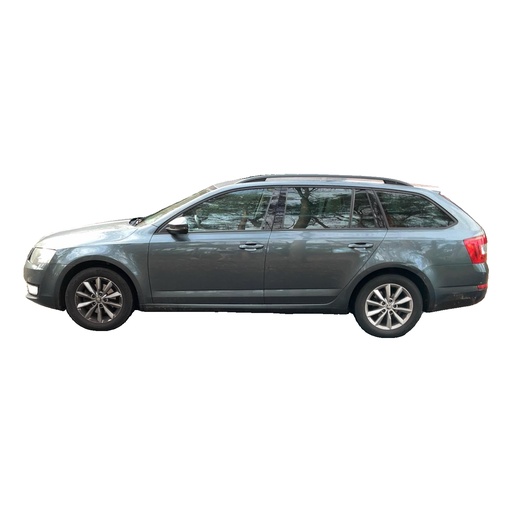} & \qualresult{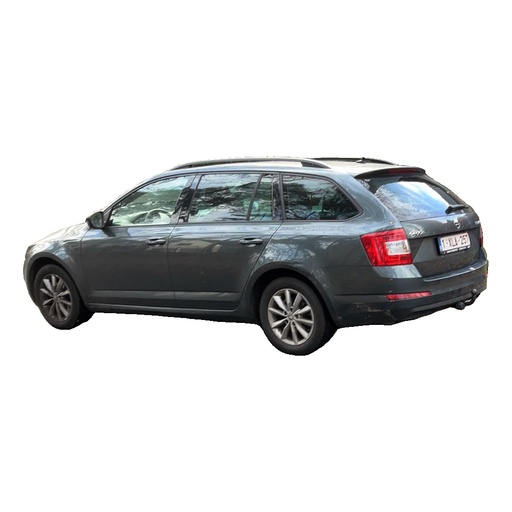} & \qualresult{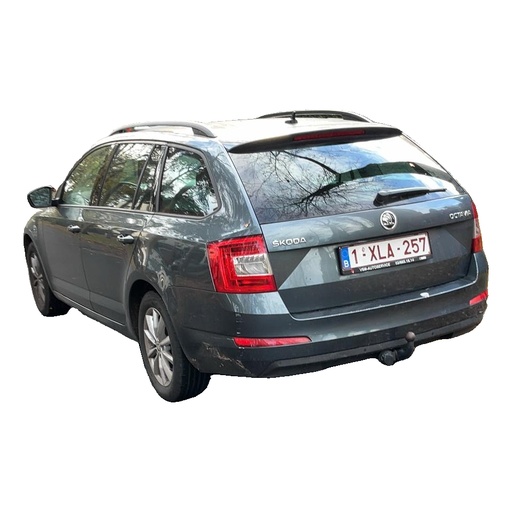} & \qualresult{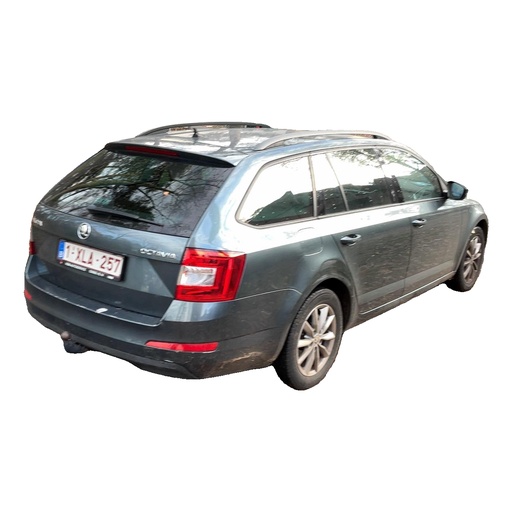} \\[3pt]
            \textbf{Ours} & \qualresult{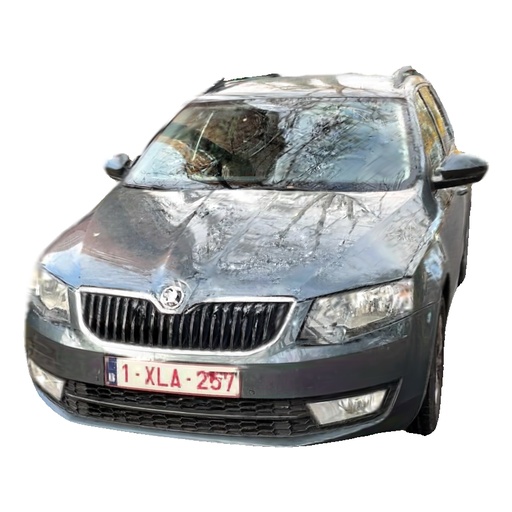} & \qualresult{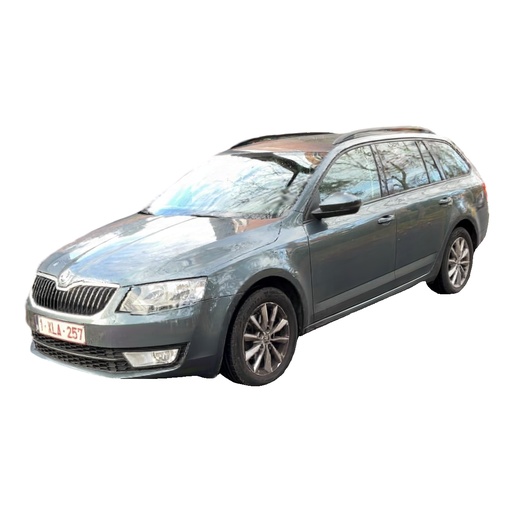} & \qualresult{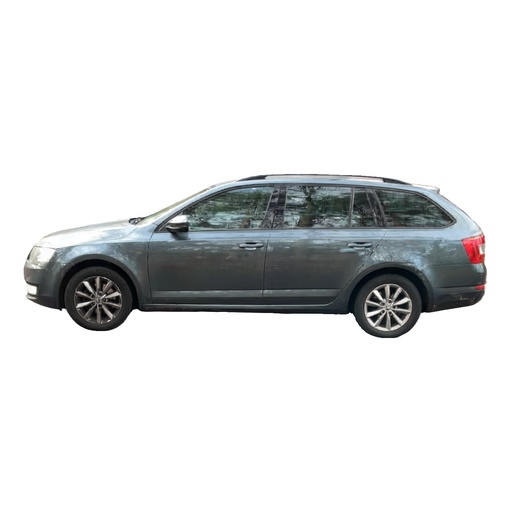} & \qualresult{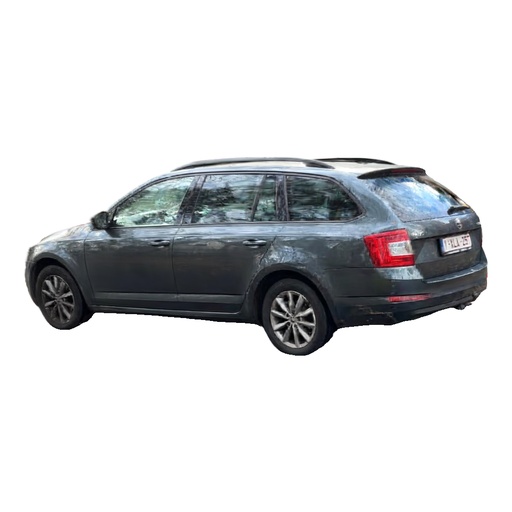} & \qualresult{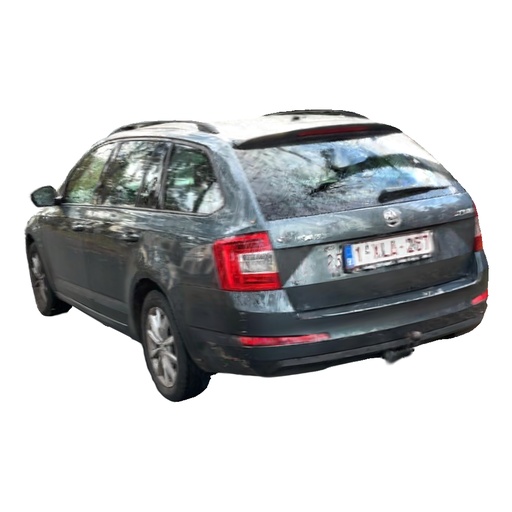} & \qualresult{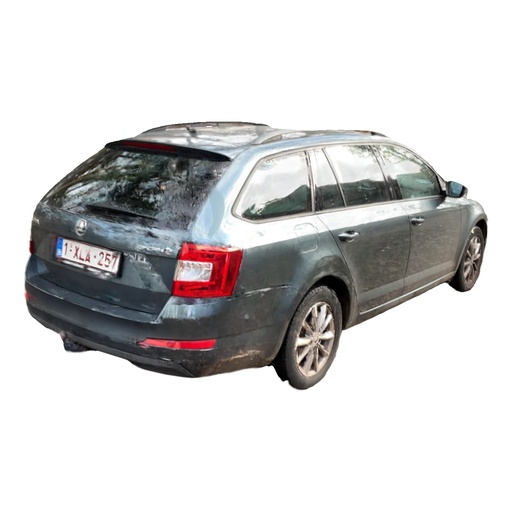} \\[3pt]
            FSGS~\cite{zhu2024fsgs} & \qualresult{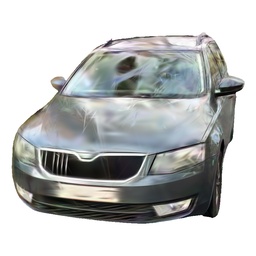} & \qualresult{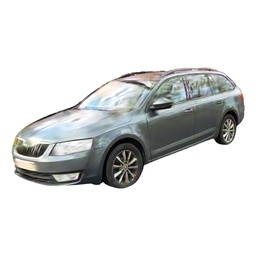} & \qualresult{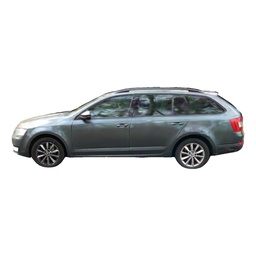} & \qualresult{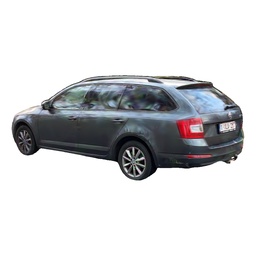} & \qualresult{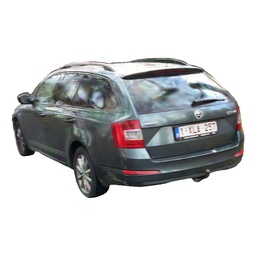} & \qualresult{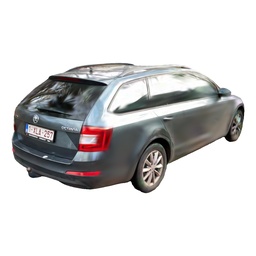} \\[3pt]
            InstantSplat~\cite{fan2024instantsplat} & \qualresult{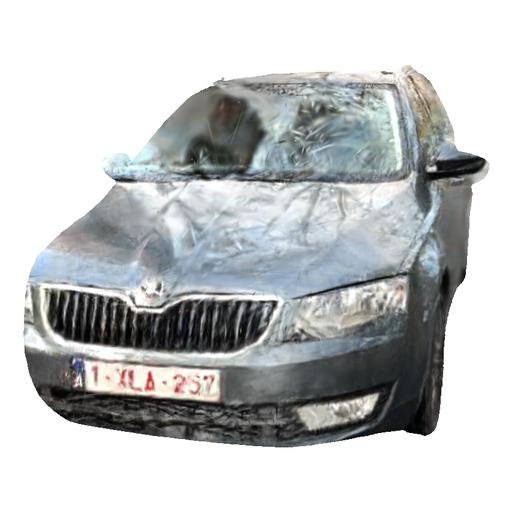} & \qualresult{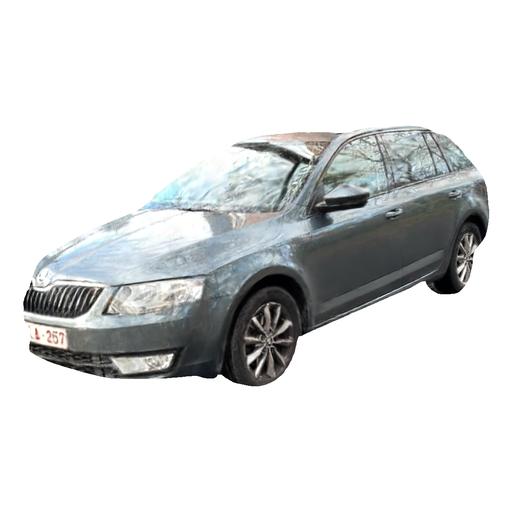} & \qualresult{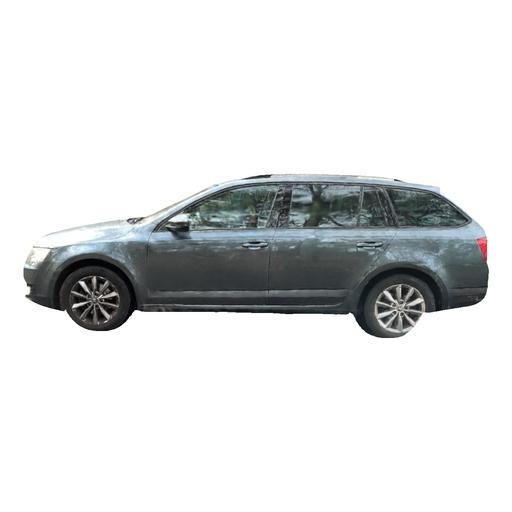} & \qualresult{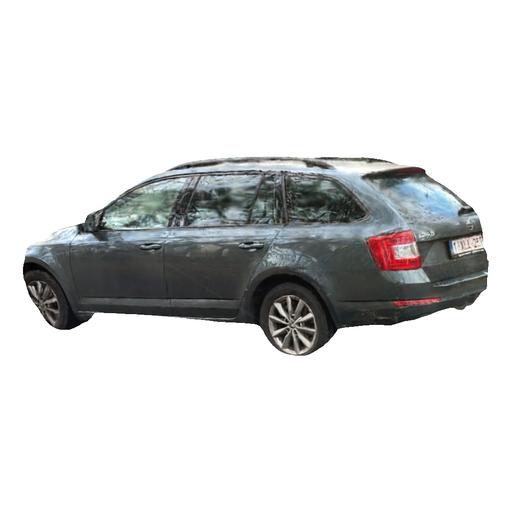} & \qualresult{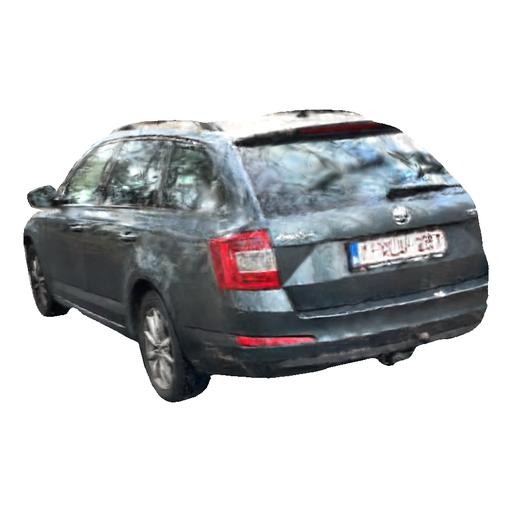} & \qualresult{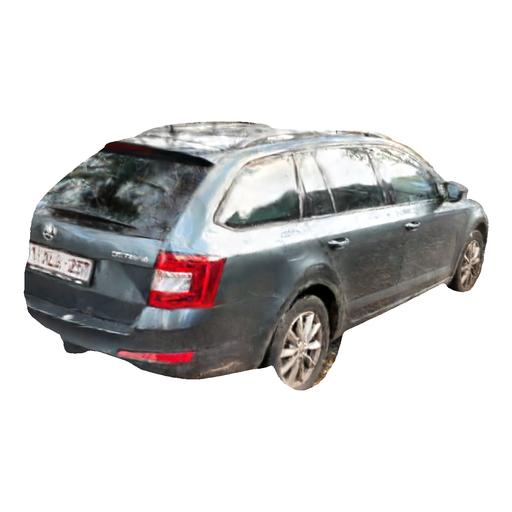} \\[3pt]
            MVS-Texturing~\cite{Waechter2014Texturing} & \qualresult{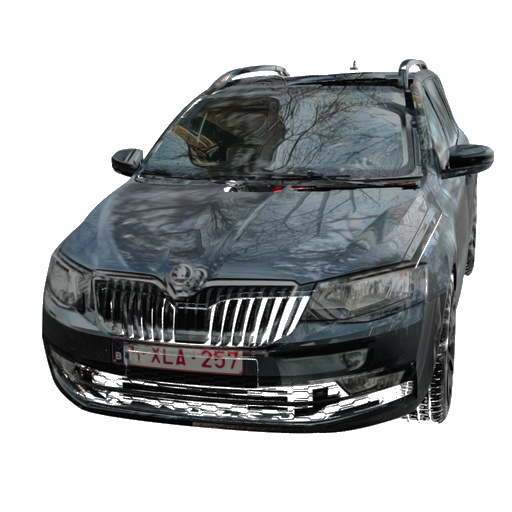} & \qualresult{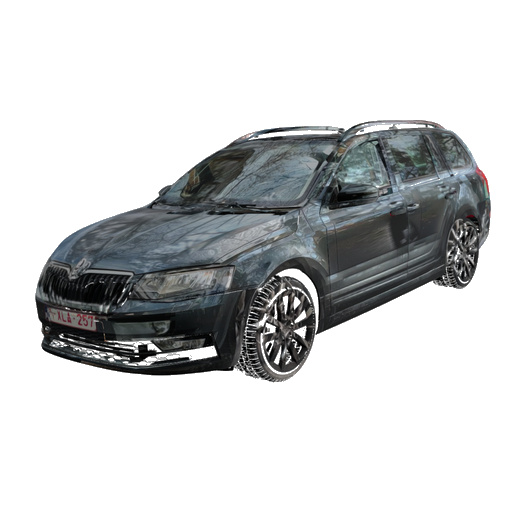} & \qualresult{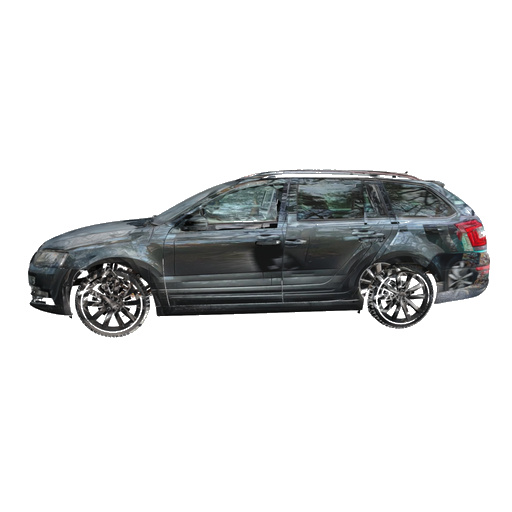} & \qualresult{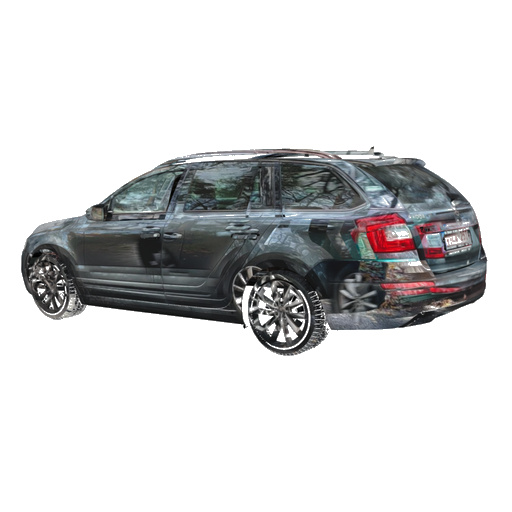} & \qualresult{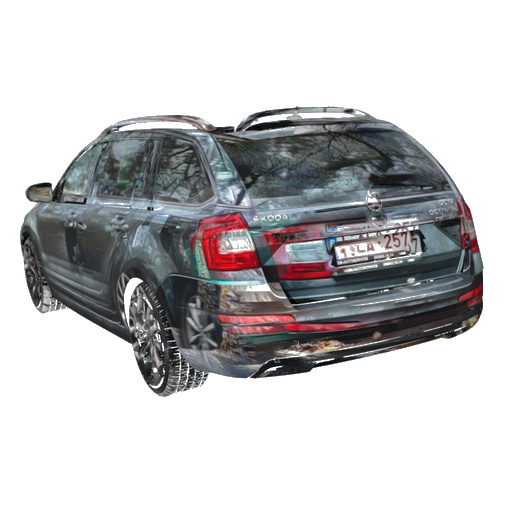} & \qualresult{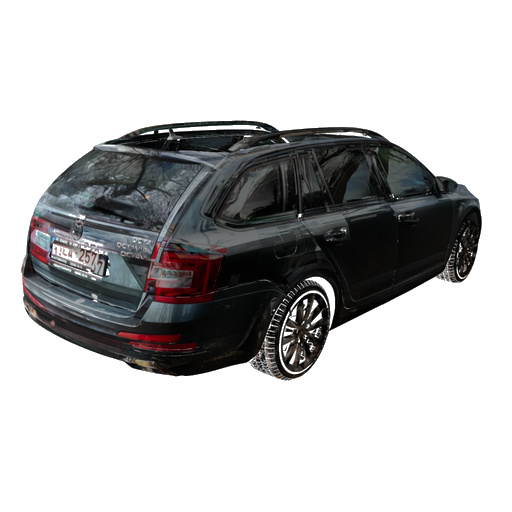} \\[14pt]
            
            GT & \qualresult{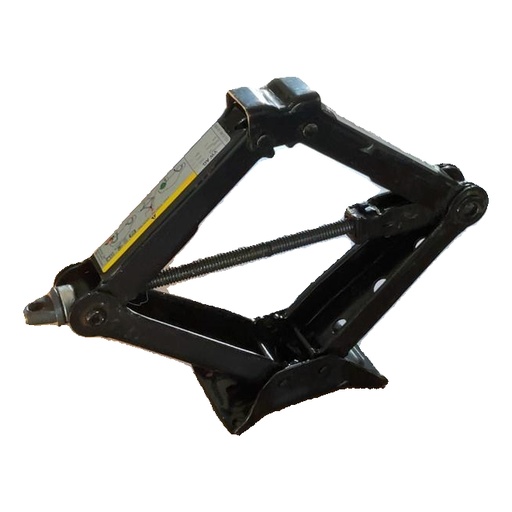} & \qualresult{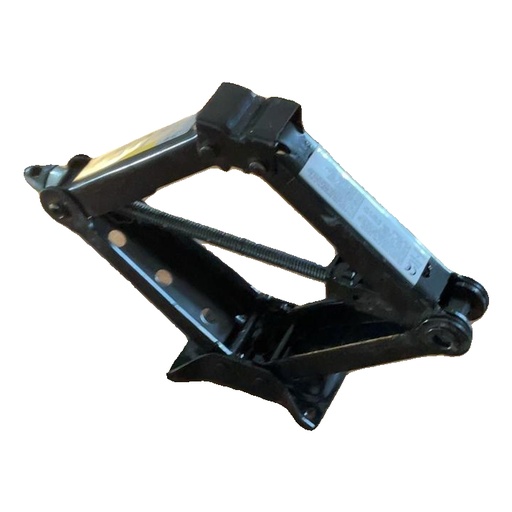} & \qualresult{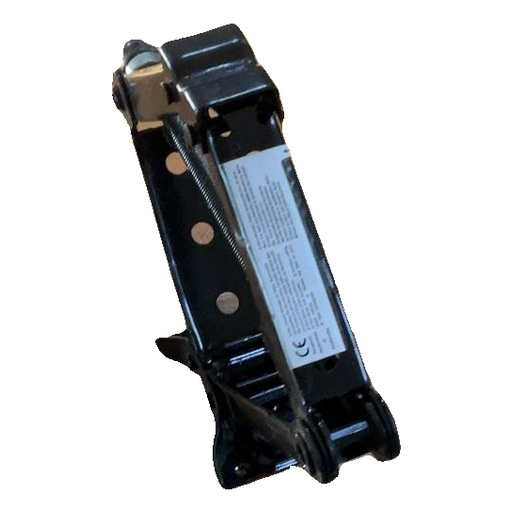} & \qualresult{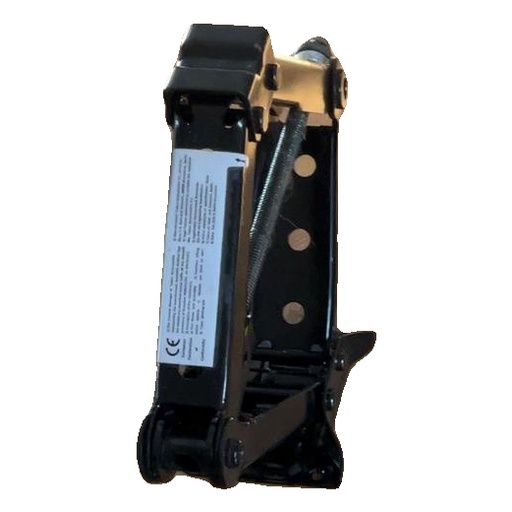} & \qualresult{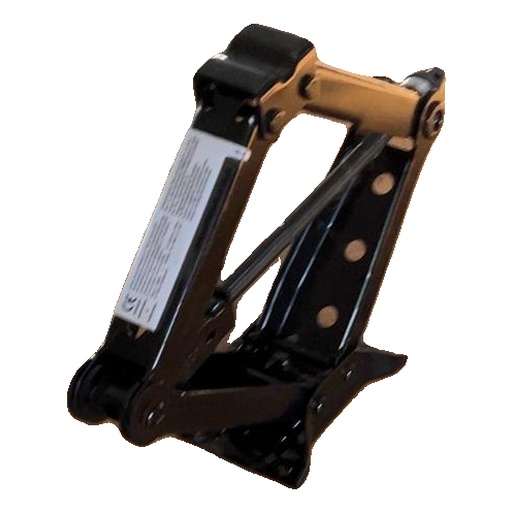} & \qualresult{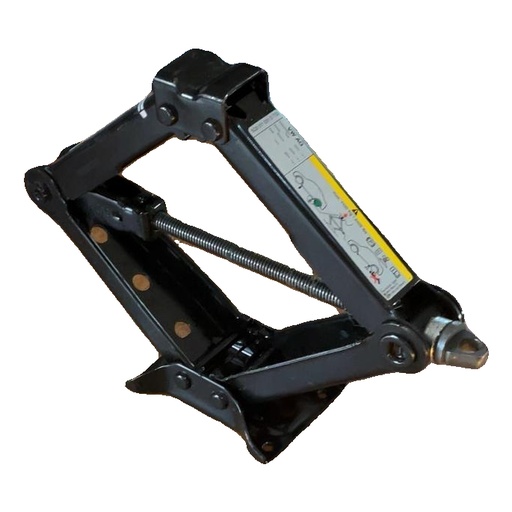} \\[3pt]
            \textbf{Ours} & \qualresult{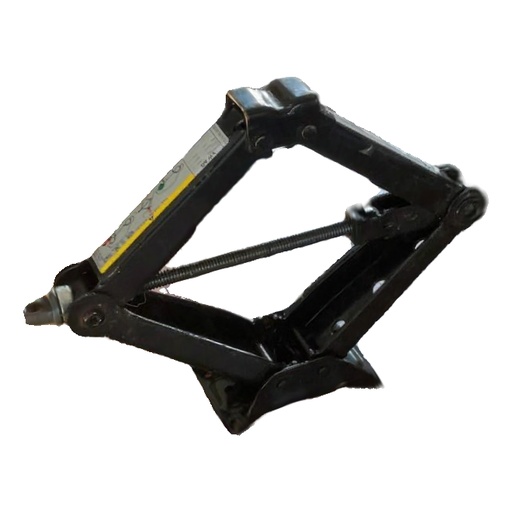} & \qualresult{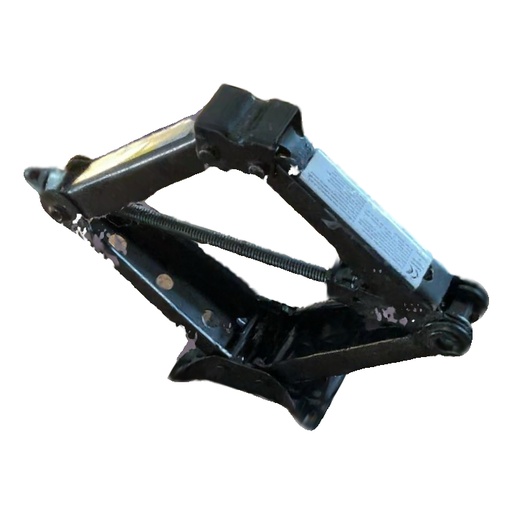} & \qualresult{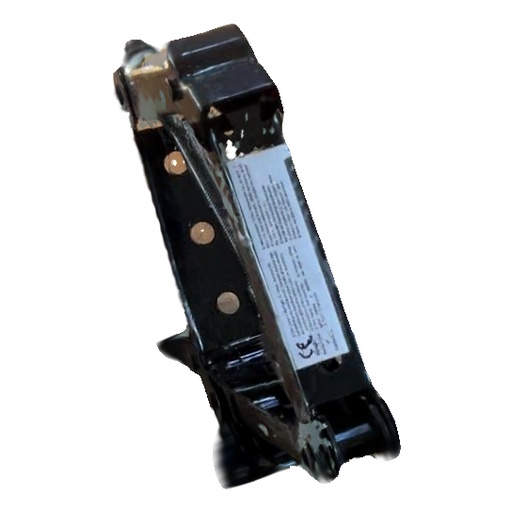} & \qualresult{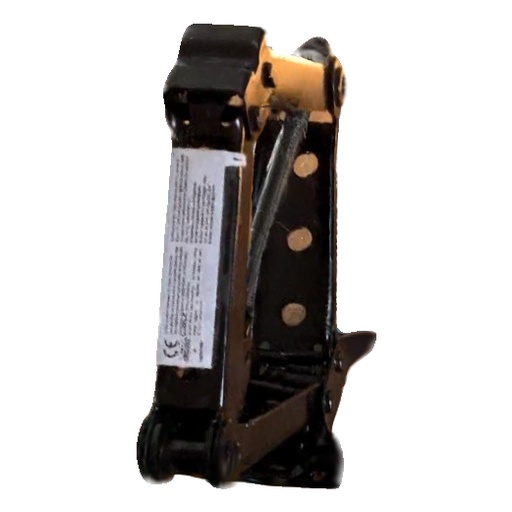} & \qualresult{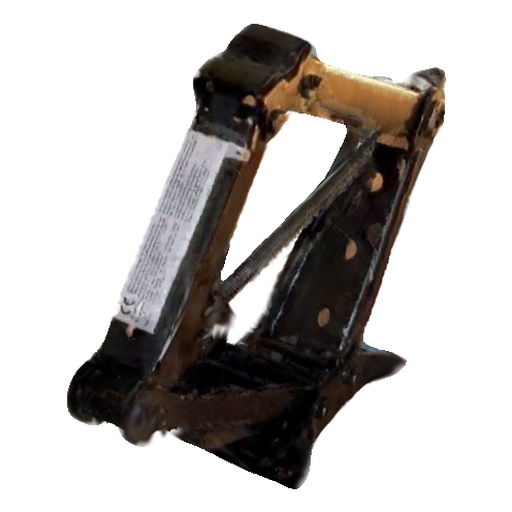} & \qualresult{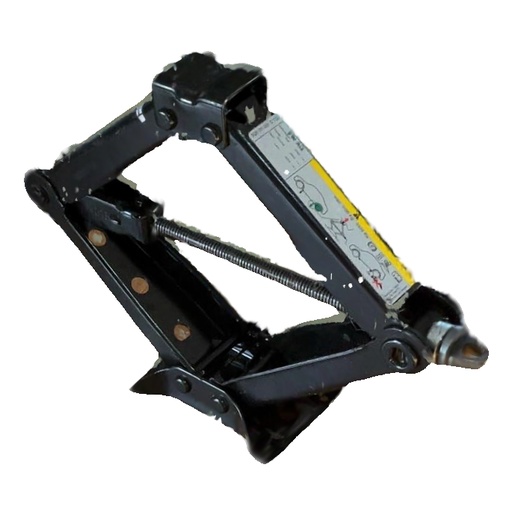} \\[3pt]
            FSGS~\cite{zhu2024fsgs} & \qualresult{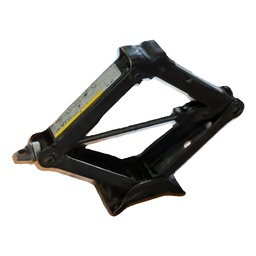} & \qualresult{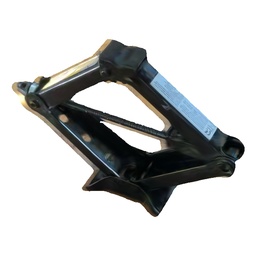} & \qualresult{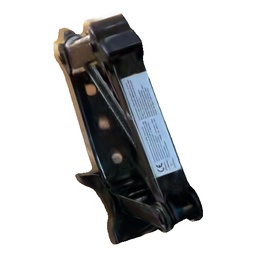} & \qualresult{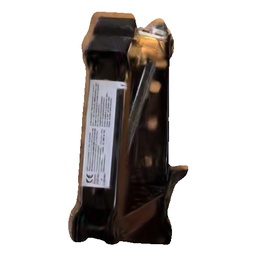} & \qualresult{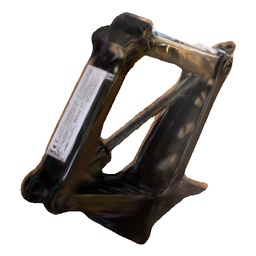} & \qualresult{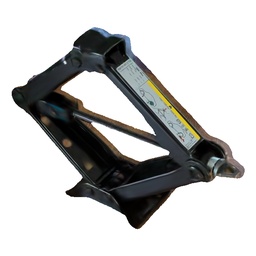} \\[3pt]
            InstantSplat~\cite{fan2024instantsplat} & \qualresult{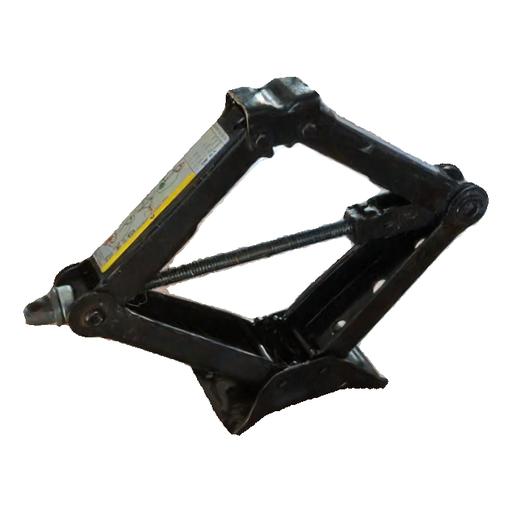} & \qualresult{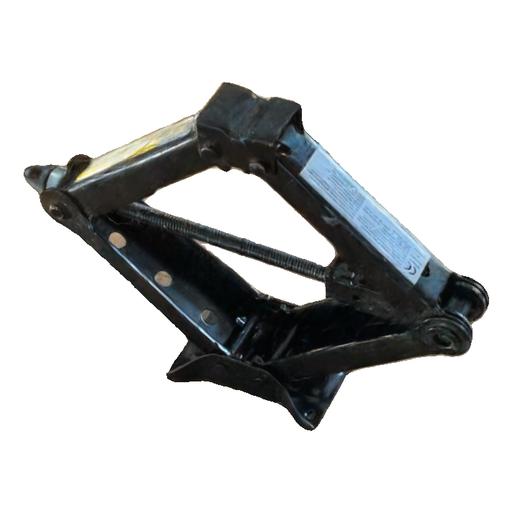} & \qualresult{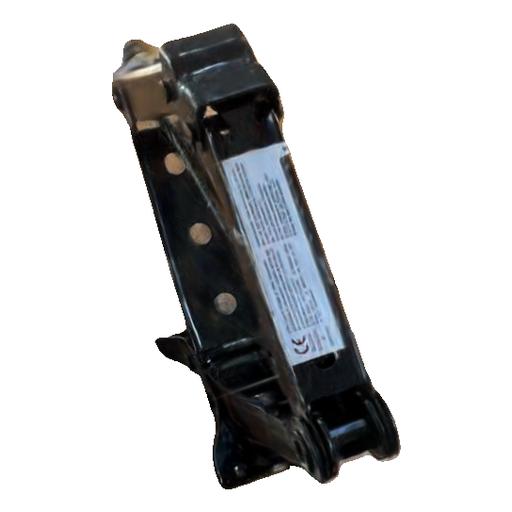} & \qualresult{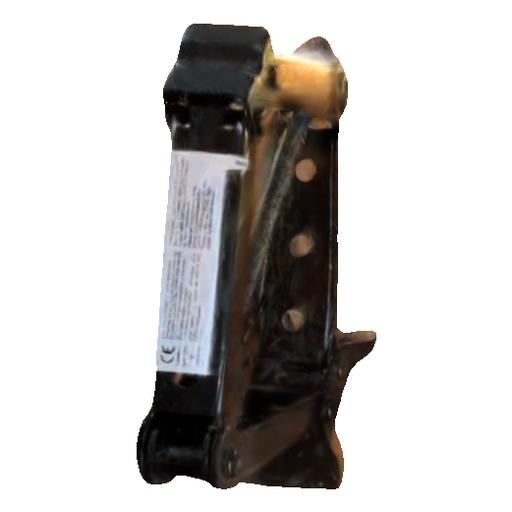} & \qualresult{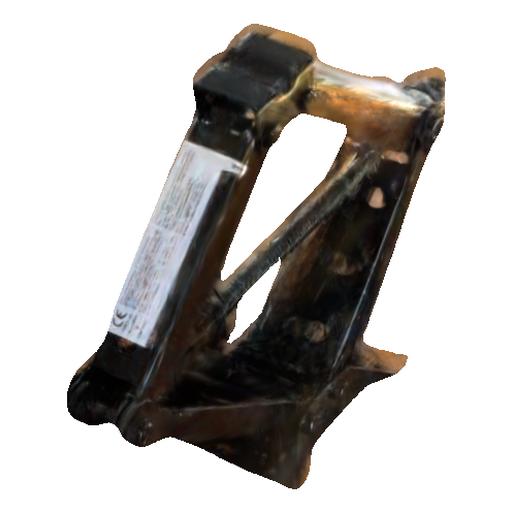} & \qualresult{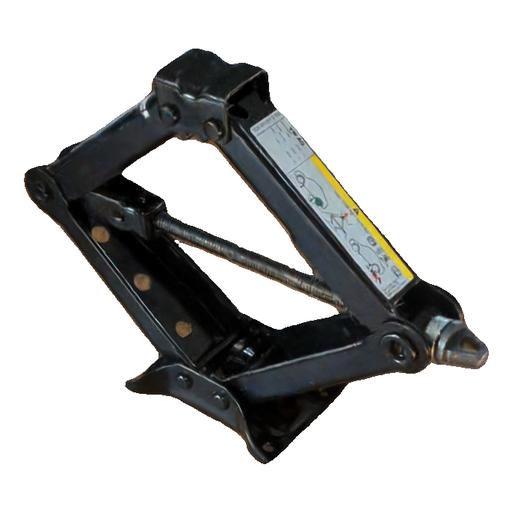} \\[3pt]
            MVS-Texturing~\cite{Waechter2014Texturing} & \qualresult{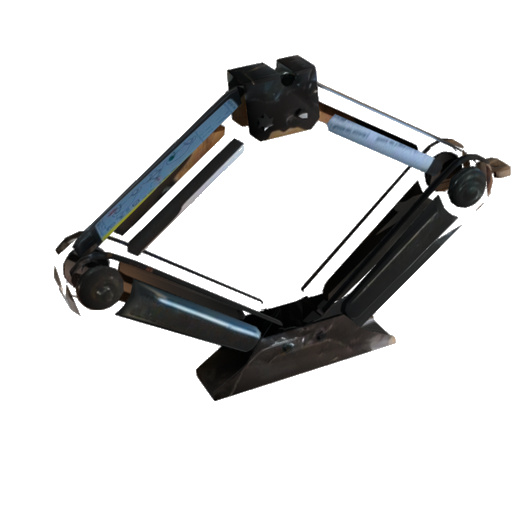} & \qualresult{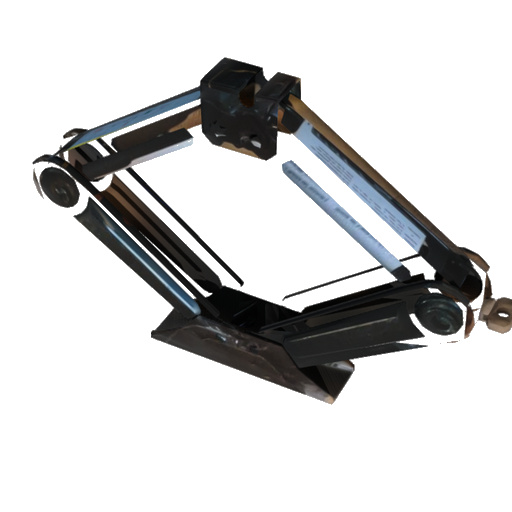} & \qualresult{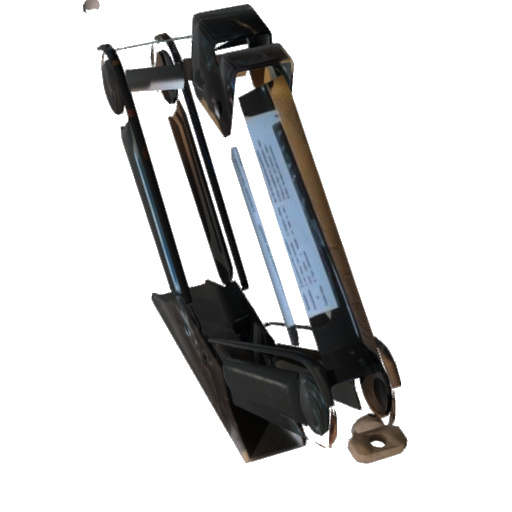} & \qualresult{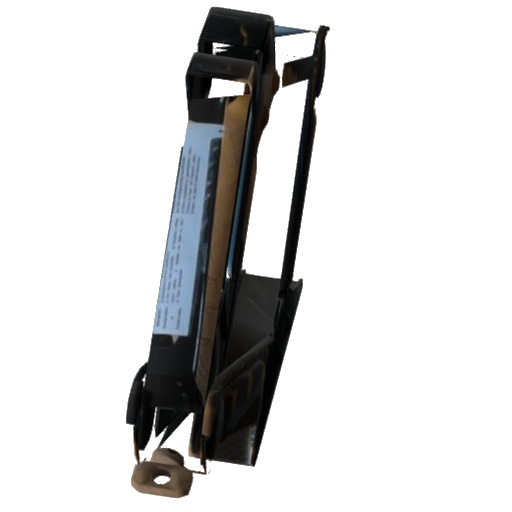} & \qualresult{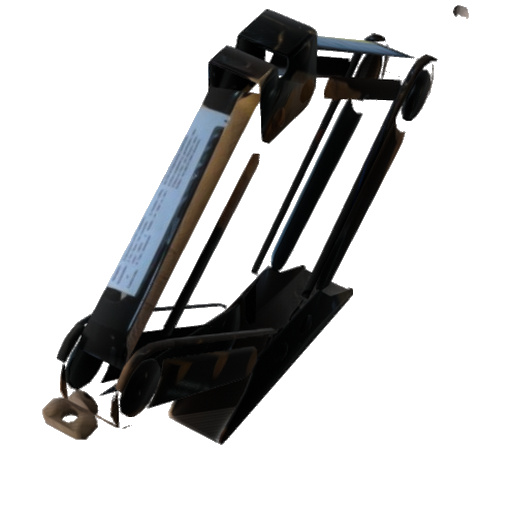} & \qualresult{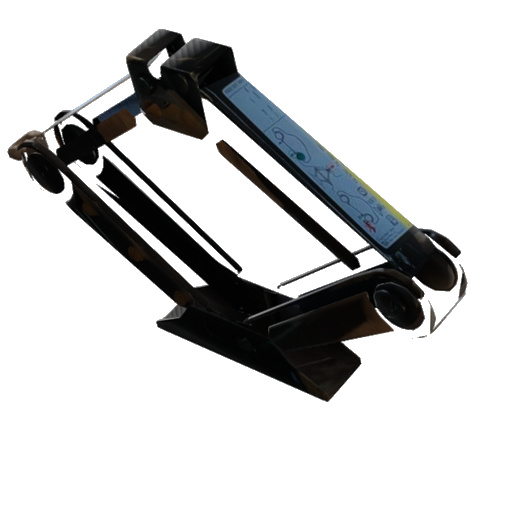} \\[14pt]

            GT & \qualresult{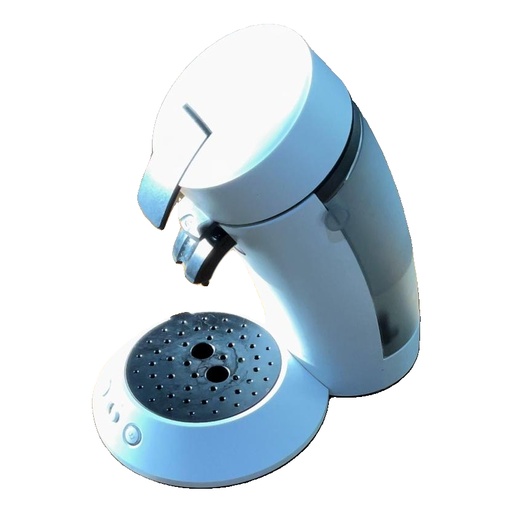} & \qualresult{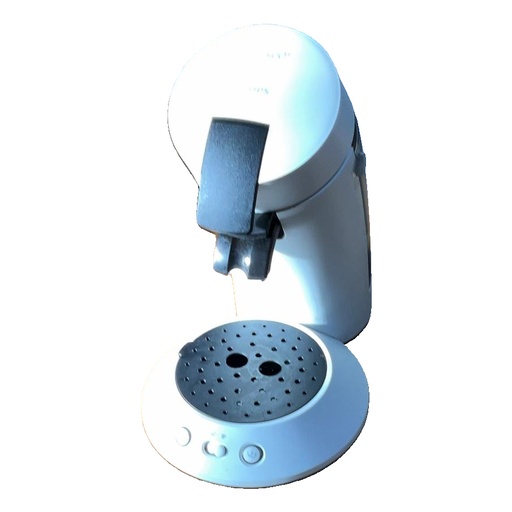} & \qualresult{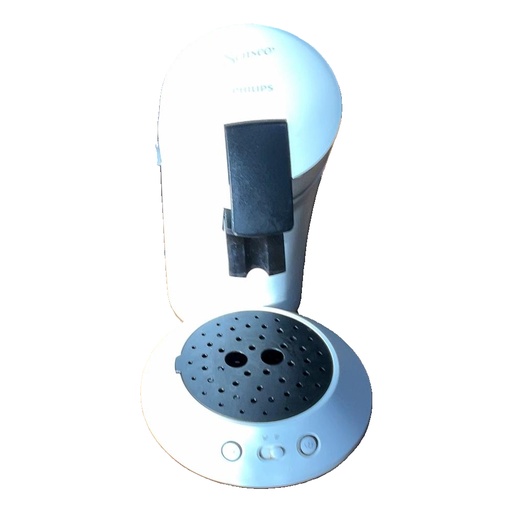} & \qualresult{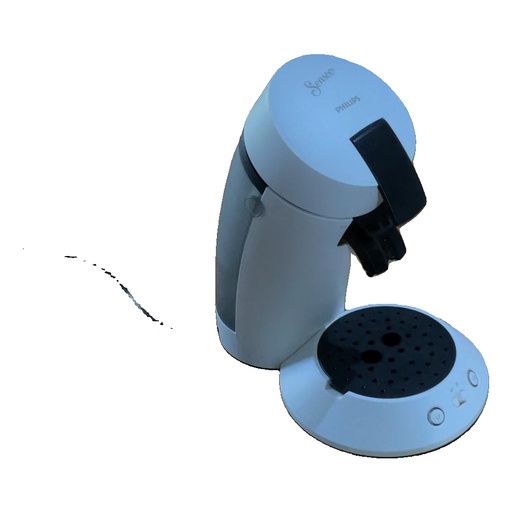} & \qualresult{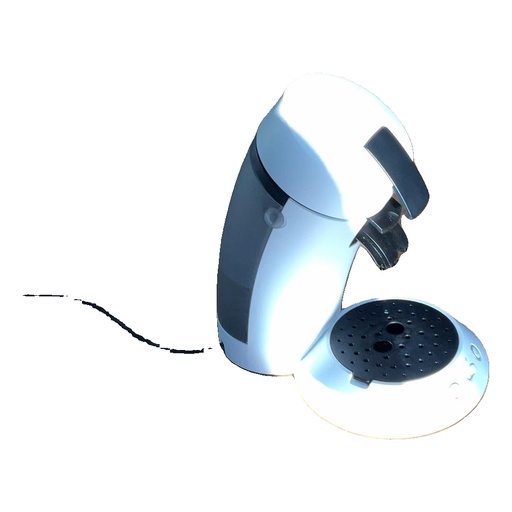} & \qualresult{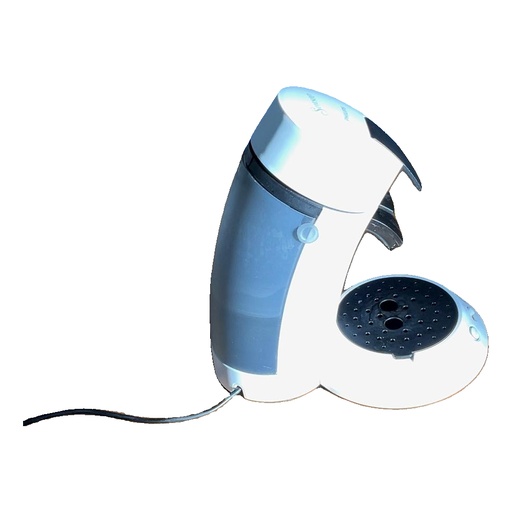} \\[10pt]
            \textbf{Ours} & \qualresult{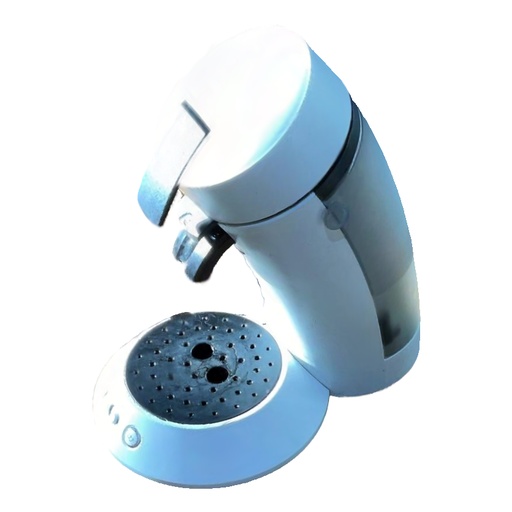} & \qualresult{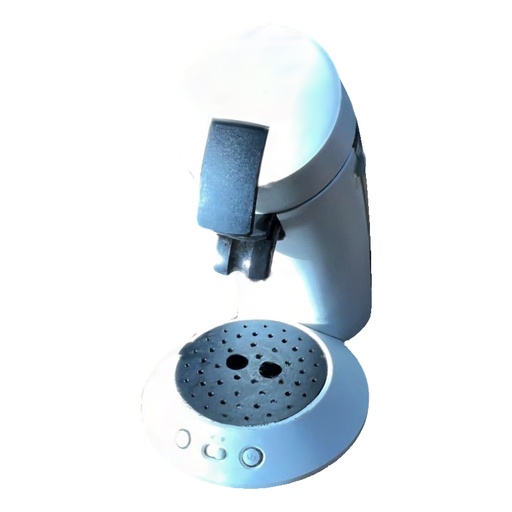} & \qualresult{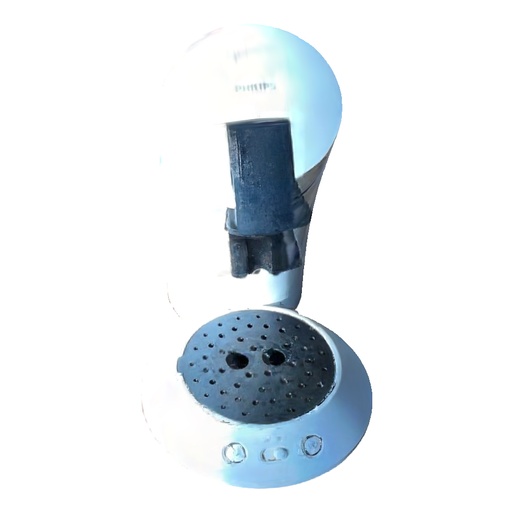} & \qualresult{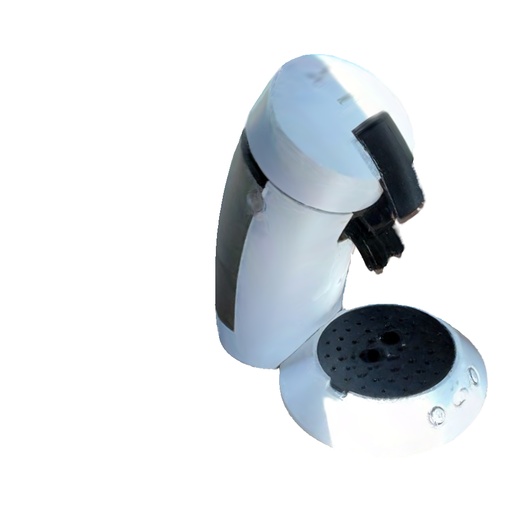} & \qualresult{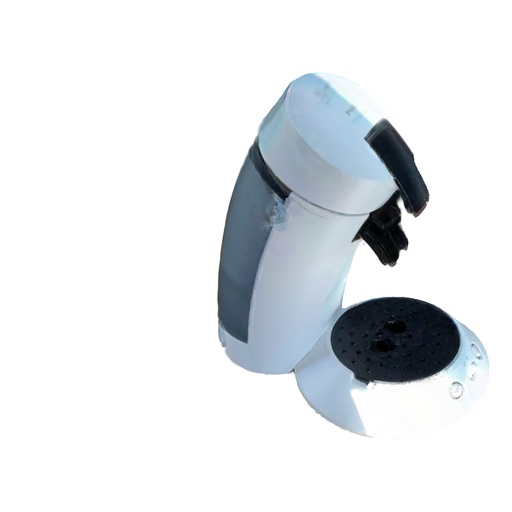} & \qualresult{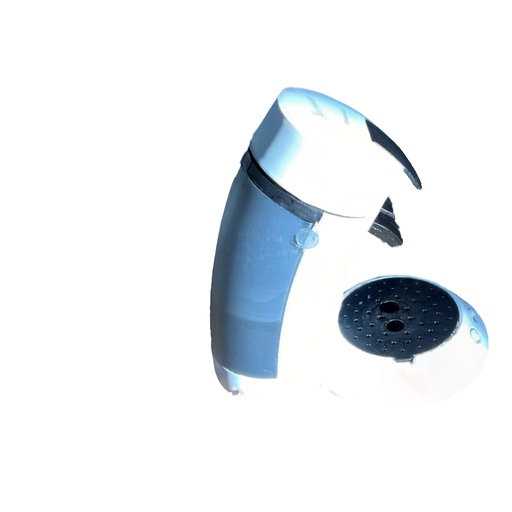} \\[10pt]
            FSGS~\cite{zhu2024fsgs} & \qualresult{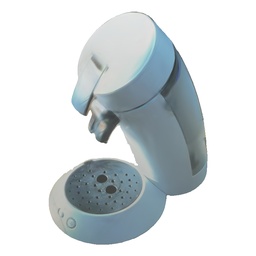} & \qualresult{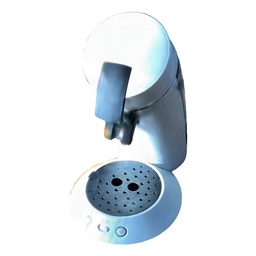} & \qualresult{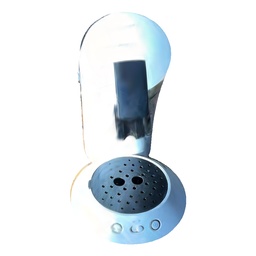} & \qualresult{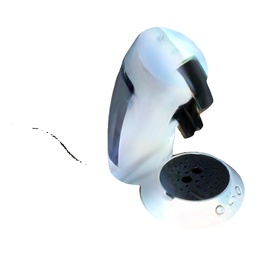} & \qualresult{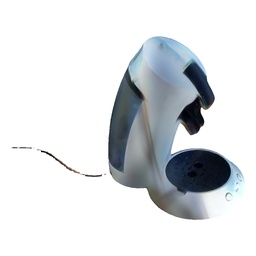} & \qualresult{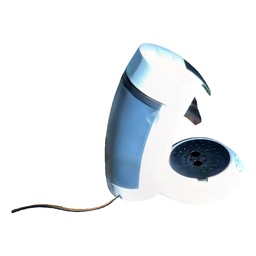} \\[10pt]
            InstantSplat~\cite{fan2024instantsplat} & \qualresult{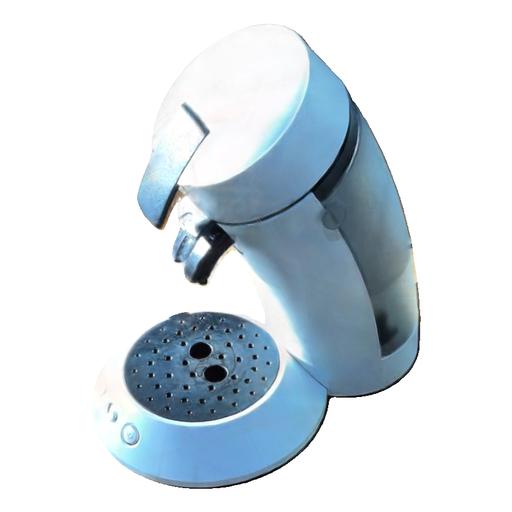} & \qualresult{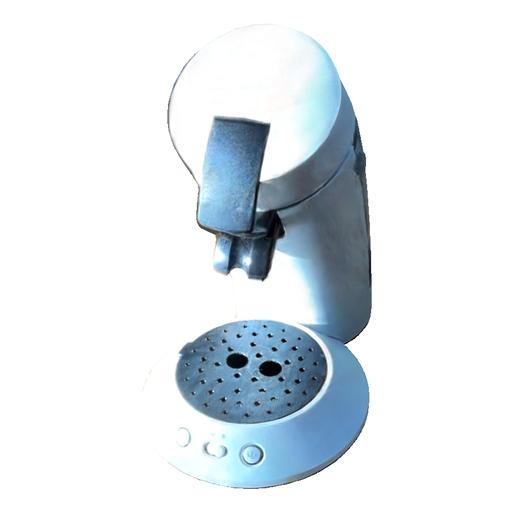} & \qualresult{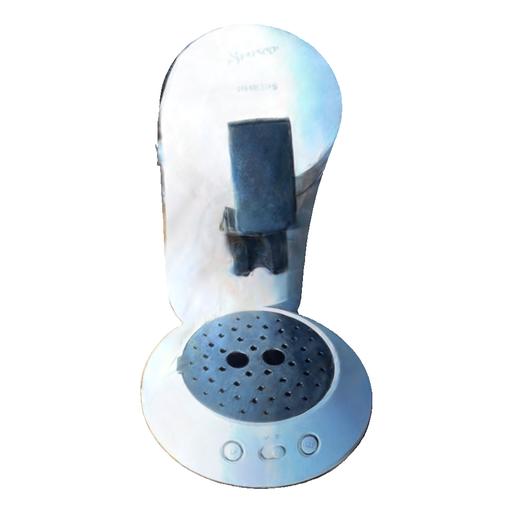} & \qualresult{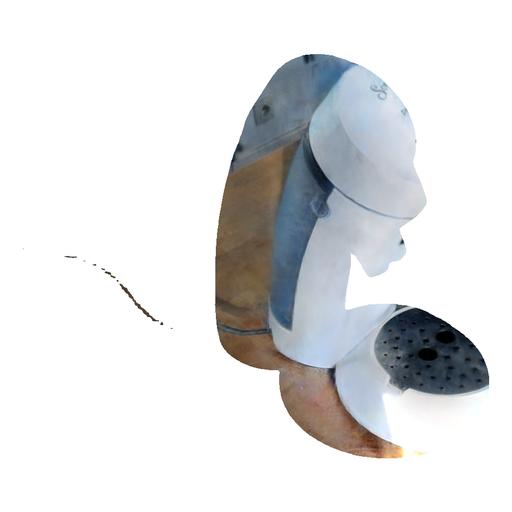} & \qualresult{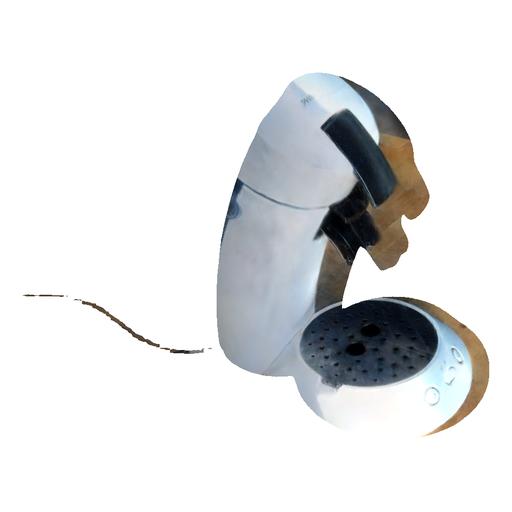} & \qualresult{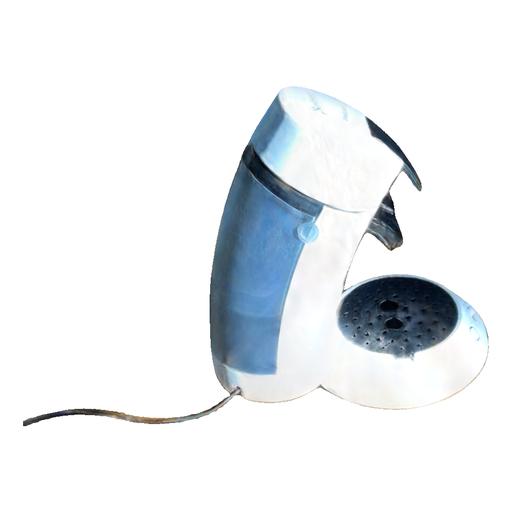} \\[10pt]
            MVS-Texturing~\cite{Waechter2014Texturing} & \qualresult{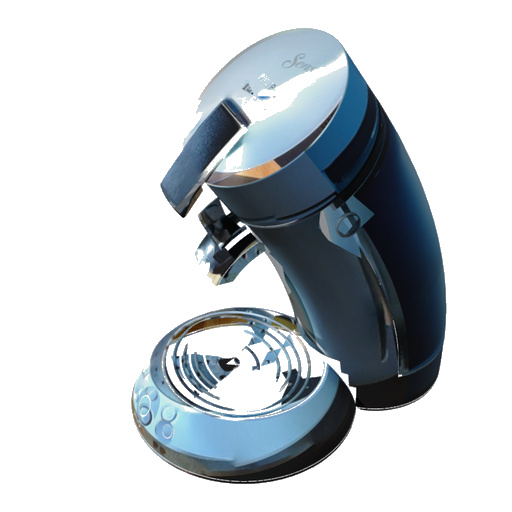} & \qualresult{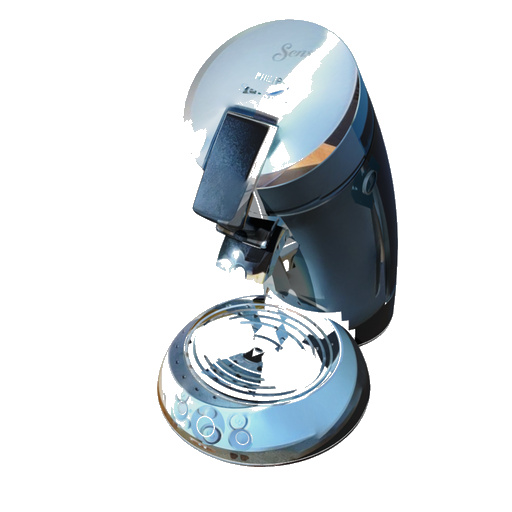} & \qualresult{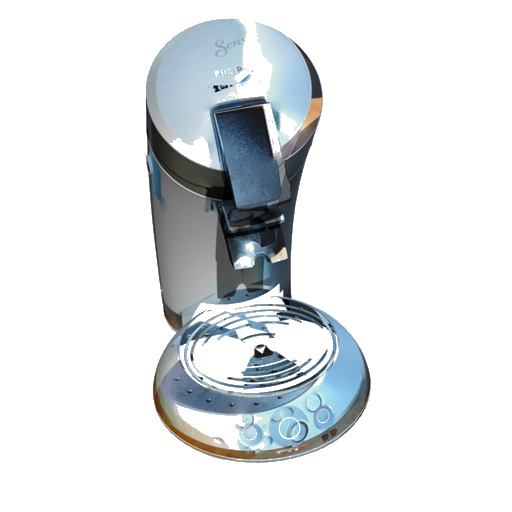} & \qualresult{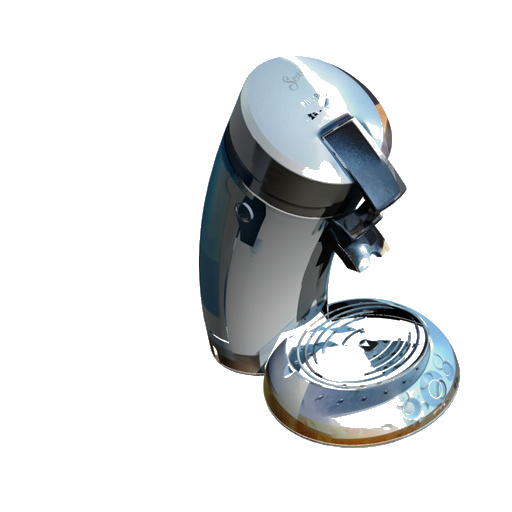} & \qualresult{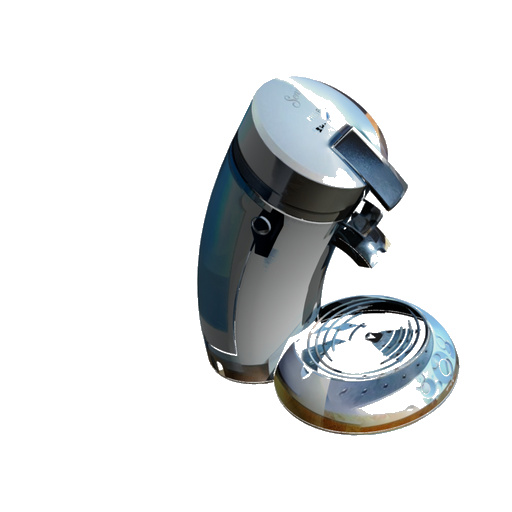} & \qualresult{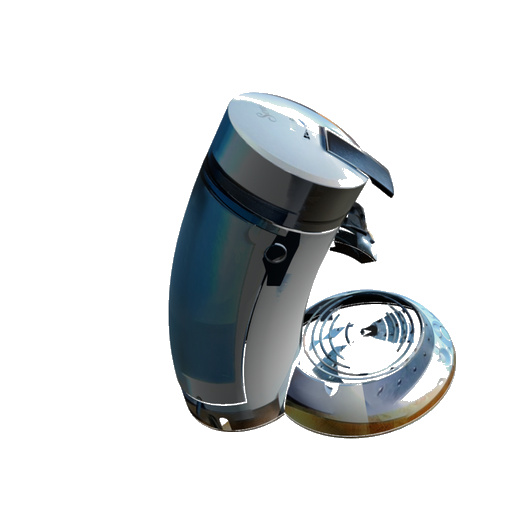}
    \end{tabular}%
    }%
\end{center}
    \caption{Qualitative comparison on sparse-view reconstruction ($9$--$13$ images). MVS-Texturing~\cite{Waechter2014Texturing} produces an inconsistent patchwork of textures and leaves weakly observed regions untextured. FSGS~\cite{zhu2024fsgs} suffers from over-smoothing. InstantSplat~\cite{fan2024instantsplat} is an SfM-free full-scene method; as it reconstructs the whole scene rather than the object, its tiles are shown composited with the ground-truth mask, and even then, it recovers only a coarse object with distorted geometry and smeared texture. CADSplat (Ours) maintains strict geometric fidelity, preserving sharp edges and accurate textures. \textit{Best viewed digitally and zoomed in.}}
    \label{fig:qualitative}
\end{figure*}

The qualitative comparison (\figurename~\ref{fig:qualitative}) shows the failure modes behind the numbers. FSGS produces reasonable reconstructions, but its textures are noticeably blurrier and over-smoothed compared to ours. MVS-Texturing~\cite{Waechter2014Texturing} struggles with wide-baseline views, leading to an inconsistent patchwork of texture fragments: individually correct from their source viewpoints, but with abrupt, incongruous transitions, misalignment artifacts, and untextured mesh regions where no source view provides coverage. A video of our reconstructions rendered along continuous novel-view trajectories across the six scenes is provided as Online Resource~1.

\subsubsection{Robustness to View Sparsity}
\label{sec:view_sparsity}

\begin{figure}[!htbp]
    \centering
    \adjustbox{max width=\columnwidth}{%
    \begin{tabular}{c@{\hspace{8pt}}c@{\hspace{8pt}}c@{\hspace{8pt}}c@{\hspace{8pt}}c}
        \textbf{3 Views} & \textbf{5 Views} & \textbf{8 Views} & \textbf{10 Views} \\
        
        \qualresulth{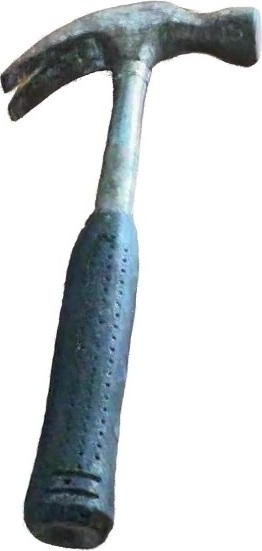} & \qualresulth{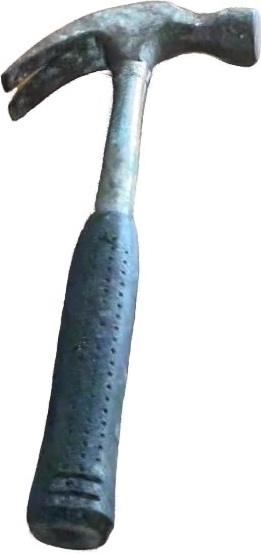} & \qualresulth{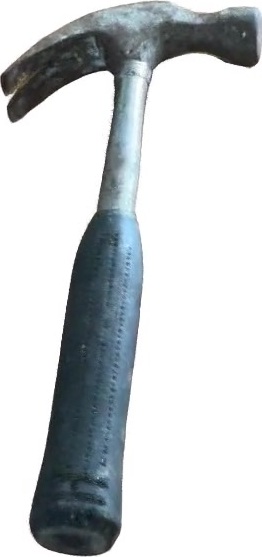} & \qualresulth{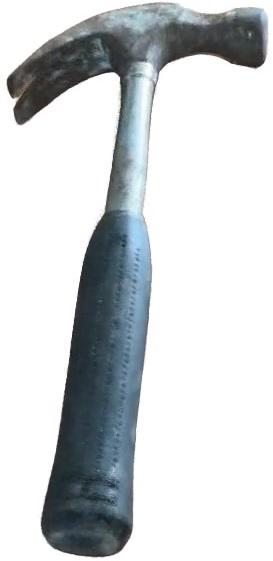} \\
        
        \qualresulth{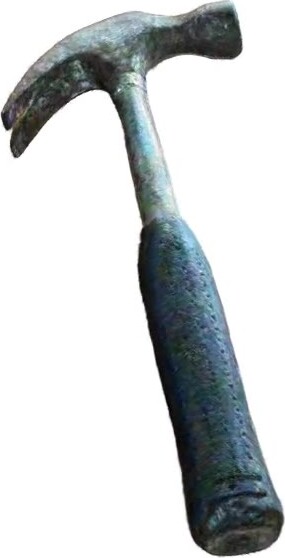} & \qualresulth{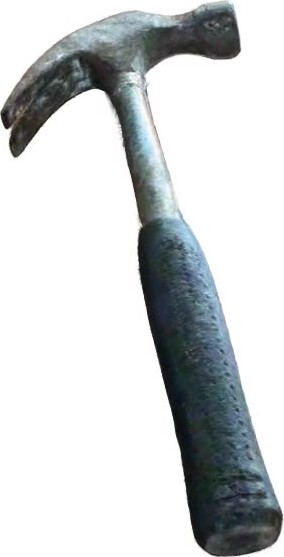} & \qualresulth{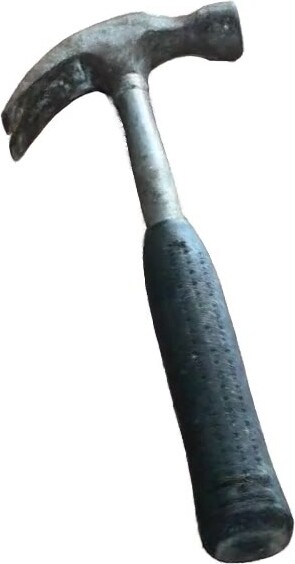} & \qualresulth{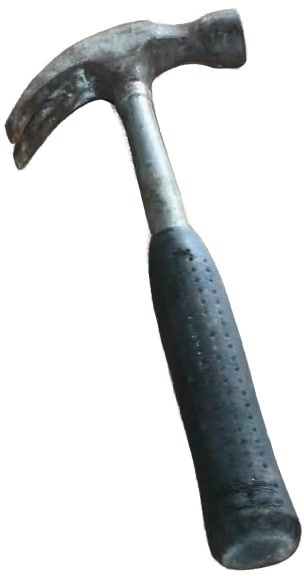} \\
    \end{tabular}%
    }%
    \caption{Qualitative evaluation of CADSplat across varying numbers of training views. Because the geometry is constrained by the CAD prior, the reconstruction degrades gracefully to blurrier textures rather than generating geometric hallucinations, even in extremely sparse regimes.}
    \label{fig:view_sparsity}
\end{figure}

\begin{table*}[!tp]
\centering
\caption{Impact of view sparsity on reconstruction quality, averaged over the six scenes of \textbf{SCO-CAD} and (per scene) over the FPS seeds, under the evaluation protocol of \cref{sec:eval_protocol}. Training views are selected by farthest-point sampling over camera centers; held-out views are the full held-out trajectory. The all-view row equals the Ours row of \cref{tab:eval_avg}.}
\label{tab:view_sparsity}
\resizebox{\textwidth}{!}{%
\begin{tabular}{l|cccc|cccc|cc}
\toprule
& \multicolumn{4}{c|}{\textbf{Object crop}} & \multicolumn{4}{c|}{\textbf{Eroded mask}} & \multicolumn{2}{c}{\textbf{Fg / Sil.}} \\
\textbf{Views} & \textbf{PSNR}\,$\uparrow$ & \textbf{SSIM}\,$\uparrow$ & \textbf{LPIPS}\,$\downarrow$ & \textbf{\flip{}}\,$\downarrow$ & \textbf{mPSNR}\,$\uparrow$ & \textbf{mSSIM}\,$\uparrow$ & \textbf{mLPIPS}\,$\downarrow$ & \textbf{m\flip{}}\,$\downarrow$ & \textbf{fgPSNR}\,$\uparrow$ & \textbf{IoU}\,$\uparrow$ \\ \midrule
3 Views & 20.710 & 0.845 & 0.122 & 0.116 & 25.171 & 0.898 & 0.083 & 0.084 & 17.012 & 0.929 \\
5 Views & 21.987 & 0.865 & 0.089 & 0.097 & 26.859 & 0.912 & 0.057 & 0.070 & 18.534 & 0.940 \\
8 Views & 22.568 & 0.876 & 0.073 & 0.086 & 27.871 & 0.921 & 0.045 & 0.059 & 19.410 & 0.942 \\
All Views (9--13) & 23.239 & 0.888 & 0.058 & 0.075 & 28.953 & 0.932 & 0.031 & 0.048 & 20.360 & 0.942 \\ \bottomrule
\end{tabular}%
}
\end{table*}

To evaluate how gracefully our method degrades under increasingly constrained data regimes, we ablate the number of training views from the full set of $9$--$13$ images down to $\{8, 5, 3\}$ on \textbf{SCO-CAD}. Naively sampling random images risks selecting clustered, nearby viewpoints, leaving the held-out views to observe regions never seen during training. Therefore, for each scene and each target count, we select the first training image randomly and the other training images by \textit{farthest-point sampling} (FPS) over the training-camera centers in the SfM frame. Because the first camera is chosen randomly, we run each scene \& number of views configuration with multiple random seeds and report the mean across seeds. 

Held-out views are scored with the full protocol of \cref{sec:eval_protocol}, and all other hyperparameters are unchanged from \cref{sec:implementation_details}. One caveat must be stated explicitly: the reduced-view conditions reuse the relative calibration and silhouette registration of the full sparse ($9$-$13$ views) capture. A genuine $3$-view capture would have to calibrate from those three wide-baseline images alone, which often exceeds what current SfM can deliver. This experiment, therefore, models the pre-calibrated-rig regime---where calibration is available regardless of how many views are captured---and measures how reconstruction quality degrades as views are removed.

\tablename~\ref{tab:view_sparsity} presents the quantitative degradation averaged over the six scenes, and \figurename~\ref{fig:view_sparsity} a qualitative visualization comparing novel viewpoints across view counts. We notice that our method degrades gracefully as views are removed. The decreased performance is concentrated mostly on appearance, not geometry: from all views down to three, the silhouette IoU falls only from $0.942$ to $0.929$ (a $1.4\%$ relative drop), whereas the perceptual distance mLPIPS grows $2.7\times$ ($0.031\rightarrow0.083$) and fgPSNR drops $3.3$\,dB ($20.36\rightarrow17.01$). Scarcity thus shows up as a noisier texture on a stable surface rather than as broken geometry. The per-scene breakdown shows that degradation depends on the object shape: geometrically simple objects where most of the shape is visible in a single image (hammer, scissors) are essentially saturated by $8$ views---hammer's $8$-view PSNR ($25.38$) even marginally exceeds its all-view score ($25.26$), and scissors gains under $0.2$\,dB from $5$ views onward---while more voluminous objects with significant per-image occlusion degrade more significantly. Even in the extreme $3$-view regime, the metrics remain usable (PSNR $20.7$, mPSNR $25.2$, IoU $0.93$). A video sweeping novel views at each view count is provided as Online Resource~2.

\subsubsection{Initialization: CAD Mesh vs.\ a Generic Sphere}
\label{sec:init_ablation}

To isolate the CAD model's contribution relative to the other pipeline components, we replace the mesh initialization with a sphere while keeping everything else identical. The sphere is positioned at the intersection of the cameras' viewing directions and rescaled to cover the same surface area as the ground-truth masks. To our genuine surprise, at the full view budget, the sphere initialization is not merely competitive but slightly ahead of the CAD initialization on the six-scene average of \emph{every} metric (\cref{tab:init}: PSNR $23.54$ vs.\ $23.24$, mPSNR $29.40$ vs.\ $28.95$, IoU $0.946$ vs.\ $0.942$). We report this openly, as it is informative rather than adverse: when $9$--$13$ views are available, the photometric and silhouette signals are sufficiently rich for the deformation field to carve the object out of a featureless sphere as well as from the CAD mesh.

The CAD prior becomes meaningful as views become scarce. \cref{tab:init} shows that as views decrease, initializing shape from CAD models becomes more important: at $3$ views the CAD initialization gives $+1.4$\,dB PSNR, $+1.1$\,dB mPSNR, $+1.0$\,dB fgPSNR, and a $+4.9$-point silhouette-IoU gap ($0.929$ vs.\ $0.879$). From only three images, the sphere cannot recover a correct silhouette without a shape prior, which the CAD prior supplies. The advantage of CAD initialization is especially visible in the group of voluminous objects with significant occlusion. The reason is clear: a generic sphere cannot infer the parts of a voluminous object that no training view observes from more than one angle---precisely the geometry the CAD prior fills in. The prior's value is therefore geometric disambiguation under occlusion and scarcity.

Rather than undermining our method, the sphere result is a testament to the strength of constrained deformation in producing high-fidelity, well-isolated objects. It generalizes better to held-out views than unconstrained masked 3DGS in sparse-view scenarios. Sphere-initialized CADSplat exceeds the vanilla 3DGS baseline of \cref{tab:eval_avg} by $+2.8$\,dB mPSNR, $+2.6$\,dB fgPSNR, and $+0.03$ IoU at the full budget. 

\begin{table*}[!tp]
\centering
\caption{Initialization ablation on \textbf{SCO-CAD}: CAD-mesh vs.\ unit-sphere initialization, six-scene averages at each view count (limited-view rows averaged over the FPS seeds), all else identical. \textbf{Bold} marks the better of the two per column; the winner crosses over from CAD at $3$ views to the sphere by $8$ views, isolating the CAD prior's value to the sparse regime.}
\label{tab:init}
\resizebox{\textwidth}{!}{%
\begin{tabular}{ll|ccc|ccc|cc}
\toprule
& & \multicolumn{3}{c|}{\textbf{Object crop}} & \multicolumn{3}{c|}{\textbf{Eroded mask}} & \multicolumn{2}{c}{\textbf{Fg / Sil.}} \\
\textbf{Views} & \textbf{Init} & \textbf{PSNR}\,$\uparrow$ & \textbf{SSIM}\,$\uparrow$ & \textbf{LPIPS}\,$\downarrow$ & \textbf{mPSNR}\,$\uparrow$ & \textbf{mLPIPS}\,$\downarrow$ & \textbf{m\flip{}}\,$\downarrow$ & \textbf{fgPSNR}\,$\uparrow$ & \textbf{IoU}\,$\uparrow$ \\ \midrule
\multirow{2}{*}{3}
 & CAD    & \textbf{20.710} & \textbf{0.845} & \textbf{0.122} & \textbf{25.171} & \textbf{0.083} & \textbf{0.084} & \textbf{17.012} & \textbf{0.929} \\
 & Sphere & 19.312 & 0.830 & 0.152 & 24.099 & 0.098 & 0.088 & 16.005 & 0.879 \\ \midrule
\multirow{2}{*}{5}
 & CAD    & \textbf{21.987} & \textbf{0.865} & \textbf{0.089} & 26.859 & \textbf{0.057} & 0.070 & 18.534 & \textbf{0.940} \\
 & Sphere & 21.711 & 0.861 & 0.099 & \textbf{26.992} & 0.063 & 0.070 & \textbf{18.687} & 0.930 \\ \midrule
\multirow{2}{*}{8}
 & CAD    & 22.568 & 0.876 & \textbf{0.073} & 27.871 & 0.045 & 0.059 & 19.410 & 0.942 \\
 & Sphere & \textbf{22.783} & \textbf{0.881} & 0.074 & \textbf{28.318} & \textbf{0.043} & \textbf{0.057} & \textbf{19.850} & 0.942 \\ \midrule
\multirow{2}{*}{All}
 & CAD    & 23.239 & 0.888 & \textbf{0.058} & 28.953 & 0.031 & 0.048 & 20.360 & 0.942 \\
 & Sphere & \textbf{23.535} & \textbf{0.895} & 0.059 & \textbf{29.403} & \textbf{0.030} & \textbf{0.046} & \textbf{20.807} & \textbf{0.946} \\ \bottomrule
\end{tabular}%
}
\end{table*}

\subsubsection{Ablation of Design Decisions}

\begin{table*}[!tp]
\centering
\caption{Ablation of CADSplat's components, six-scene averages under the protocol of \cref{sec:eval_protocol}. Components accumulate down the table, from the CAD mesh with mask supervision only (first row) to the full method (last row, which equals the Ours row in \cref{tab:eval_avg}). ``Mask align'' is the warmup phase that delays appearance optimization until deformation and pose have settled against the silhouette. \textbf{Bold} marks the best value per column.}
\label{tab:ablation}
\resizebox{\textwidth}{!}{%
\begin{tabular}{l|cccc|cccc|c|c}
\toprule
& \multicolumn{4}{c|}{\textbf{Object crop}} & \multicolumn{4}{c|}{\textbf{Eroded mask}} & \textbf{Fg} & \textbf{Sil.} \\
\textbf{Configuration} & \textbf{PSNR}\,$\uparrow$ & \textbf{SSIM}\,$\uparrow$ & \textbf{LPIPS}\,$\downarrow$ & \textbf{\flip{}}\,$\downarrow$ & \textbf{mPSNR}\,$\uparrow$ & \textbf{mSSIM}\,$\uparrow$ & \textbf{mLPIPS}\,$\downarrow$ & \textbf{m\flip{}}\,$\downarrow$ & \textbf{fgPSNR}\,$\uparrow$ & \textbf{IoU}\,$\uparrow$ \\ \midrule
No deformation, no pose opt.            & 17.944 & 0.796 & 0.237 & 0.155 & 21.881 & 0.874 & 0.135 & 0.096 & 13.810 & 0.793 \\
Deformation, no pose opt., mask align   & 22.917 & 0.879 & 0.072 & 0.076 & 28.376 & 0.924 & 0.041 & 0.049 & 19.629 & 0.941 \\
No deformation, pose opt., mask align   & 22.667 & 0.878 & 0.081 & 0.079 & 28.193 & 0.925 & 0.041 & 0.050 & 19.505 & 0.934 \\
Deformation, pose opt., no mask align   & 18.880 & 0.821 & 0.217 & 0.140 & 24.461 & 0.897 & 0.111 & 0.075 & 16.231 & 0.784 \\
Deformation, pose opt., mask align      & 22.926 & 0.879 & 0.068 & 0.076 & 28.412 & 0.924 & 0.039 & 0.049 & 19.663 & 0.941 \\
\quad + SH neighbor smoothness          & 23.126 & 0.886 & 0.061 & 0.075 & 28.740 & 0.930 & 0.032 & 0.048 & 20.183 & 0.942 \\
\quad + Splat dropout (\textbf{Ours})   & \textbf{23.239} & \textbf{0.888} & \textbf{0.058} & \textbf{0.075} & \textbf{28.953} & \textbf{0.932} & \textbf{0.031} & \textbf{0.048} & \textbf{20.360} & \textbf{0.942} \\
\bottomrule
\end{tabular}%
}
\end{table*}

\Cref{tab:ablation} isolates each component's contribution as a six-scene average, accumulating one component at a time from the bare CAD-plus-masked ground truth baseline (PSNR $17.94$, IoU $0.79$) to the full method (PSNR $23.24$, IoU $0.94$).

\begin{figure}[!htbp]
        \centering
        \adjustbox{max width=\columnwidth}{%
        \begin{tabular}{m{2.5cm}c@{}c@{\hspace{10pt}}c@{\hspace{10pt}}}
            \textbf{Experiment} & & \\
            Ground-Truth Image & \qualresult{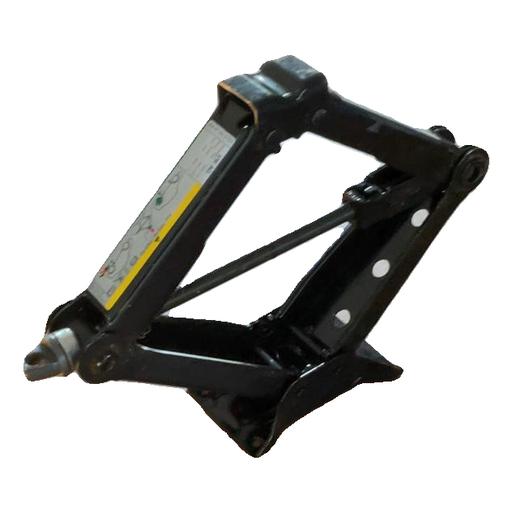} & \qualresult{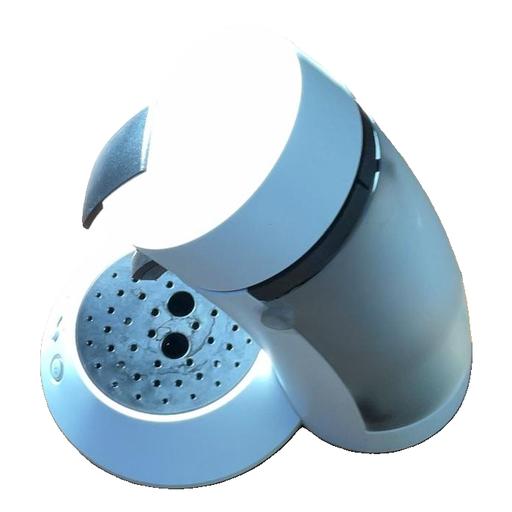} \\[10pt]
            
            No Deformation, No Pose Opt. & \qualresult{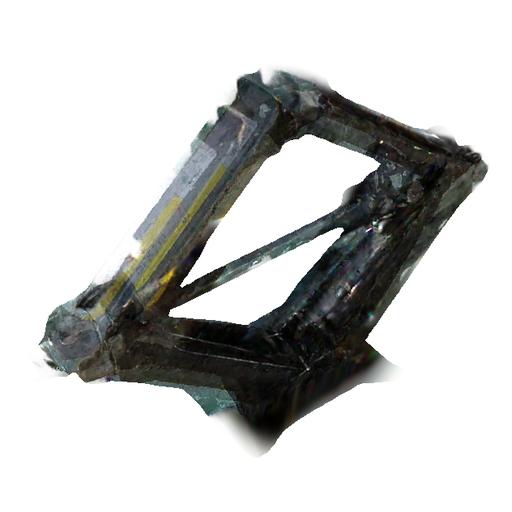} & \qualresult{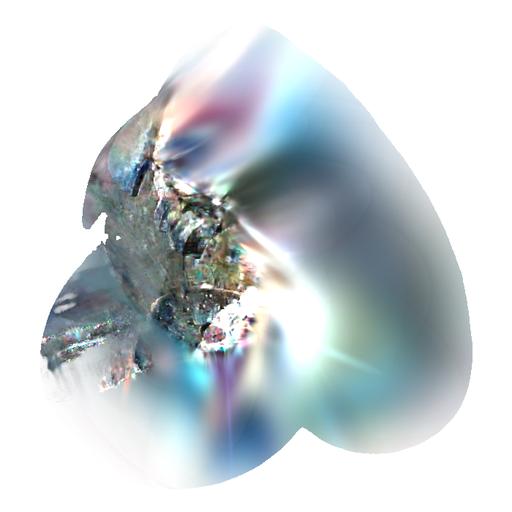} \\[10pt]
            
            Deformation, No Pose Opt. & \qualresult{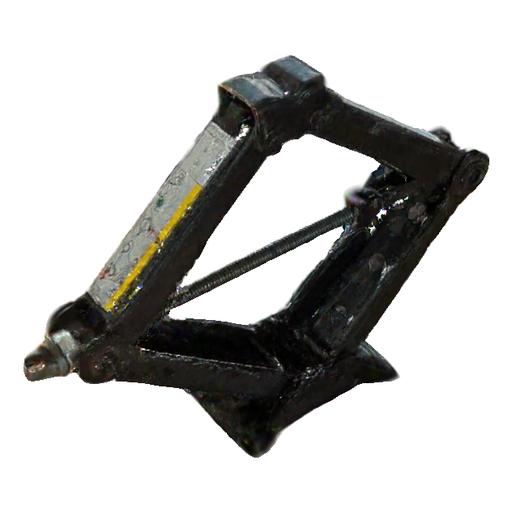} & \qualresult{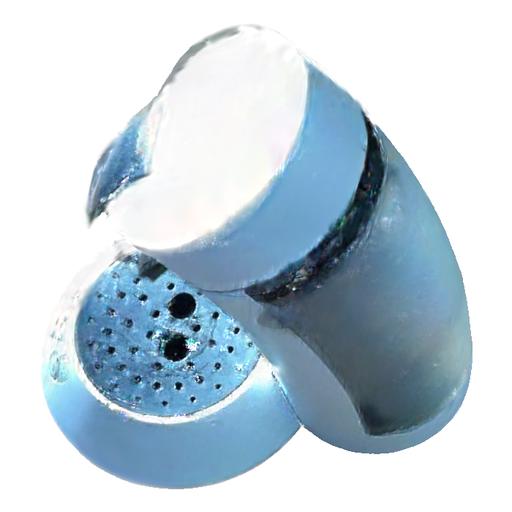} \\[10pt]
            
            No Deformation, Pose Opt. & \qualresult{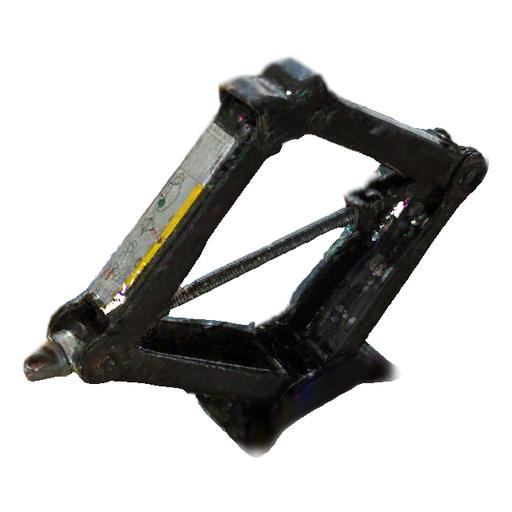} & \qualresult{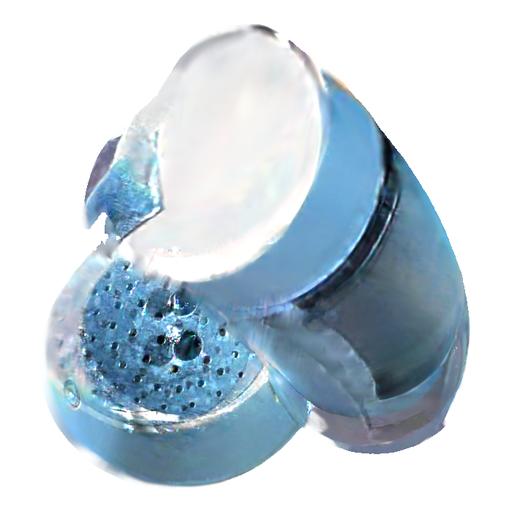} \\[10pt]
            
            Deformation, Pose Opt. & \qualresult{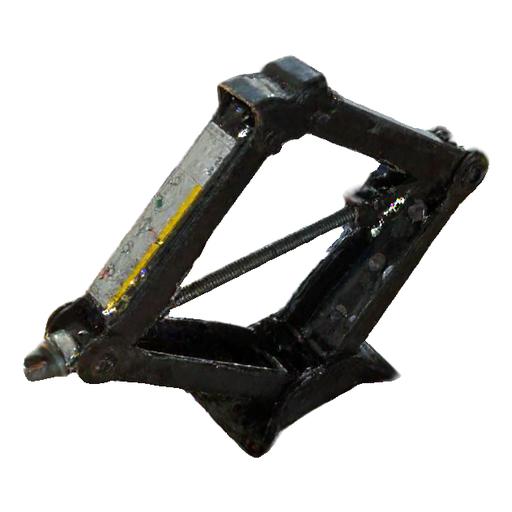} & \qualresult{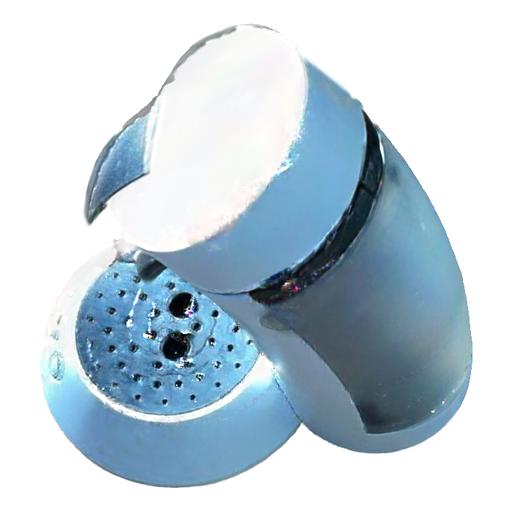} \\[10pt]
            
            +SH smoothness & \qualresult{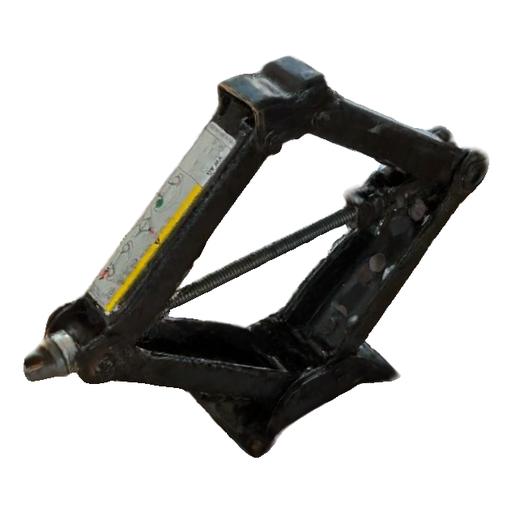} & \qualresult{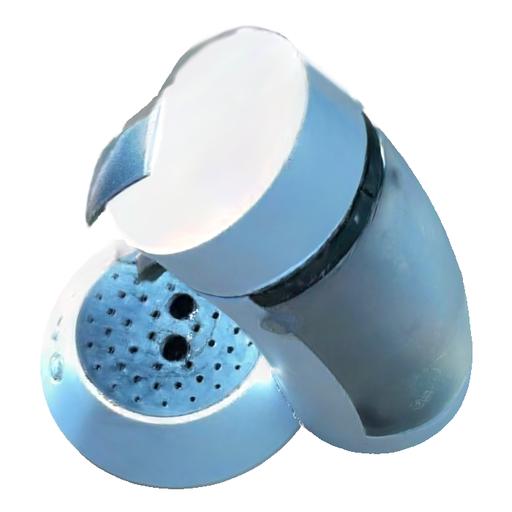} \\[10pt]
            
            +dropout & \qualresult{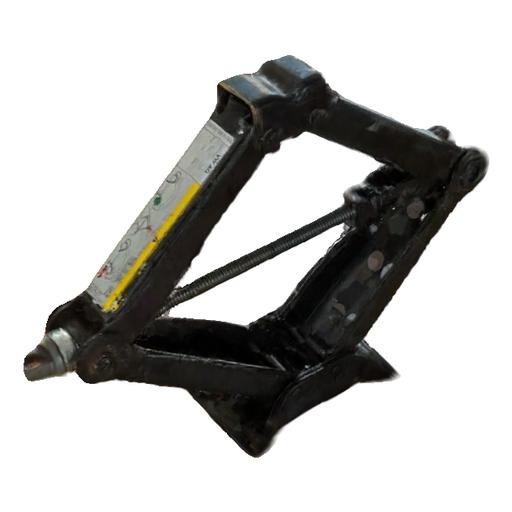} & \qualresult{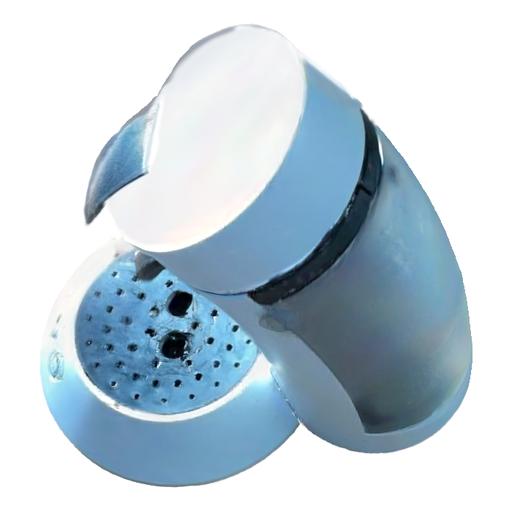} \\[10pt]
            
    \end{tabular}%
    }%
    \caption{Visual ablation study demonstrating the impact of architectural decisions on the reconstruction quality.}
    \label{fig:ablation}
\end{figure}

\textbf{Effect of Deformation:} Disabling the deformation network $\mathcal{D}_\theta$ forces the system to treat the CAD model as perfectly rigid, so residual CAD-to-object mismatch surfaces as ghosting (\cref{fig:ablation}). Deformation alone recovers most of the fidelity---PSNR $22.92$, mPSNR $28.38$, IoU $0.941$, within $0.3$\,dB of the full method---as the field ``wraps'' the CAD mesh onto the physical observation and can autonomously correct for pose misalignment by rigidly deforming the object to match camera viewpoints. However, because we care about the correctness of camera-to-object poses as well, we also include a separate pose optimization step.

\textbf{Effect of Pose Optimization:} Refining a single global Sim(3) during training mitigates inaccuracies from the consensus registration. On its own, it reaches PSNR $22.67$. Its marginal gain on top of deformation is small in the six-scene average, but is concentrated in the scenes whose calibration is least accurate. In contrast to the deformation network, it cannot recover non-rigid shape discrepancies, so on its own it only suffices when the CAD model is accurate.

\textbf{Effect of Mask Alignment:} We investigate whether the geometric warmup is needed, i.e., whether pose and deformation can be learned from the photometric signal directly, alongside the appearance. Without it, the texture is fitted to a still-misregistered surface and the geometry never recovers: the reconstruction collapses to PSNR $18.88$ and IoU $0.784$, below either geometric component used alone.

\textbf{Effect of SH Neighbor Smoothness and Splat Dropout:} The two appearance regularizers introduced in \cref{sec:reg} each add a small, consistent gain on top of the joint deformation/pose/alignment baseline. Pulling each splat's SH coefficients toward those of its $20$ canonical-neighbor splats (PSNR $22.93{\rightarrow}23.13$, mLPIPS $0.039{\rightarrow}0.032$) removes high-frequency color changes in unobserved views; splat dropout (PSNR $23.13{\rightarrow}23.24$, fgPSNR $20.18{\rightarrow}20.36$) ensures that splats occluded in the training views---but visible or sorted differently in held-out views---still receive an optimization signal. 

The visual impact of these decisions is shown in \cref{tab:ablation} and \cref{fig:ablation}, and animated across novel views in Online Resource~3.

\subsubsection{Camera-to-Object Registration Accuracy}
\label{sec:reg_accuracy}

The metrics above measure image quality. We also measure the accuracy of recovered camera-to-object pose directly, as it matters for downstream applications (\cref{sec:applications}). The ground-truth camera-to-object poses are annotated manually. For each scene, we annotate keypoint correspondences between the CAD model and several held-out views, triangulate those keypoints in the dense COLMAP frame, and fit the similarity transform between the triangulated points and their CAD counterparts \cite{umeyama1991least}. This yields a reference pose of the CAD model in the dense frame and, with it, a reference camera-to-object pose for every camera. We report two errors with respect to this reference: the geodesic rotation error of the camera orientations in degrees, and the camera-center error as a percentage of the camera-to-object distance, which makes it independent of the reconstruction's scale. For the two near-symmetric objects (carjack, skateboard), a silhouette cannot distinguish the object from its $180^\circ$ rotation, so we report the error against the closer of the two symmetric reference poses. Errors are medians over the training cameras, averaged over the six scenes. The silhouette consensus alone ($\mathbf{T}_{init}$) registers the object to within \textbf{$6.2^\circ$ rotation and $18.6\%$ position error} (per-scene range $3.1$--$10.2^\circ$ and $14$--$24\%$), and the joint pose optimization refines this to \textbf{$2.9^\circ$ and $6.1\%$} ($\mathbf{T}_{opt}$; range $1.2$--$4.2^\circ$ and $1.6$--$14.3\%$), reducing both errors on every scene except senseo's rotation, which is already accurate after the consensus step ($3.1^\circ\rightarrow3.3^\circ$). The largest remaining error is on carjack ($4.2^\circ$, $14.3\%$), the scene with the least accurate relative calibration.

\subsection{NeRS MVMC}
\label{sec:ners_mvmc}
The \textbf{NeRS MVMC} dataset \cite{zhang2021ners} consists of $20$ sparse-view car listings captured by untrained people who took a few pictures of their cars for online listing. For this dataset, we follow the evaluation methodology established by NeRS \cite{zhang2021ners}, performing both qualitative and quantitative evaluation. The pipeline is identical to that on \textbf{SCO-CAD}, except for the camera calibration. \textbf{SCO-CAD} provides an accurate relative calibration, whereas MVMC does not. Its listings are casual internet photos---few, wide-baseline views with inconsistent backgrounds---in which neither COLMAP~\cite{schoenberger2016sfm}, HLoc~\cite{sarlin2019coarse} nor VGGT~\cite{wang2025vggt} could identify poses for all viewpoints. We therefore use the camera-to-object poses shipped with the dataset, as NeRS itself does. These provided poses are only rough per-view estimates, not accurate relative calibrations. Consequently, whereas on \textbf{SCO-CAD} the pose optimization in \cref{sec:pose_opt} refines a single global $\mathrm{Sim}(3)$, here we let it learn an independent correction for each camera. The CAD library used to find the best shape match is the ShapeNetCore~\cite{chang2015shapenet} cars category.

\begin{figure}[!htbp]
    \centering
    \setlength{\tabcolsep}{2pt}
    \begin{tabular}{ccc}
        GT & NeRS \cite{zhang2021ners} & \textbf{Ours} \\
        \includegraphics[width=0.31\linewidth]{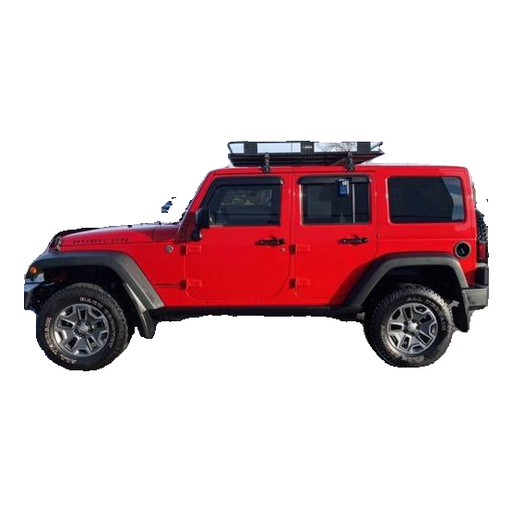} & \includegraphics[width=0.31\linewidth]{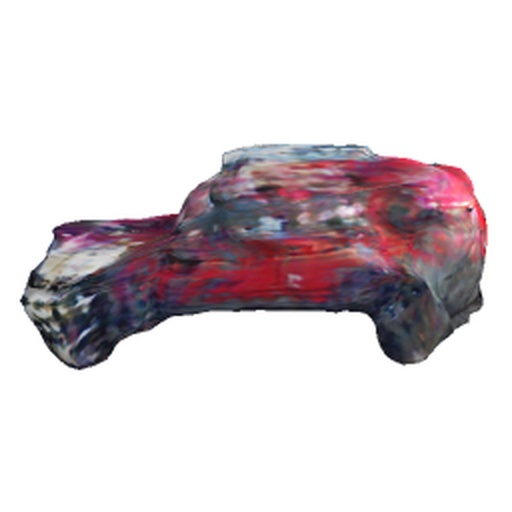} & \includegraphics[width=0.31\linewidth]{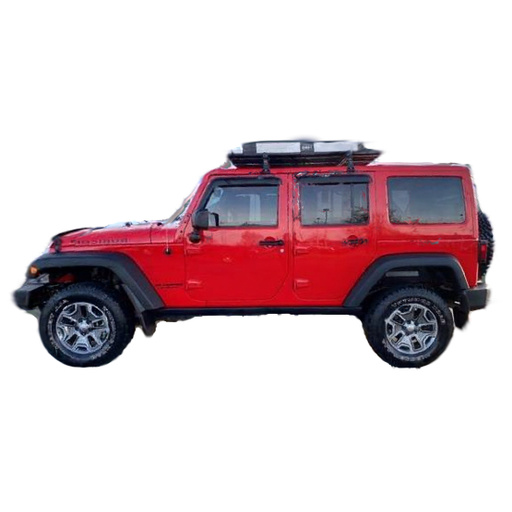} \\
        \includegraphics[width=0.31\linewidth]{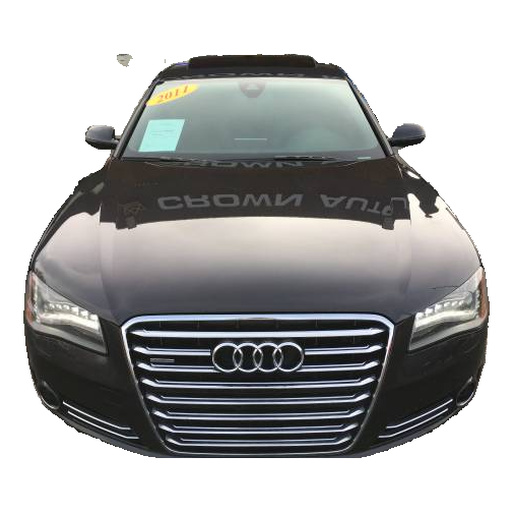} & \includegraphics[width=0.31\linewidth]{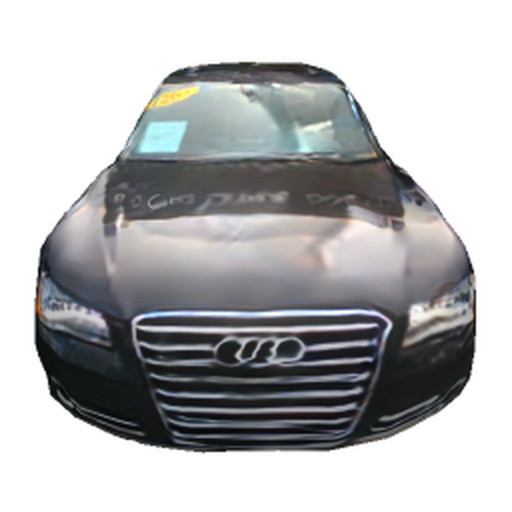} & \includegraphics[width=0.31\linewidth]{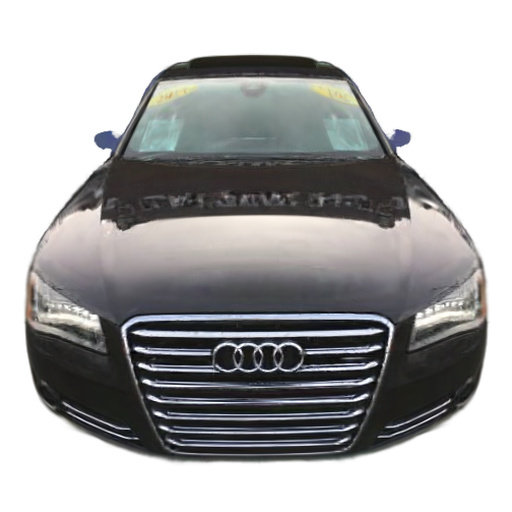} \\
        \includegraphics[width=0.31\linewidth]{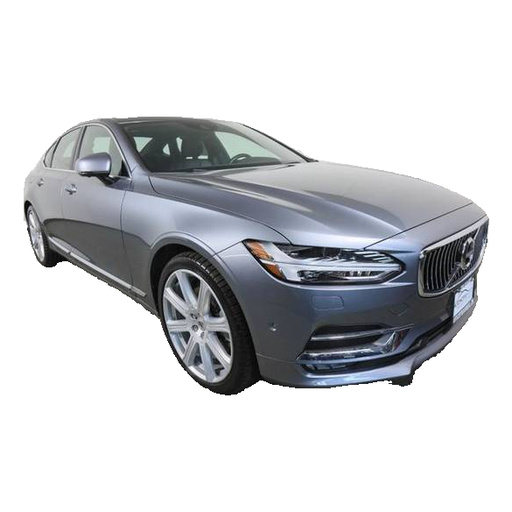} & \includegraphics[width=0.31\linewidth]{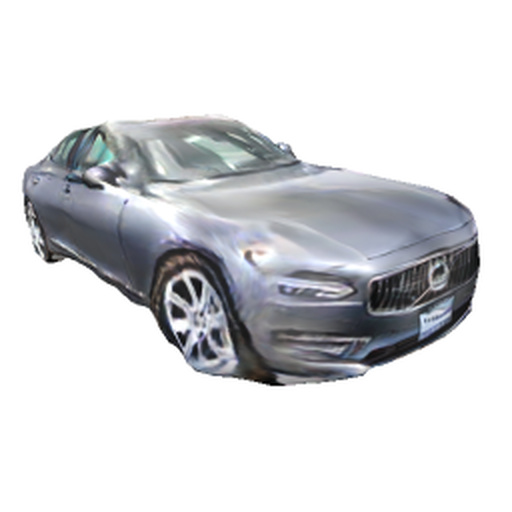} & \includegraphics[width=0.31\linewidth]{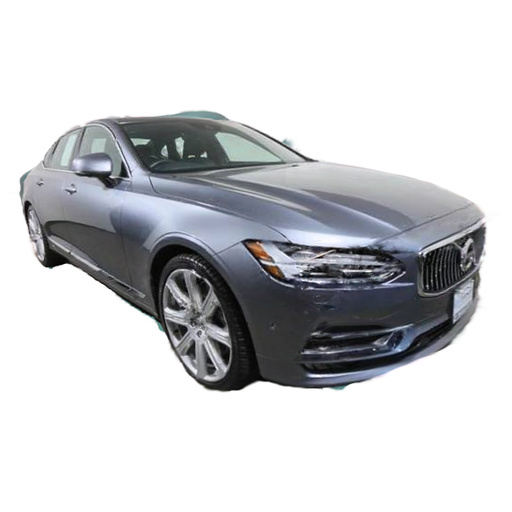} \\
        \includegraphics[width=0.31\linewidth]{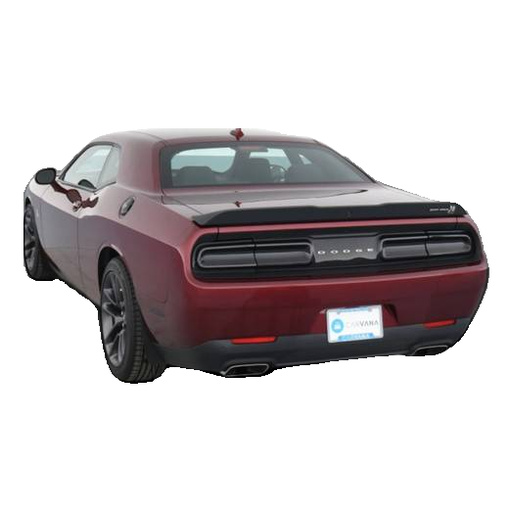} & \includegraphics[width=0.31\linewidth]{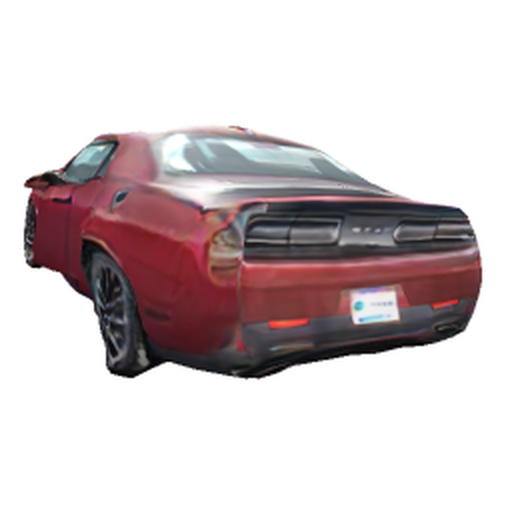} & \includegraphics[width=0.31\linewidth]{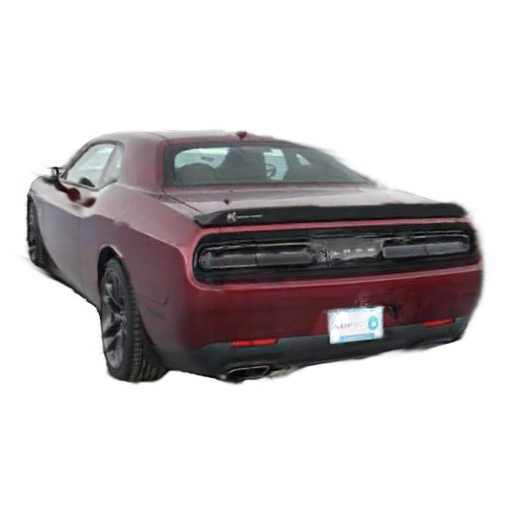} \\
    \end{tabular}
    \caption{Qualitative comparison on held-out views of the NeRS MVMC dataset. The NeRS renders are those released by its authors, produced in the same fixed, optimized-camera setting as in \cref{tab:ners_mvmc_results}. NeRS suffers from distorted geometry, smeared texture, and residual viewpoint misalignment, while our renders remain sharp and pose-accurate. CADSim \cite{cadsim} is excluded because the authors did not include renders in their paper, and their code is not open source.}
    \label{fig:ners_mvmc_qualitative}
\end{figure}

We compare against NeRS \cite{zhang2021ners} and CADSim \cite{cadsim}, which both report results on this dataset (CADNeRF~\cite{wen2025cad} released neither metrics nor code). All rows of \cref{tab:ners_mvmc_results} report the object-crop PSNR/SSIM/LPIPS (\cref{sec:eval_protocol})---the protocol NeRS established for this dataset and the one in which both baselines publish their numbers. The NeRS row is its strongest published setting (\emph{fixed optimized cameras}), which matches the per-camera test-time refinement we use. CADSim's row is copied from its paper, as its code is not released. Reading \Cref{tab:ners_mvmc_results} fairly requires spelling out how much category-specific knowledge each method builds in, because cars have a very specific appearance, material, and symmetrical properties. Both baselines rely heavily on these priors. NeRS represents shape as a network that maps every point on a unit sphere to a point on the object's surface, thereby keeping the surface watertight; on MVMC, however, this network is not initialized neutrally but is pre-trained so that the sphere maps onto a category-specific car template mesh before the input images are ever seen. Optimization, therefore, starts from an already car-shaped surface. NeRS further restricts appearance to a Phong reflectance model under a learned environment map---an illumination family that suits smooth, uniformly painted car bodies particularly well. CADSim embeds even stronger priors: it starts from a small curated set of part-annotated vehicle CAD models; it optimizes a vehicle-specific articulation model in which all wheels share a single mesh, the wheel positions are symmetric by construction; it enforces \emph{hard} left--right symmetry of the recovered geometry through an explicit symmetric Chamfer loss between the mesh and its own reflection; and where the capture provides it, it additionally consumes LiDAR depth. These priors are highly effective for vehicles and largely meaningless elsewhere. Our pipeline, by contrast, is the same generic one used throughout this paper; its only car-specific ingredient is a soft symmetry assumption over the car's length. Despite carrying the lightest prior load, our method achieves the best PSNR and SSIM among the comparison methods and improves on NeRS across all three metrics, while also enabling real-time rendering; only CADSim's LPIPS remains ahead of ours. The qualitative comparison in \cref{fig:ners_mvmc_qualitative} shows that NeRS's renders exhibit distorted geometry and smeared textures, whereas ours remain sharp and pose-accurate.

\begin{table}[!htbp]
\centering
\caption{Quantitative evaluation on the NeRS MVMC dataset, in the object-crop metrics (\cref{sec:eval_protocol}): the evaluation protocol established by NeRS, in which both baselines publish their numbers. The NeRS row is its strongest published setting (fixed optimized cameras); CADSim's row is transcribed from its paper, as its code is not released.}
\label{tab:ners_mvmc_results}
\begin{tabular*}{\linewidth}{@{\extracolsep{\fill}}l|ccc@{}}
\toprule
\textbf{Method} & \textbf{PSNR} $\uparrow$ & \textbf{SSIM} $\uparrow$ & \textbf{LPIPS} $\downarrow$ \\ \midrule
NeRS \cite{zhang2021ners} & 16.5 & 0.720 & 0.172 \\
CADSim \cite{cadsim} & 17.7 & 0.751 & \textbf{0.147} \\ \midrule
\textbf{Ours} & \textbf{18.7} & \textbf{0.767} & 0.157 \\ \bottomrule
\end{tabular*}
\end{table}

\subsection{NeRS Misc}
\begin{figure*}[!tp]
    \centering
    \adjustbox{max width=\textwidth}{%
    \begin{tabular}{lrc}
        \raisebox{\dimexpr-\height+0.9cm\relax}[0pt][0pt]{\includegraphics[height=140pt]{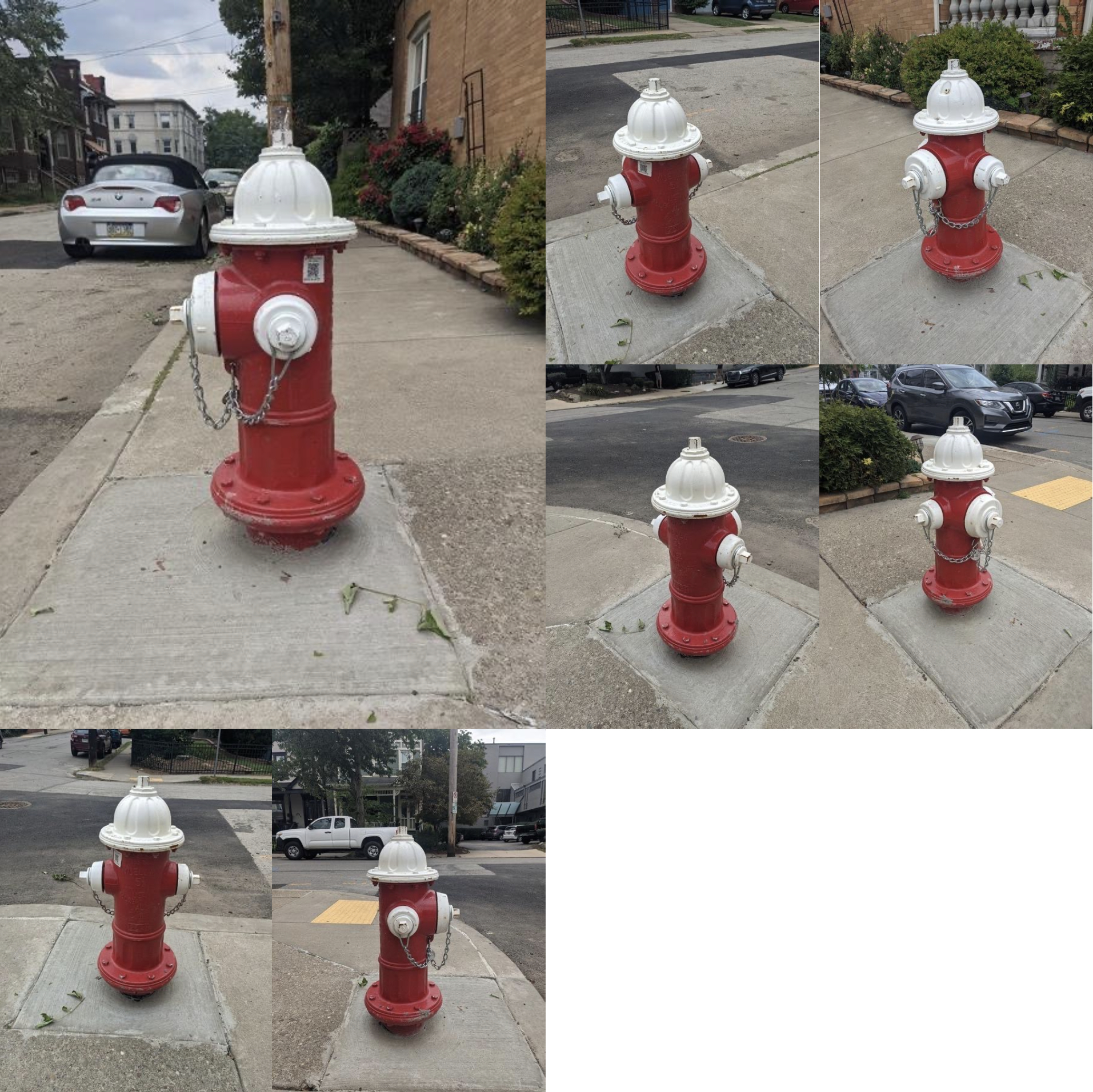}} & Ours & \qualresultners{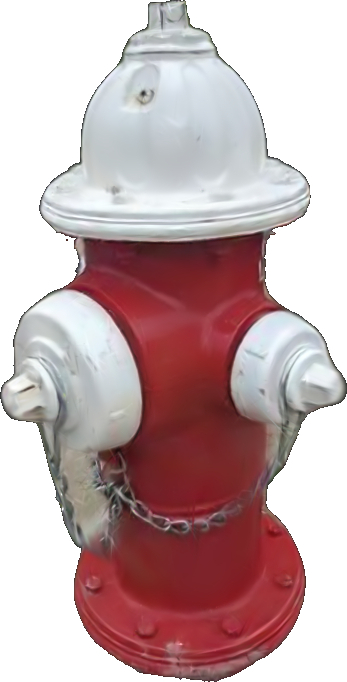} \qualresultners{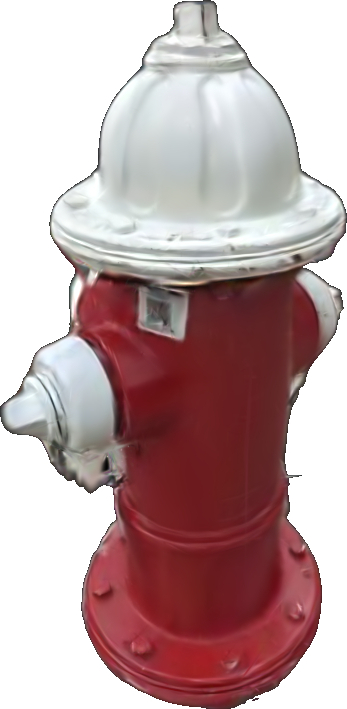} \qualresultners{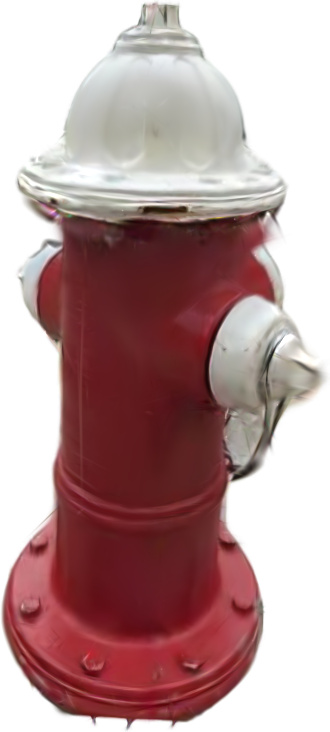} \\
        
        & CADNeRF \cite{wen2025cad} & \qualresultners{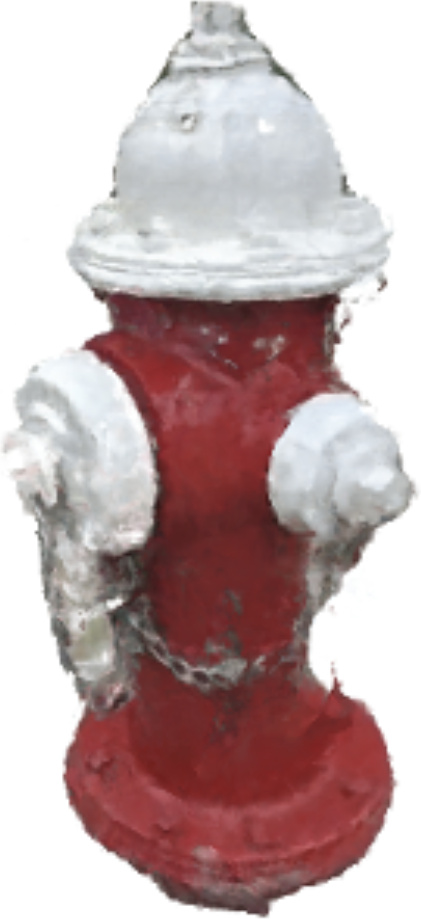} \qualresultners{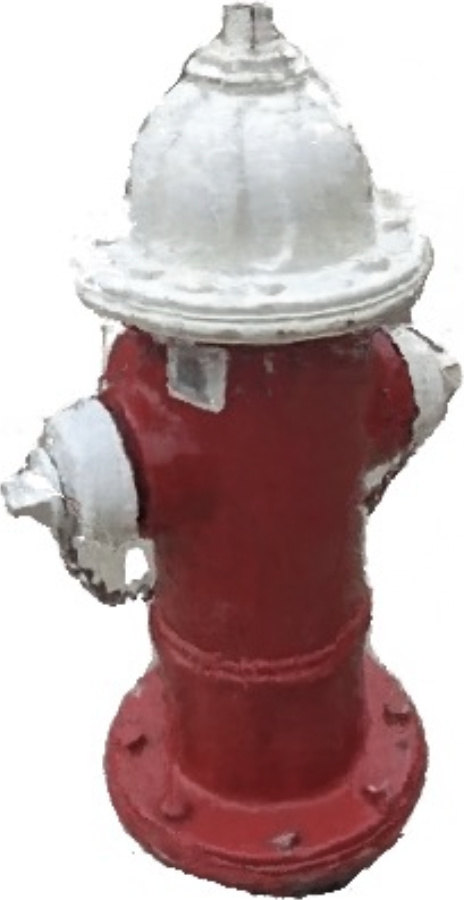} \qualresultners{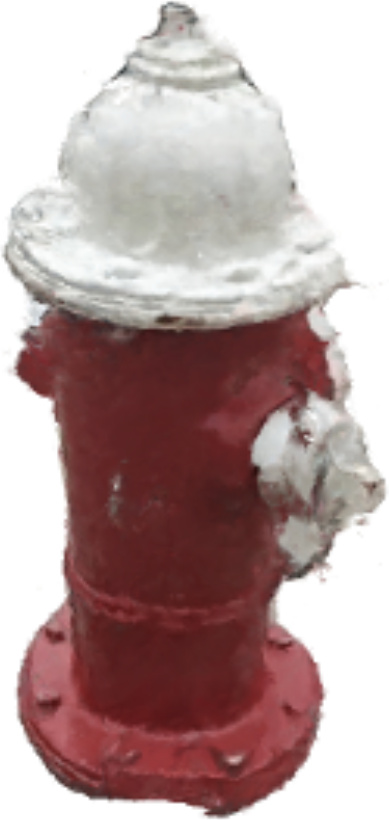} \\
        
        & NeRS \cite{zhang2021ners} & \qualresultners{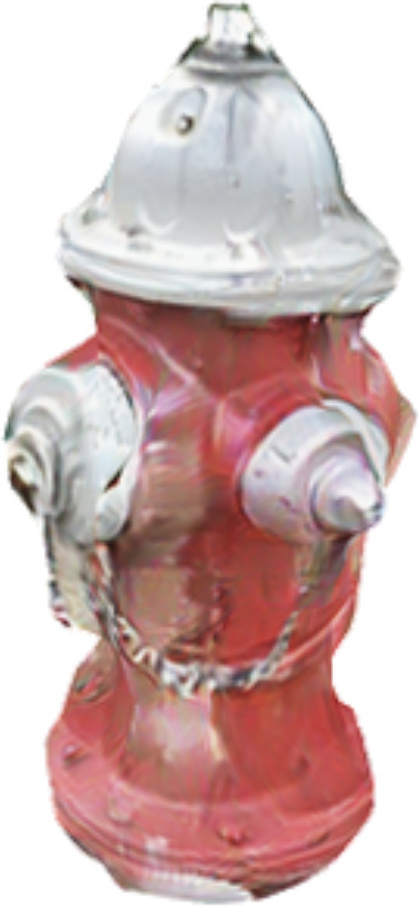} \qualresultners{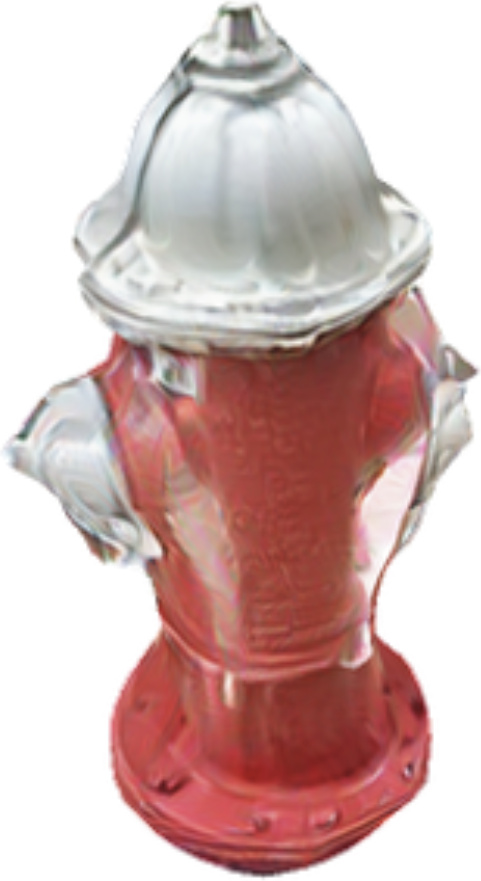} \qualresultners{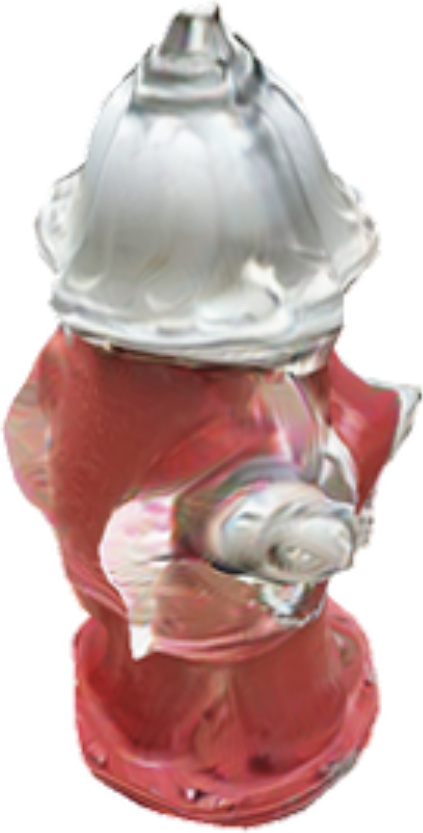} \\[30pt]

        \raisebox{\dimexpr-\height+0.9cm\relax}[0pt][0pt]{\includegraphics[height=130pt]{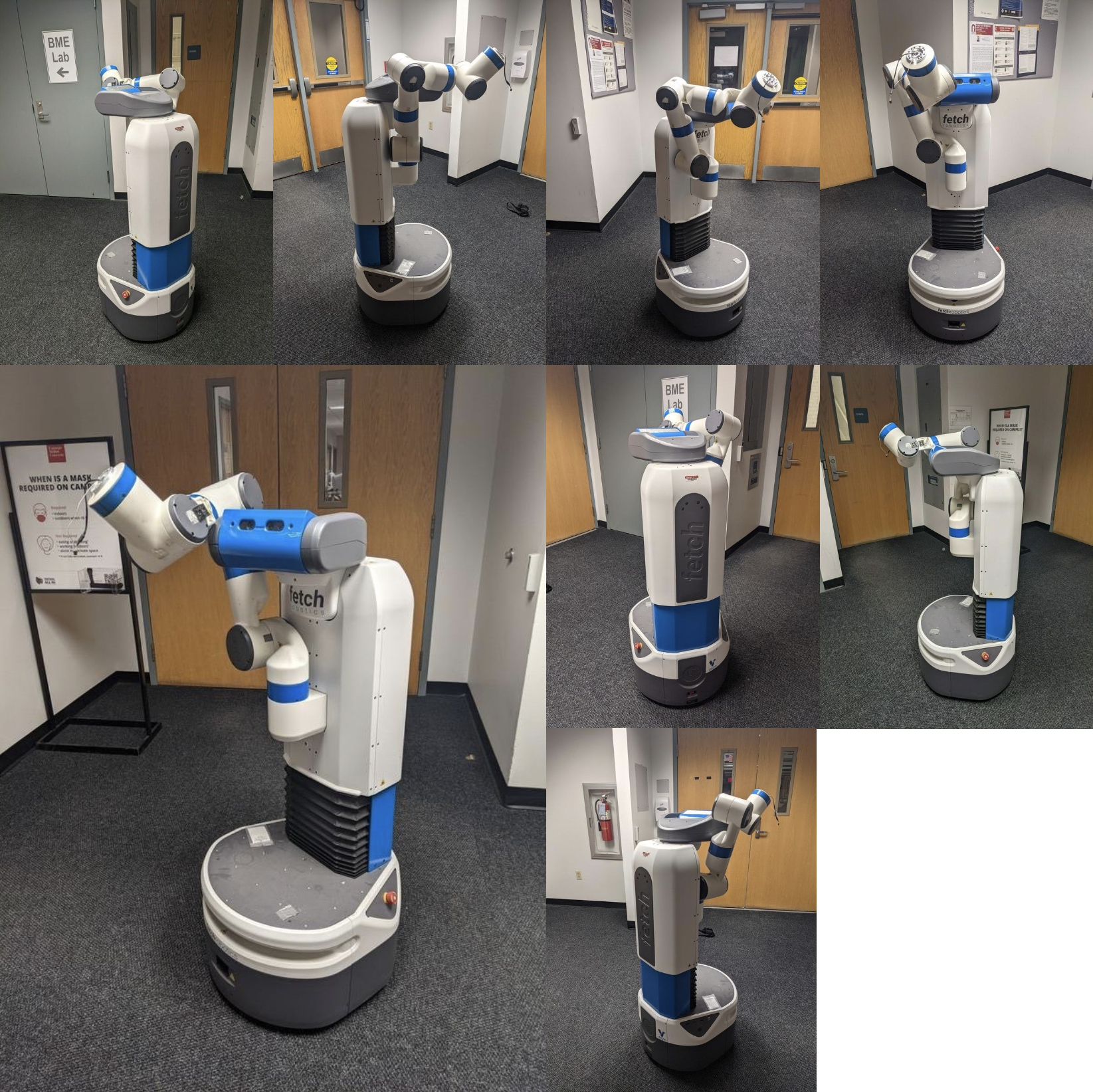}} & Ours & \qualresultners{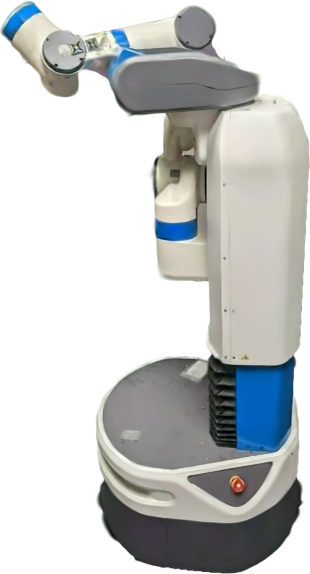} \qualresultners{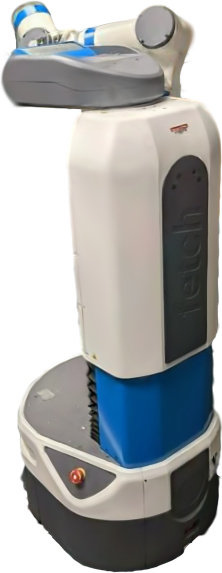} \qualresultners{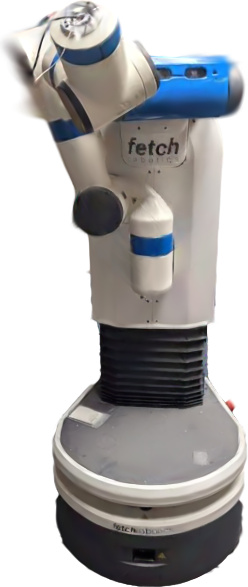} \\
        
        & CADNeRF \cite{wen2025cad} & \qualresultners{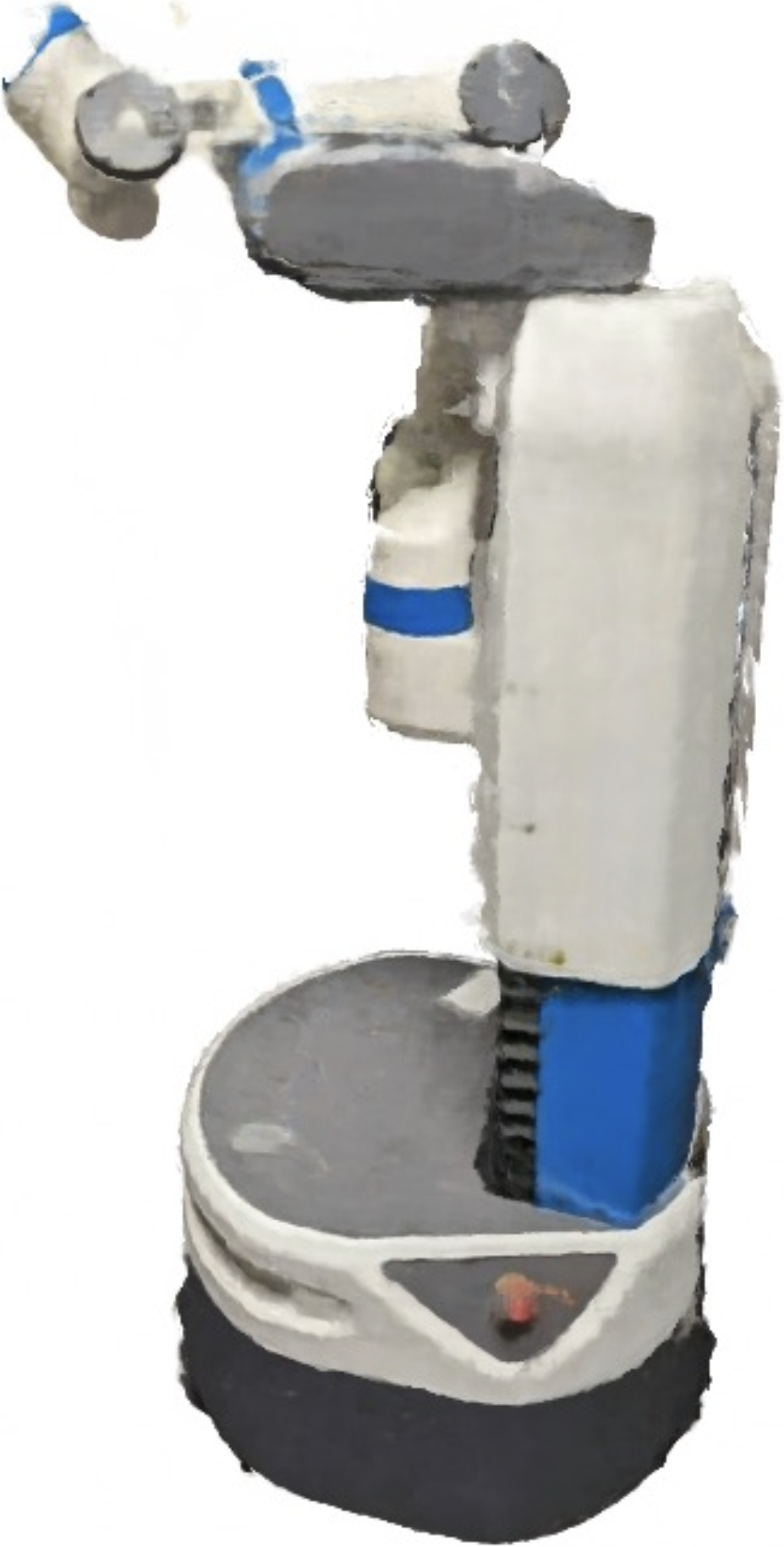} \qualresultners{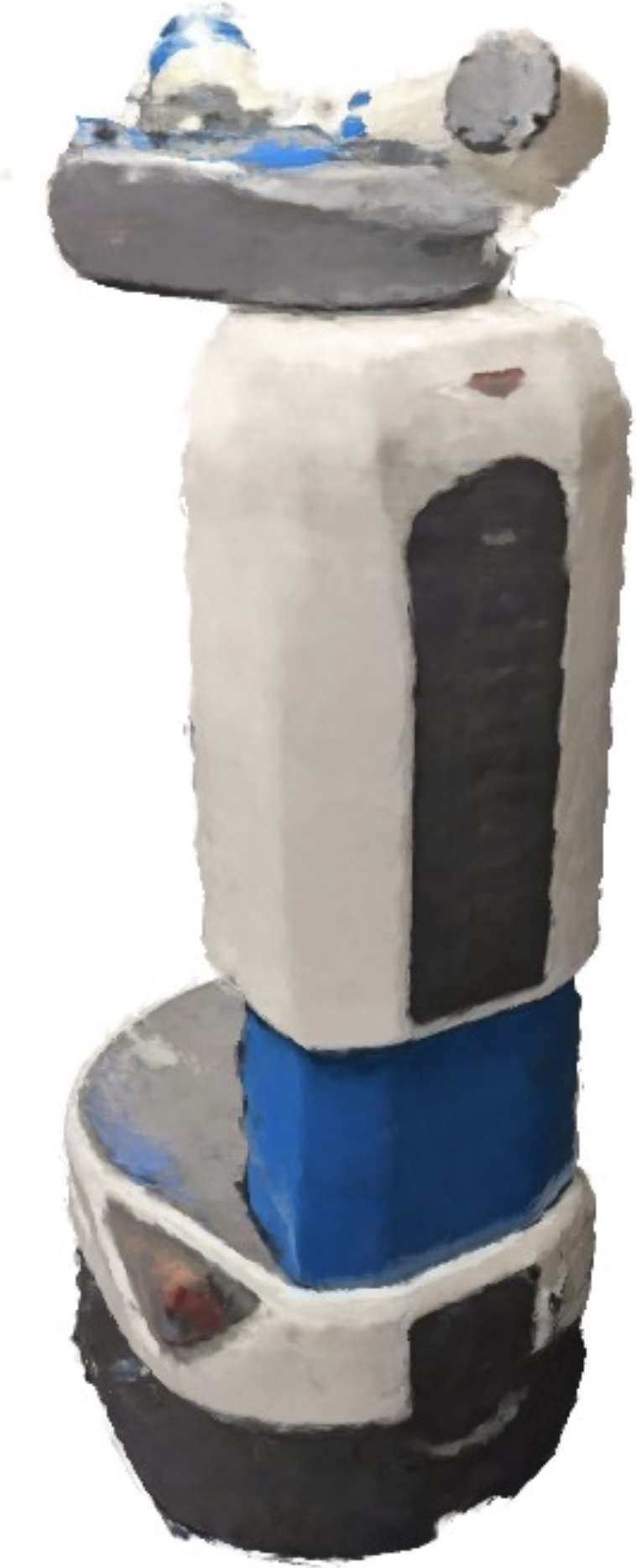} \qualresultners{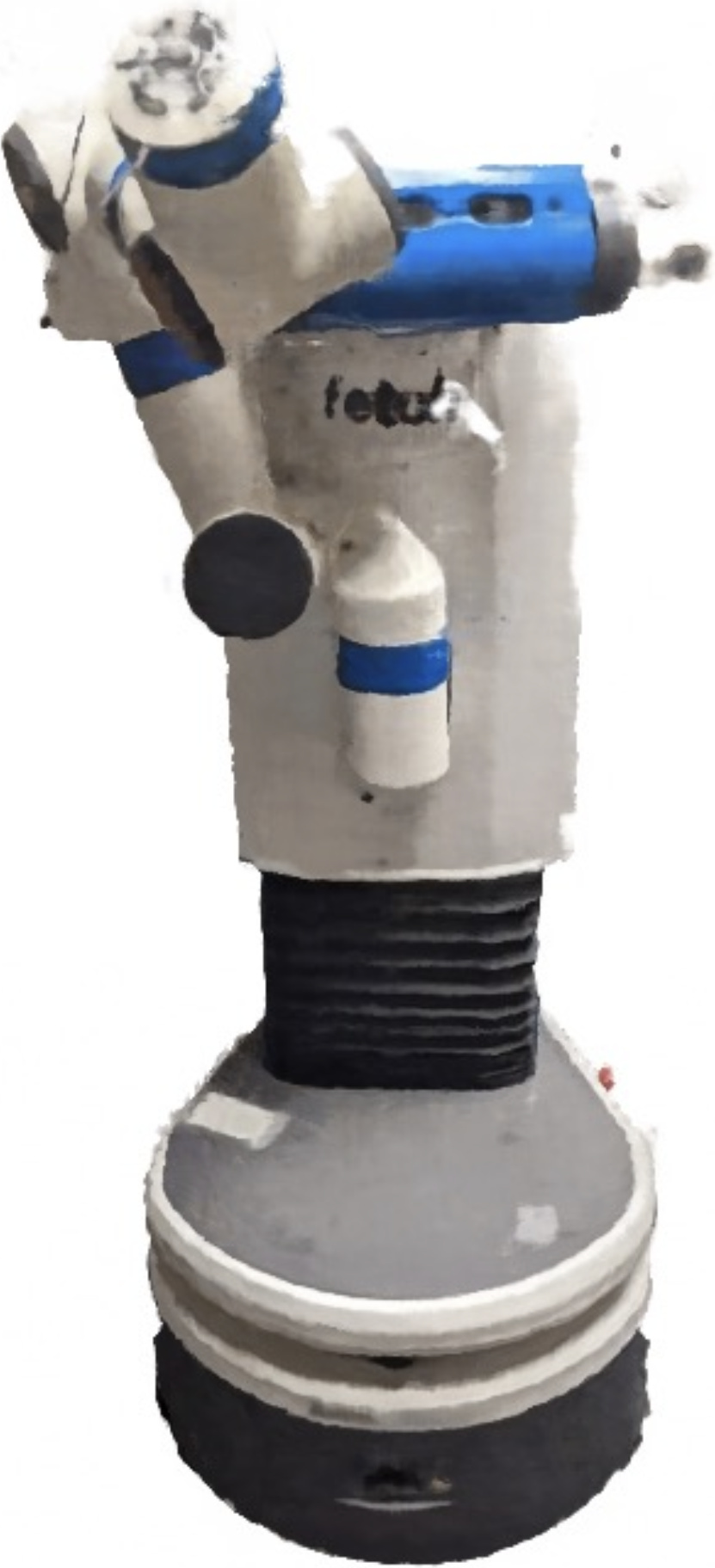} \\
        
        & NeRS \cite{zhang2021ners} & \qualresultners{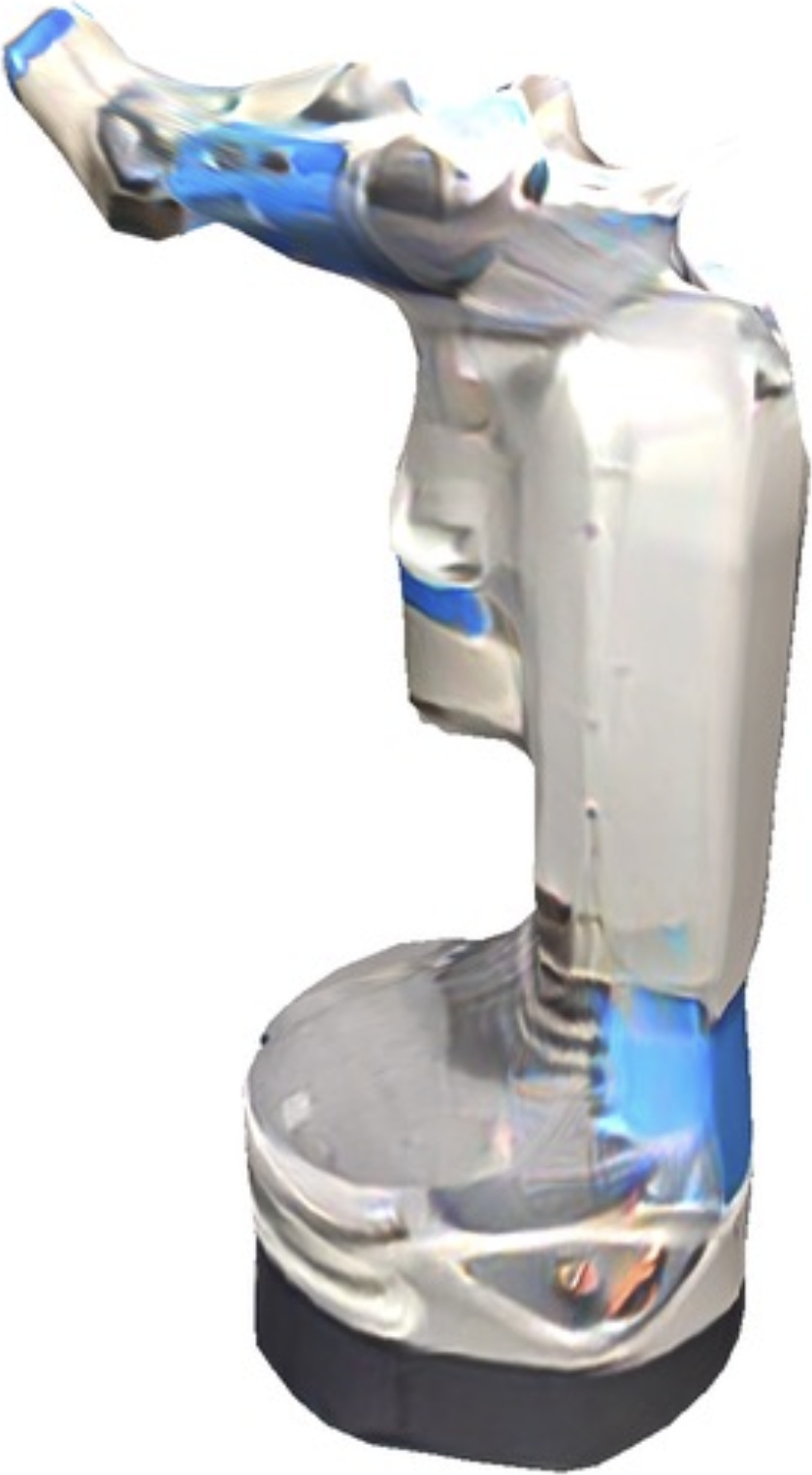} \qualresultners{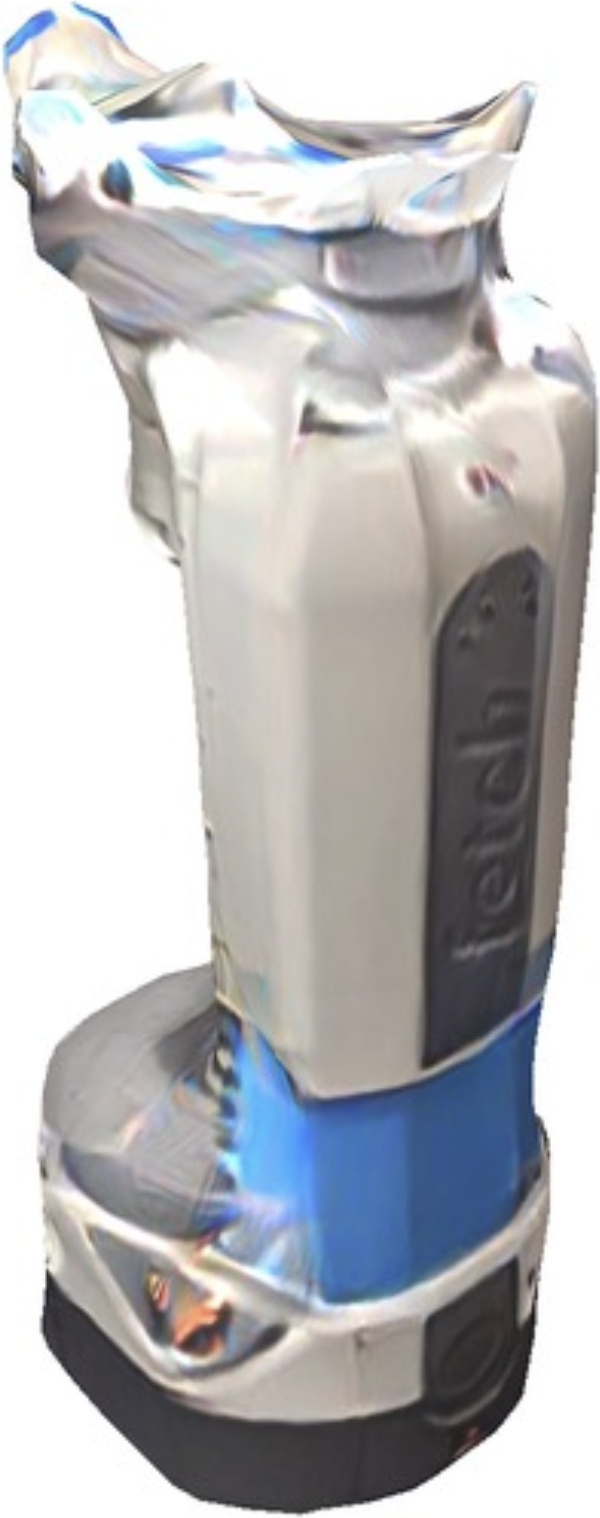} \qualresultners{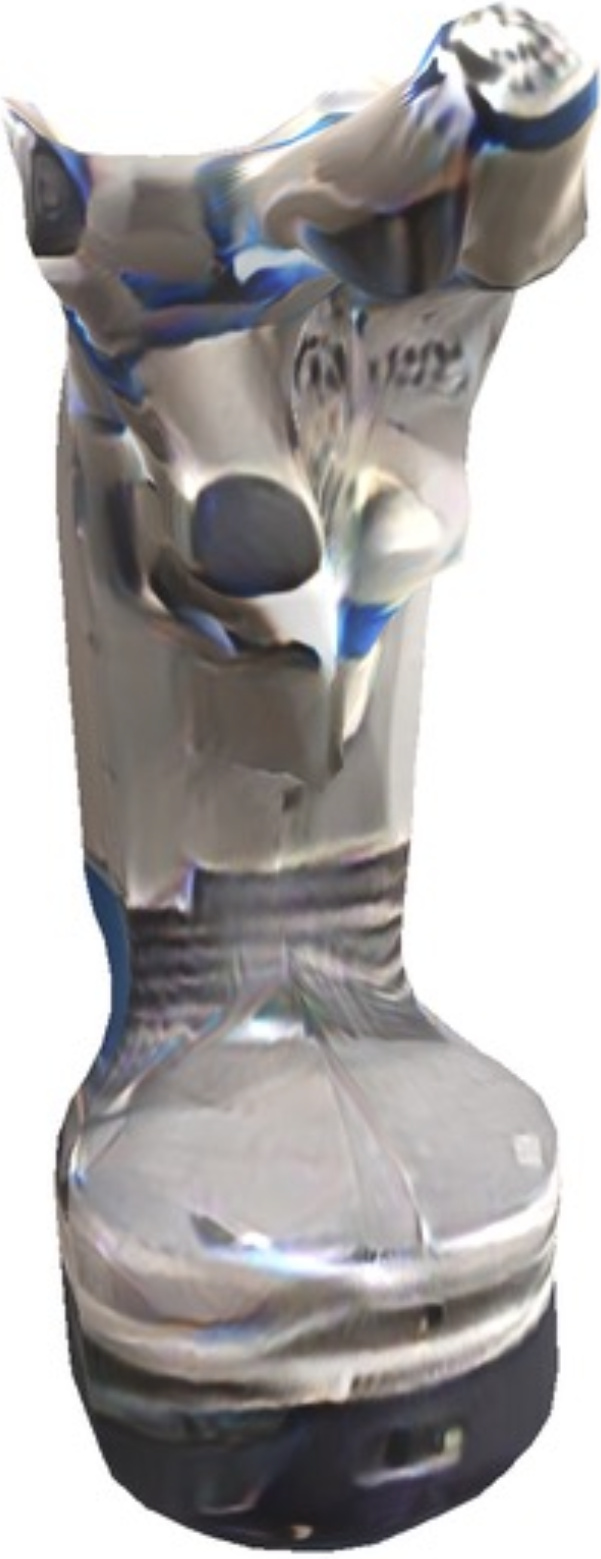} \\[30pt] 

        \raisebox{\dimexpr-\height+0.9cm\relax}[0pt][0pt]{\includegraphics[height=80pt]{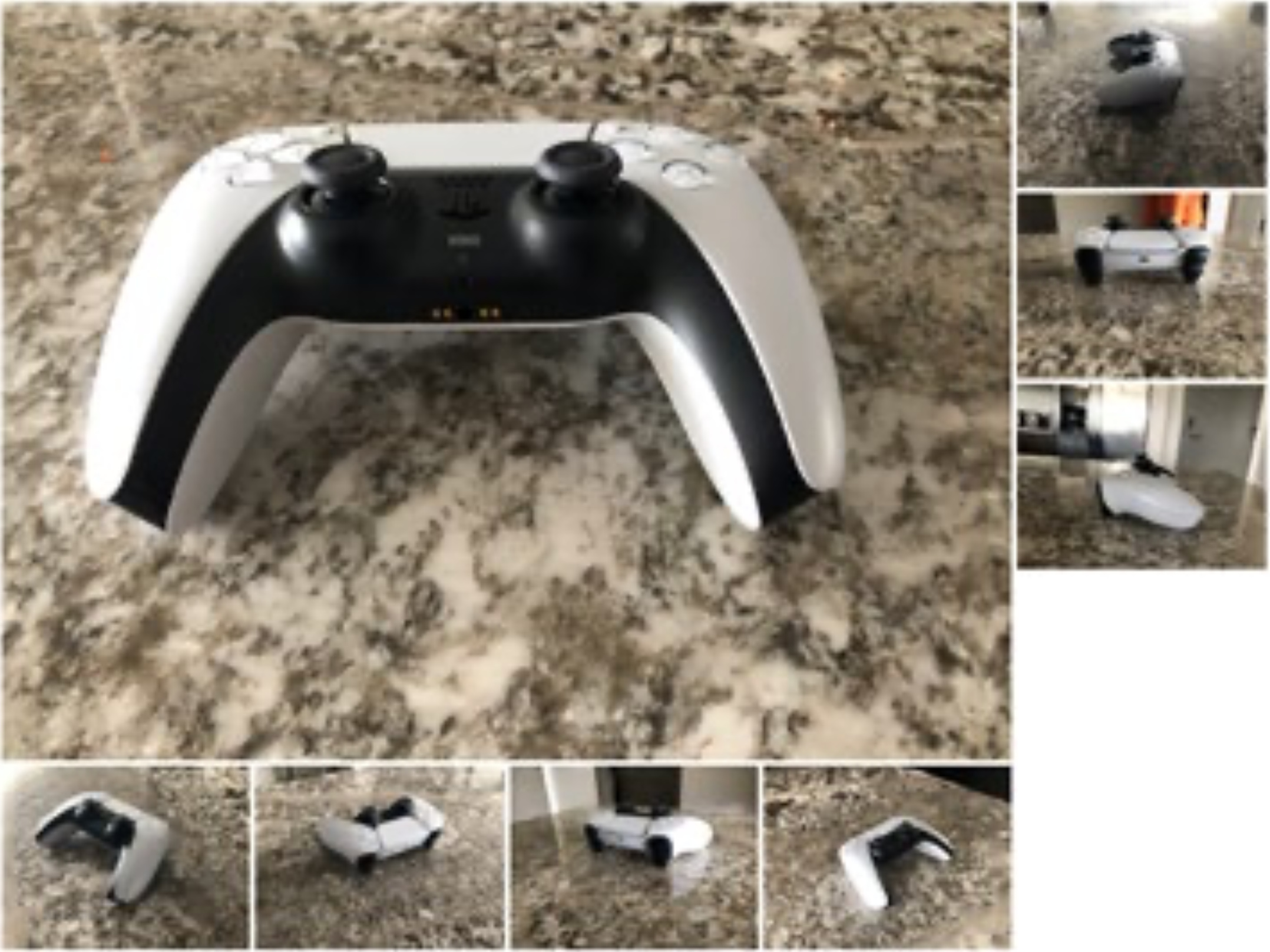}} & Ours & \qualresultners{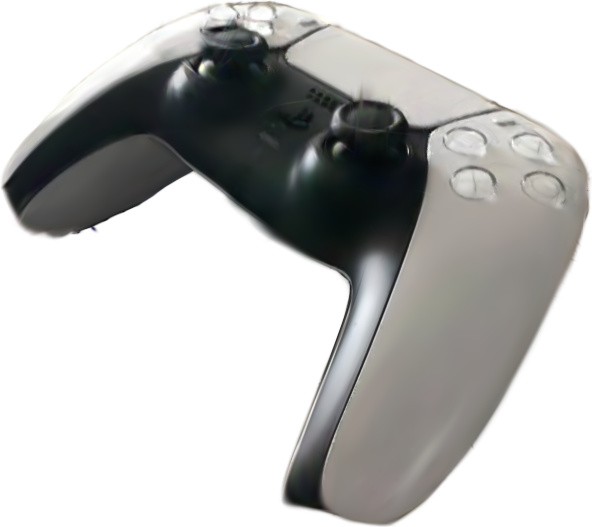} \qualresultners{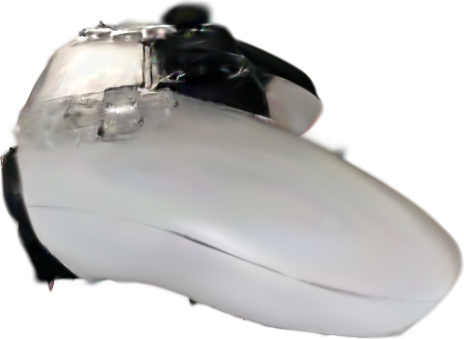} \qualresultners{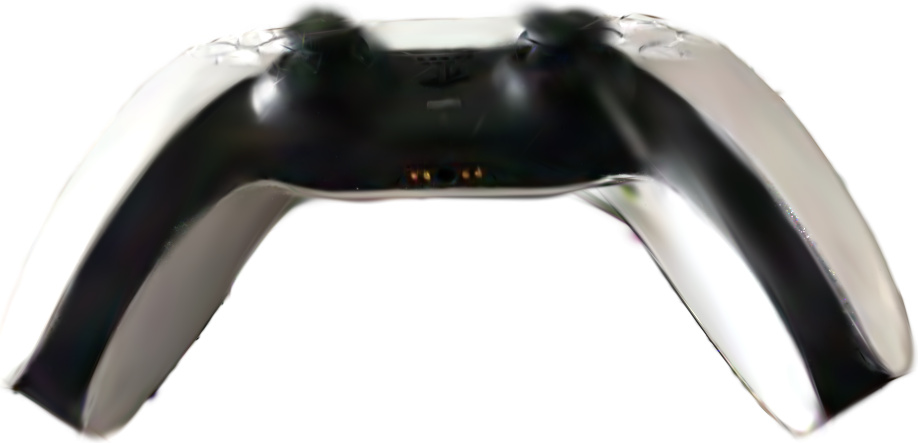} \qualresultners{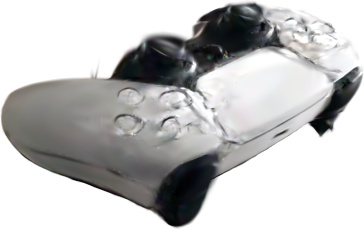} \\
        
        & NeRS \cite{zhang2021ners} & \qualresultners{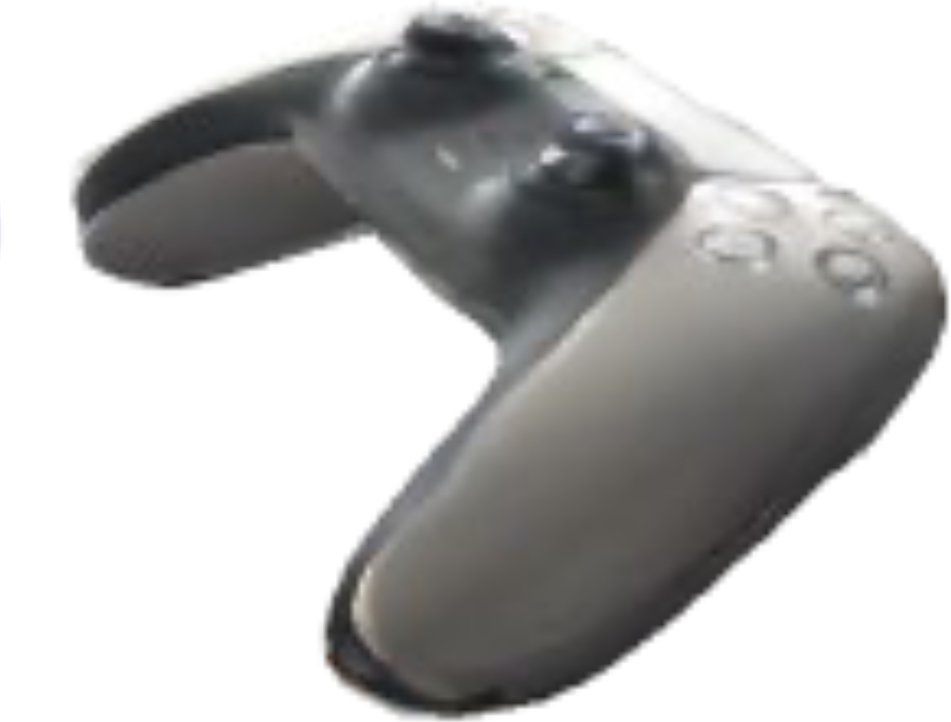} \qualresultners{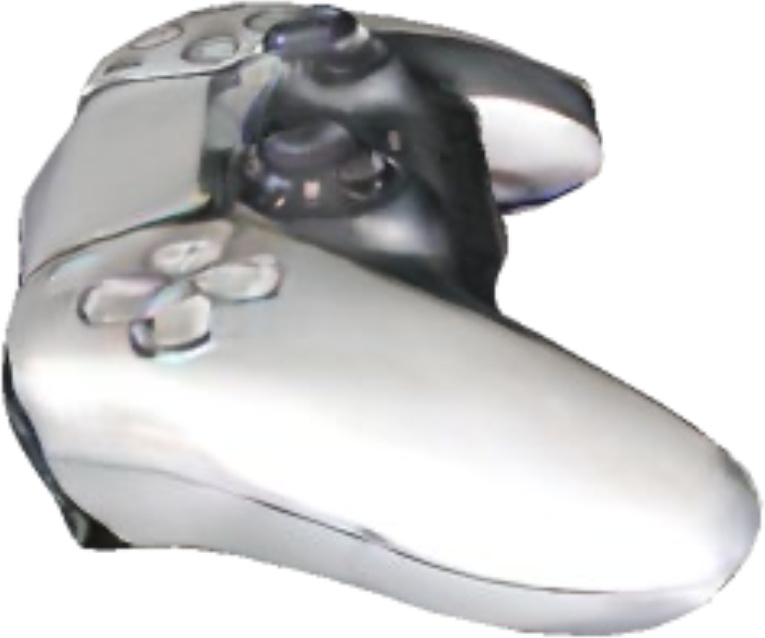} \qualresultners{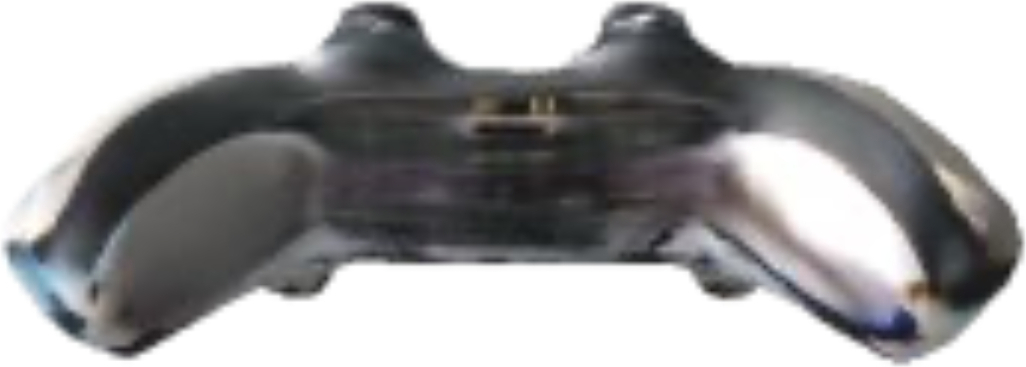} \qualresultners{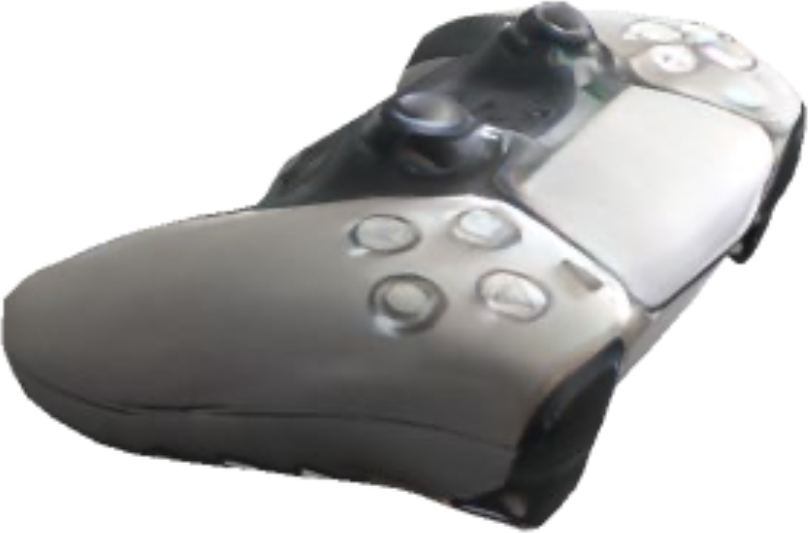} \\[30pt]

    \end{tabular}%
    }%
    \caption{Qualitative evaluation on the NeRS Misc dataset. The input images are shown on the left, and the novel viewpoint renderings are shown on the right. The renders on NeRS~\cite{zhang2021ners} and CADNeRF~\cite{wen2025cad} were reproduced from the NeRS dataset and the CADNeRF paper, where available.}
    \label{fig:ners_misc_qualitative}
\end{figure*}

The \textbf{NeRS Misc} dataset \cite{zhang2021ners} comprises $8$ various household objects captured sparsely in the wild. As in MVMC, we use the camera poses provided with the dataset as the camera-to-object poses. Silhouettes are matched against all ShapeNetCore~\cite{chang2015shapenet} categories except cars. Following NeRS, we qualitatively evaluate this set (\figurename~\ref{fig:ners_misc_qualitative}) and compare against NeRS~\cite{zhang2021ners} and CADNeRF~\cite{wen2025cad}. This dataset requires significant deformation from the deformation network, as several objects have no counterpart category in ShapeNetCore. Therefore, the consensus registration selects the closest geometrically compatible shape instead (a bottle for the fire hydrant, a table for the PS5 controller) and the deformation field bridges the remaining category-level gap. The results demonstrate that our pipeline produces clean, artifact-free reconstructions from casual in-the-wild captures, even when the prior is only a rough geometric approximation of the imaged object rather than its actual design.

\section{Applications}
\label{sec:applications}
\begin{figure*}[!tp]
    \centering
    \begin{minipage}{0.32\linewidth}\centering
        \includegraphics[width=\linewidth]{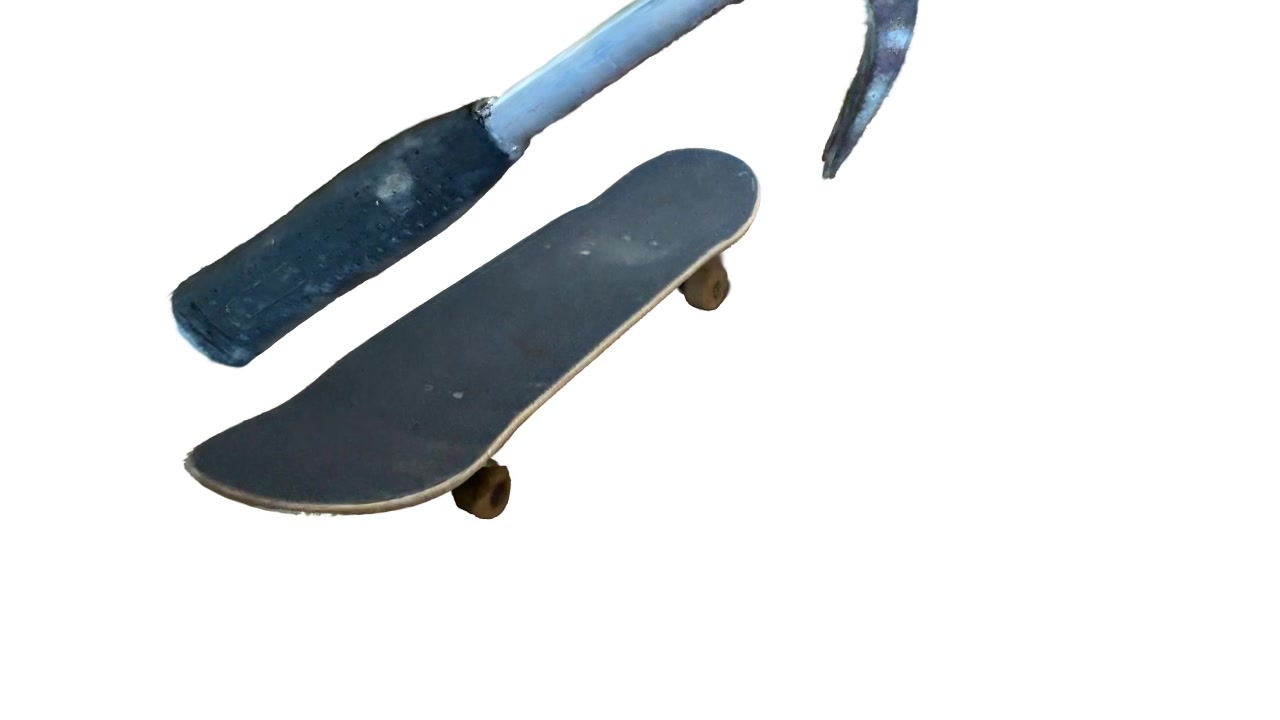}\\
        \footnotesize (a) Hammer falling
    \end{minipage}\hfill
    \begin{minipage}{0.32\linewidth}\centering
        \includegraphics[width=\linewidth]{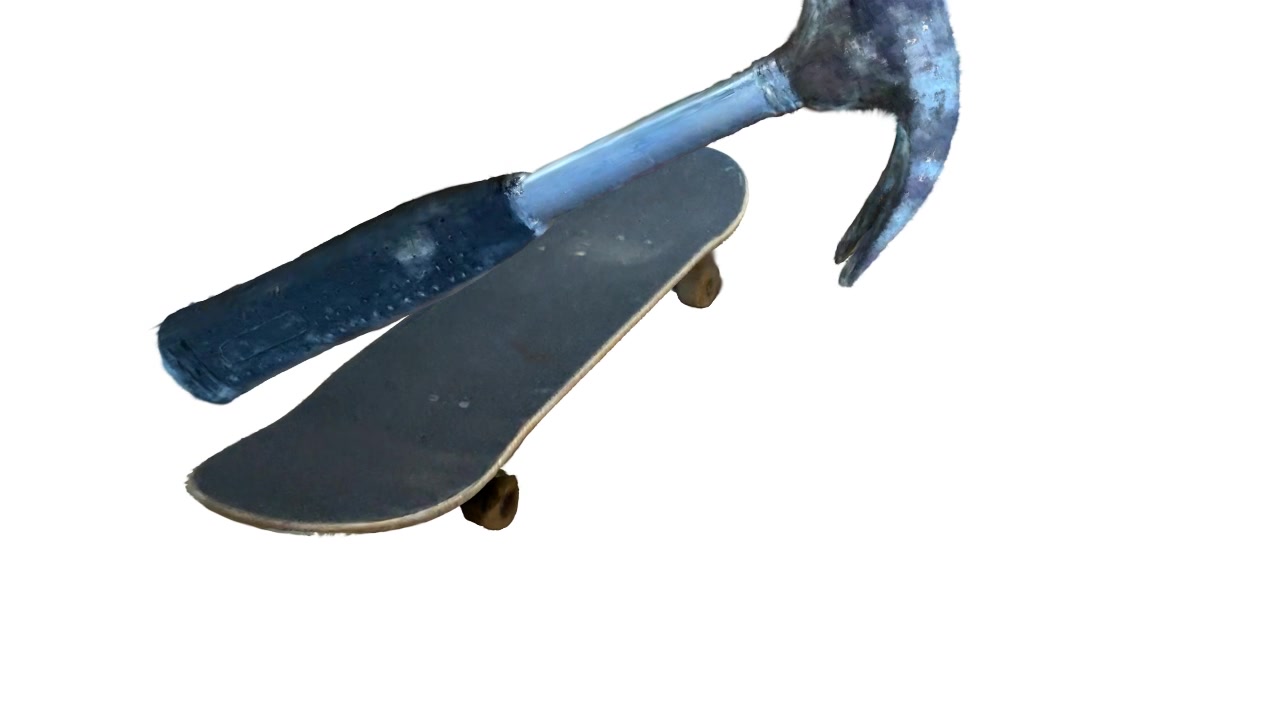}\\
        \footnotesize (b) Just after first contact
    \end{minipage}\hfill
    \begin{minipage}{0.32\linewidth}\centering
        \includegraphics[width=\linewidth]{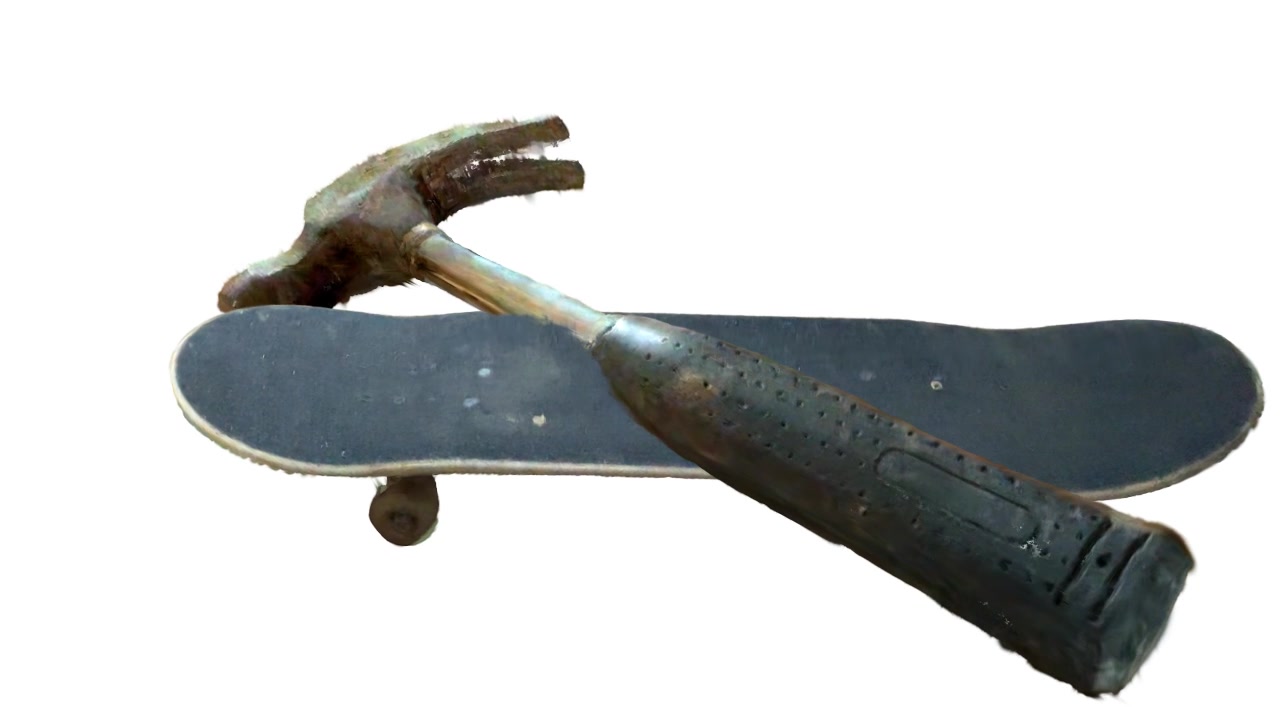}\\
        \footnotesize (c) At rest on the deck
    \end{minipage}
    \caption{Physically-based interaction between two reconstructed assets. The CAD meshes act as collision geometry while the Gaussians provide appearance: a rigid-body simulation drops the reconstructed hammer onto the reconstructed skateboard, and the poses it produces are applied to the surface-anchored Gaussians, which follow rigidly. Contact is resolved from the true object geometry---not a bounding proxy---and rendered photorealistically.}
    \label{fig:app_physics}
\end{figure*}

Because our method aligns the CAD model to the reconstruction and anchors every Gaussian to the CAD surface, the fitted asset inherits the CAD model's structure, semantic information, and motion properties (if available). We illustrate two example applications enabled by this:

\noindent\textbf{Physically-based interaction.} The CAD mesh serves as a collision proxy while the Gaussians provide photorealism. We drop a reconstructed hammer onto a reconstructed skateboard and run a rigid-body simulation on the two deformed CAD meshes. The interaction is resolved from the true object geometry and rendered photorealistically throughout (\cref{fig:app_physics}).

\begin{figure}[!htbp]
    \centering
    \begin{minipage}{\linewidth}\centering
       \includegraphics[width=\linewidth]{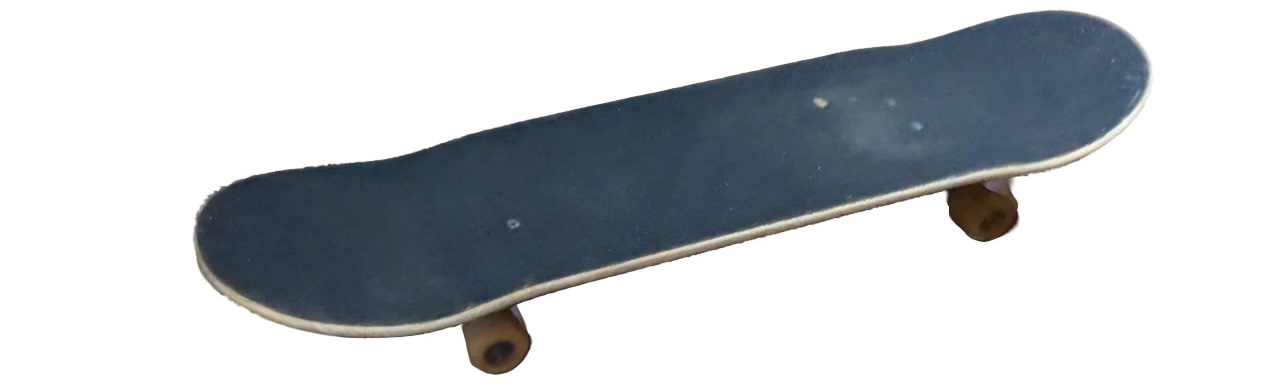}\\
        \footnotesize (a) Assembled
    \end{minipage}\\[2mm]
    \begin{minipage}{\linewidth}\centering
        \includegraphics[width=\linewidth]{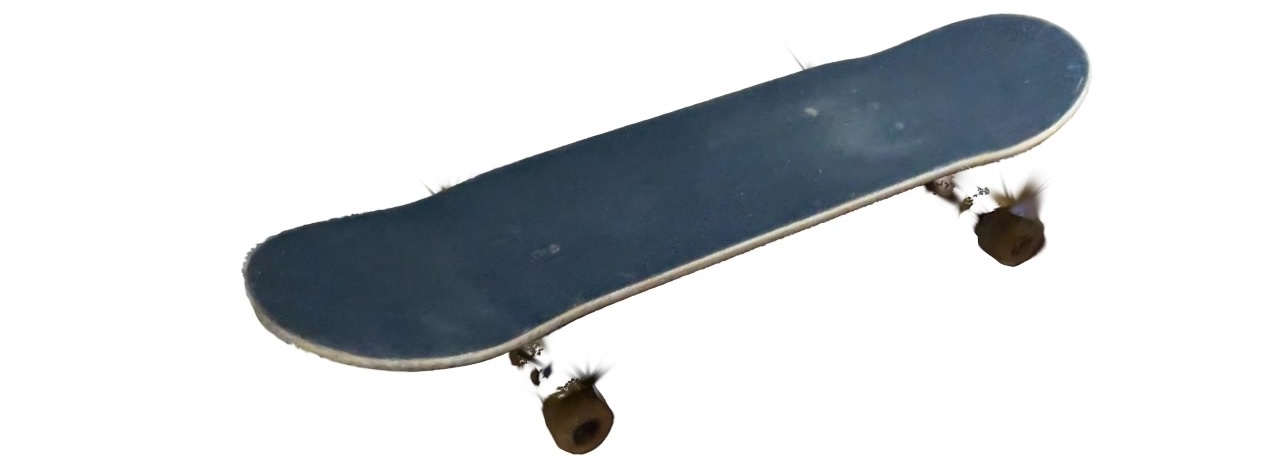}\\
        \footnotesize (b) Exploded
    \end{minipage}
    \caption{Photorealistic exploded view. Because each Gaussian is anchored to a CAD part, translating individual parts of the CAD model carries their Gaussians along with them: the skateboard's wheels slide off their axles while the deck stays fixed. The part structure comes from the CAD model; the appearance comes from the reconstruction.}
    \label{fig:app_explode}
\end{figure}

\noindent\textbf{Exploded views.} Because the Gaussians are bound to CAD parts, the reconstruction inherits the CAD model's part structure. Translating individual parts turns the photorealistic reconstruction into an exploded view: here the skateboard's wheels slide off their axles while the deck stays fixed (\cref{fig:app_explode}). This links the CAD model's semantic, editable structure to the photorealistic appearance of the reconstruction. Both applications are shown animated across novel views in Online Resource~4.

\section{Limitations}
\label{sec:limitations}
The method currently requires a good initial estimate of the relative calibration, whether from manual pre-calibration or SfM on the sparse views, both for silhouette matching to function properly and for the camera pose optimization to converge to a good minimum. The results are best when the CAD library contains a geometrically similar shape to the depicted object in the posed images. As the NeRS Misc experiments and sphere initialization experiments show, this is a loose requirement. Nevertheless, reconstruction quality in weakly observed or ambiguous regions benefits from a closer initial CAD match. Finally, the deformation field is optimized for reconstruction, not measurement: combined with the splats, it produces visually accurate renders, but the underlying deformed mesh is not itself optimized to be a plausible surface. We leave improving the surface quality of the deformed mesh toward metrology-grade measurement to future work.

\begin{figure}[!htbp]
    \centering
    \includegraphics[width=\linewidth]{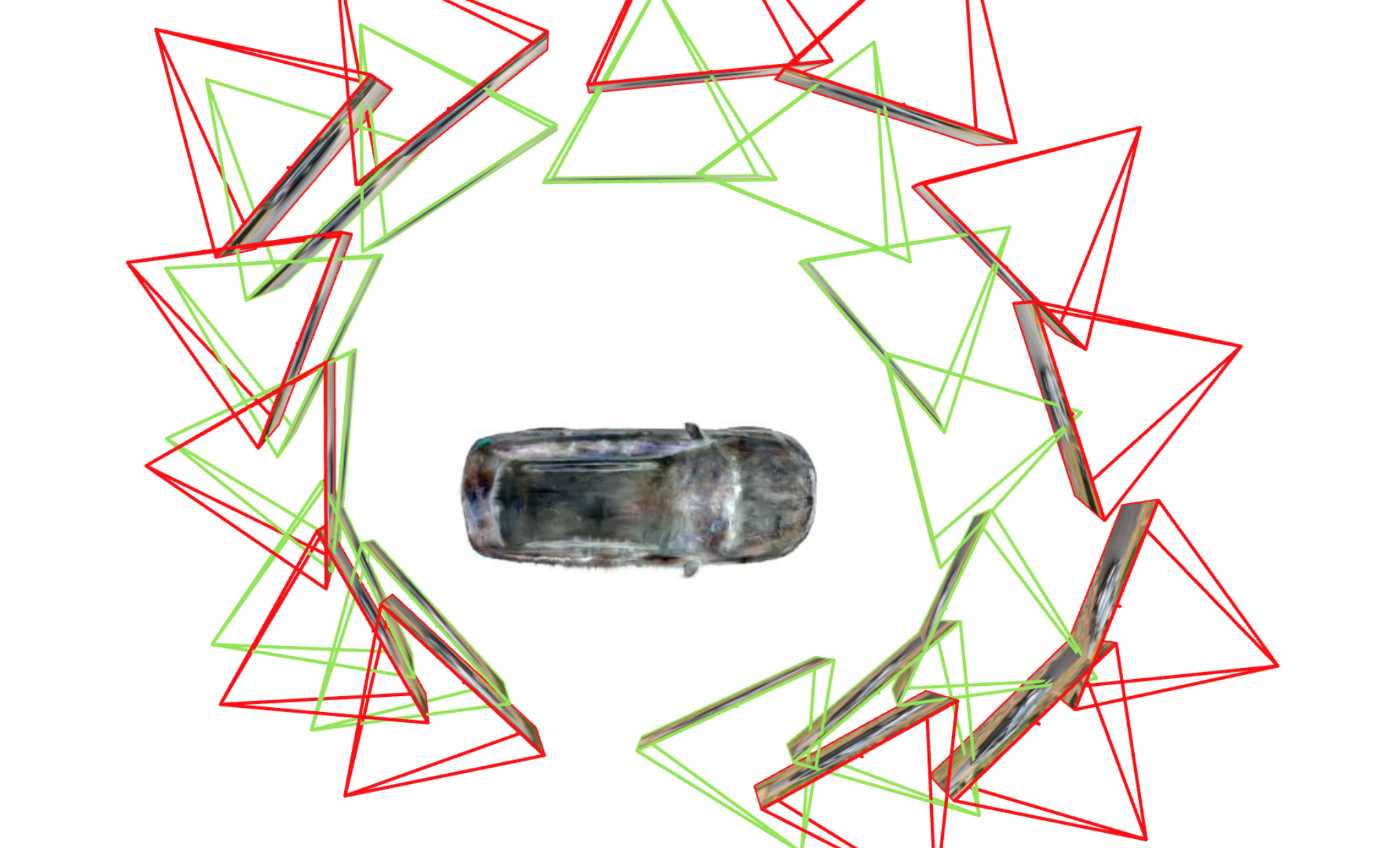}
    \caption{Failure case: extreme-viewpoint extrapolation. The capture orbits the car at mid-height; rendered from directly overhead, far outside the observed viewing directions, the appearance breaks down, although the geometry stays on the CAD surface.}
    \label{fig:failure}
\end{figure}

Our method generalizes well to novel views near the capture trajectory. Extreme extrapolation, e.g., rendering a top-down view of a car observed only from the sides (\cref{fig:failure}), yields unusable results, because the appearance model---per-splat opacities and view-dependent spherical harmonics---remains partly overfit to the training viewpoints. The SH-neighbor smoothness and splat-dropout regularizers of \cref{sec:reg} reduce this overfitting but do not remove it entirely. Extending the appearance model to generalize to such out-of-distribution viewpoints requires priors on object materials as well as their geometry.

\section{Conclusion}

We presented \textbf{CADSplat}, a novel approach to reconstruct a photorealistic, geometrically clean digital twin from a handful of wide-baseline images by anchoring 3D Gaussians to a retrieved CAD model, registering the cameras to that model through silhouette consensus, and jointly refining the registration, a non-rigid deformation field, and the appearance. In our experiments, it outperforms unconstrained, few-shot, and mesh-texturing baselines, and its quality degrades gracefully down to three views. We demonstrate that much of the rendering-quality improvement over unconstrained 3DGS comes from the constrained deformable representation. The CAD geometry prior matters most when the visual information in the images is ambiguous (e.g., extreme view sparsity or significant self-occlusion). Independently of rendering quality, the CAD prior yields camera-to-object poses, which enables applications such as markerless augmented reality registration, per-image object pose estimation, physically based interaction, and part-aware editing. Improving the deformation field toward metrology-grade surface quality and extending the framework to extreme-view synthesis are the main directions for future work.

\FloatBarrier 

\backmatter

\bmhead{Acknowledgements}
This work was supported by the Flanders Make ADDIL project. We gratefully acknowledge imec IDLab for computational resources via the imec iLab.t infrastructure.

\bibliography{references}

\end{document}